\documentclass[10pt,twocolumn,letterpaper]{article}

\usepackage[pagenumbers]{cvpr} % To force page numbers, e.g. for an arXiv version

\usepackage{graphicx}
\usepackage{booktabs}
\usepackage[table]{xcolor}
\usepackage{pifont}
\usepackage{cuted}
\usepackage{multirow}
\usepackage{algorithm}
\usepackage{algpseudocode}
\usepackage{float}

\definecolor{green}{HTML}{C2E5D3}
\definecolor{lightgreen}{HTML}{E1F2E9}
\definecolor{orange}{HTML}{FFDBBB}
\definecolor{lightorange}{HTML}{FFF3E8}
\definecolor{retPink}{HTML}{FF8088}
\definecolor{retGreen}{HTML}{80E890}

\newcommand{\effG}[2]{\cellcolor{retGreen!#1}#2}
\newcommand{\effP}[2]{\cellcolor{retPink!#1}#2}

\usepackage{microtype}

\renewcommand{\paragraph}[1]{\vspace{.5em}\noindent\textbf{#1}}

\definecolor{cvprblue}{rgb}{0.21,0.49,0.74}
\usepackage[pagebackref,breaklinks,colorlinks,allcolors=cvprblue]{hyperref}

\makeatletter

\makeatletter
\@ifpackagewith{cvpr}{review}{%
    \newcommand{\teaserfile}{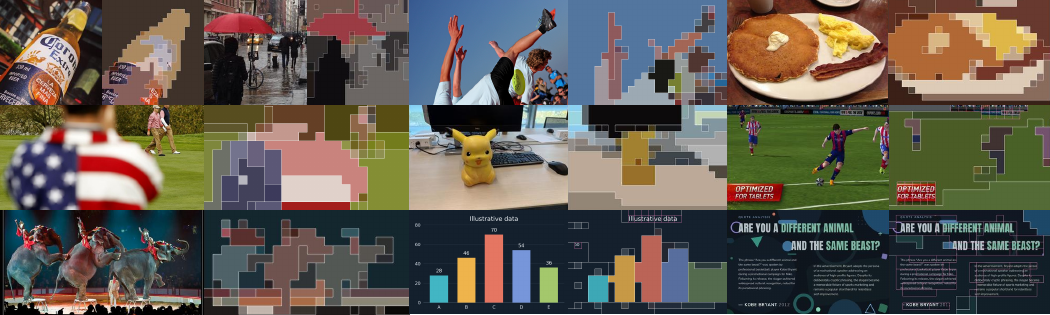}
}{%
    \newcommand{\teaserfile}{images/teaser_b32_preprint.pdf}
}

\apptocmd{\@maketitle}{%
    \noindent\begin{minipage}{\textwidth}
        \centering
        \captionsetup{type=figure}
        \includegraphics[width=\linewidth]{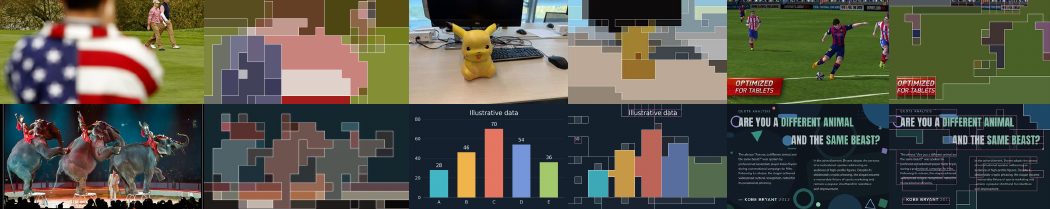}
        \caption{\textbf{\method{} tokenization from 576 tokens to 51 tokens.} The \method{} tokenizer preserves semantics and protects key text patches. We represent merged regions by their mean colour, with identified text patches kept raw. At the lowest evaluated budget ($K_m=51$ tokens before LLM and $K=32$ on average across layers), the original images remain easily recognizable in the merged views on the right.}
        \label{fig:teaser}
    \end{minipage}\par
    \vspace{8pt}
}{}{}
\makeatother

\def\confName{CVPR}
\def\confYear{2027}

\newcommand{\method}{Varest}

\title{Find the Fungible \includegraphics[scale=0.5]{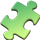}: Variable-Resolution Tokenization that Compacts Semantics with a Granularity Dial as a VLM ``Catalyst''}

\author{
\href{https://orcid.org/0009-0007-2342-5350}{\includegraphics[scale=0.06]{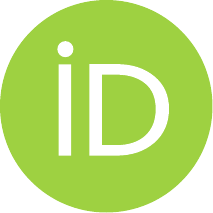}\hspace{1mm}Tianhao Fu} $^\dagger$\textsuperscript{, 1, 2, \thanks{
Project lead with correspondence at \href{mailto:tianhao.fu@mail.utoronto.ca}{tianhao.fu@mail.utoronto.ca}.\\
Project website: \url{https://varest.github.io}.\\
Project repository: \url{https://github.com/Varest/Varest}.
}}\and
\href{https://orcid.org/0009-0008-3817-0589}{\includegraphics[scale=0.06]{images/icons/orcid.pdf}\hspace{1mm}Austin Wang} $^\dagger$\textsuperscript{, 3}\and
\href{https://orcid.org/0009-0005-7789-3876}{\includegraphics[scale=0.06]{images/icons/orcid.pdf}\hspace{1mm}Charles Chen} $^\dagger$\textsuperscript{, 1}\and
\href{https://orcid.org/0009-0008-0149-2250}{\includegraphics[scale=0.06]{images/icons/orcid.pdf}\hspace{1mm}Roby Aldave-Garza} \textsuperscript{4}\and
\href{https://orcid.org/0009-0007-8118-4049}{\includegraphics[scale=0.06]{images/icons/orcid.pdf}\hspace{1mm}Victor Xie} \textsuperscript{3}\and
\hspace{0.9\linewidth}
\\
\small$^\dagger$ Equal contributions.\hspace{2mm}
\small\textsuperscript{1} University of Toronto\hspace{2mm}
\small\textsuperscript{2} Vector Institute\hspace{2mm}
\small\textsuperscript{3} McGill University\hspace{2mm}
\small\textsuperscript{4} University of Waterloo\\
}

\begin{document}
\maketitle

\begin{abstract}
Efficient vision-language models should preserve a useful representation of the whole image, even under tight budgets. However, aggressive pruning or indiscriminate merging can erase critical details and disrupt the spatial structure. We thus propose \method{}, which maximizes \emph{semantic fungibility}: tokens describing coherent content can be summarized together with minimal information loss. Object boundaries distilled from a visual foundation model guide grouping, while text localization protects details needed for reading. At lower budgets, \method{} merges the image into coarser semantic regions, while maintaining full spatial coverage. After early language decoder layers process this representation, a learned selector prunes visual tokens that later layers no longer need, reducing repeated access to already integrated information. Across three LLaVA backbones and 13 benchmarks, \method{} outperforms 10 training-free and 5 trained baselines in mean performance retention at every evaluated budget, preserving 86.10--92.33\% of uncompressed performance using only 5.56\% of tokens. This SOTA performance further generalizes to modern VLMs like Qwen3.8-27B.
\end{abstract}

\makeatletter
\@ifpackagewith{cvpr}{review}{%
  \begingroup
  \renewcommand{\thefootnote}{}
  \footnotetext{Core implementation of the method is attached in the supplementary material. Source code, model weights, and raw evaluation results will be released. Interactive visualizations will be available on the project website.}
  \endgroup
}{}
\makeatother

\section{Introduction}
\label{sec:introduction}

\begin{figure*}
    \centering
    \includegraphics[width=1\linewidth]{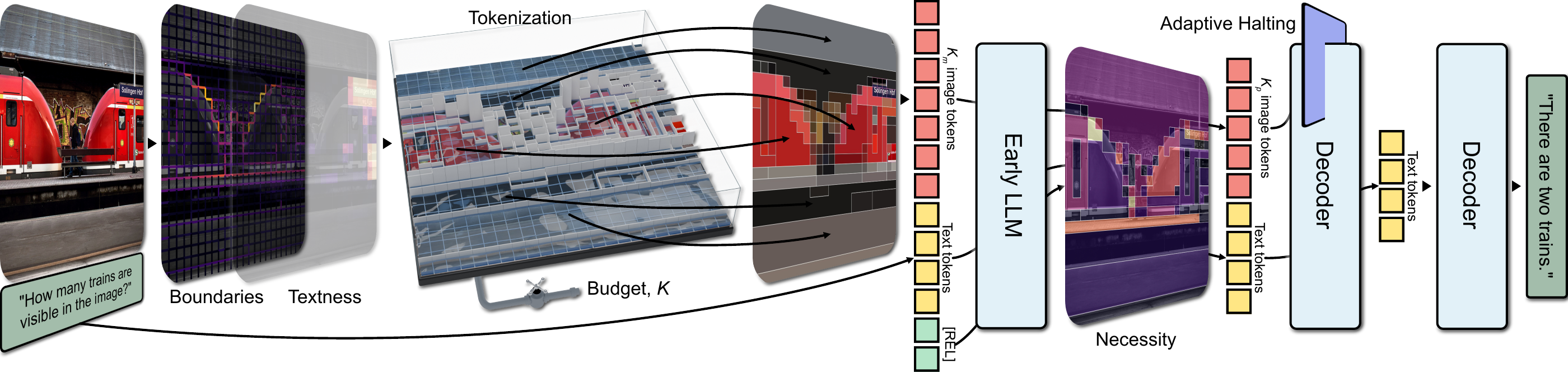}
    \caption{\textbf{Overview of \method.} Object boundaries and textness are passed through the \method{} tokenizer with post-projector visual tokens, which are then partitioned into $K_m$ semantic regions and each region is merged into one image token. At layer $L/2$, four learned \texttt{[REL]} registers (the figure is not to scale) assign a relative necessity score to each image token and keep the $K_p$ highest-scoring tokens for the remaining decoder layers. Optional adaptive halting removes all remaining image tokens once the stability and confidence criteria are met.}
    \label{fig:method}
\end{figure*}

Visual-token compression reduces the substantial compute that the huge number of image tokens imposes on the language-model decoder of vision-language models (VLMs). Prior work has explored merging and pruning using various signals, including feature geometry~\cite{bolya2023tome,alvar2025divprune,chen2026otprune,ma2026apet}, attention scores~\cite{yang2025visionzip,yang2025vflowopt,chen2024fastv,xing2025pyramiddrop,zhang2025sparsevlm,wu2026hidrop,huang2026nuwa,singh2026duet}, and learned retention utility~\cite{takezoe2026learnpruner}. These are natural ways to approach compression, yet they do not respect the semantics. Merging across two neighbouring objects creates a token that loses the semantics of both. Pruning too early hurts spatial coverage, leaving whole image regions with no token to represent them. Effective compression thus requires deciding where aggregation is most efficient and when to drop which image tokens.

We frame this requirement as a property of a set of tokens, \emph{semantic fungibility}: tokens depicting a person's head, body, and limbs are interchangeable with their mean because they are linked by a semantic prior of human structure learned by the LLM, and thus form a fungible set. We can decrease the semantic granularity by merging two instances of person into ``people''. In contrast, a set of tokens that mixes a person and a snowboard, which have distinct functions and categories, is more complex (higher variance) and thus less fungible. Text has low fungibility for literal reading, since neighbouring characters can have different meanings. These two exact experiments of cross-instance and cross-category object merging and text merging are presented with the mentioned person-snowboard image in Appendix~\ref{sec:supp_additional_experiments}. Intuitively, if all sets are wisely chosen to maximize fungibility, we expect the merged view, similar to the ones in Figure~\ref{fig:teaser}, to preserve as much information as possible. Therefore, we can quantitatively approximate the total fungibility through the merged view's reconstruction quality against the original image using RMSE, PSNR, SSIM~\cite{wang2004ssim}, and LPIPS~\cite{zhang2018lpips}. The guiding principle is therefore simple: \emph{wisely partition the image to maximize the total fungibility of all token sets}.

We propose \textbf{\method{}}, which implements this principle by combining semantic variable-resolution tokenization and relative necessity pruning. Lightweight object and text localizers provide the high-level semantic prior. Since the LLM has not yet integrated visual evidence, we ensure full coverage by treating budget as a semantic granularity dial. After early decoder layers process the merged tokens, learned \texttt{[REL]} registers retain only the tokens that remain necessary. Optional adaptive halting can drop all image tokens after visual integration. One unified training serves all budgets per VLM.

Our key contributions are as follows:
\begin{itemize}
    \item We introduce \method, a semantics-preserving variable-resolution tokenizer guided by object and text localizers. It allocates higher resolution to low-fungibility features while maintaining full spatial coverage by controlling semantic granularity to meet the desired visual-token budget.
    \item We combine it with learned decoder-stage necessity pruning, allowing broader visual access during early processing and retaining only the tokens the LLM still needs. Optional adaptive halting can further save compute by dropping all image tokens after the estimated information horizon.
    \item Across three LLaVA backbones (LLaVA-1.5-7B~\cite{liu2024llava15}, LLaVA-1.5-13B~\cite{liu2024llava15}, and LLaVA-NeXT-8B~\cite{li2024llavanext-strong}) and 13 benchmarks, \emph{\method{} achieves the highest mean performance retention among 15 baselines} at every evaluated budget with a lower computational cost. We also show that this performance generalizes to Qwen3.8-27B~\cite{qwen38}.
\end{itemize}
\section{Method}
\label{sec:method}

% Throughout the text, we define $K=\frac{1}{L}\sum_i^LK_i$ as the layer-averaged image-token budget. $K_m$/$K_p$ denote the merging/pruning-stage target image-token budget, respectively, and $\frac{K_m+K_p}{2}=K$. $r_m$/$r_p$ denote the merging/pruning-stage token retain ratio and $r=\frac{r_m+r_p}{2}$. For LLaVA-1.5, $r=\frac{K}{576}$. For LLaVA-NeXT, since its \texttt{AnyRes} yields a dynamic number of image tokens, $K$ is the maximum budget and $r$ is the exact retain ratio regardless of the actual number of image tokens. $k$ denotes the number of object regions. $L$ denotes the total number of decoder layers of the LLM. $\lambda$ denotes the textness weight. $\tau$ denotes the text localizer confidence threshold.

\method{} allocates visual resolution across the image and visual access across language decoder depth, as shown in Figure \ref{fig:method}. Given an image and user prompt, we first partition the image into regions guided by the object boundaries and textness. Merging within these regions produces $K_m$ image tokens at different resolutions while preserving full spatial coverage. Early decoder layers process this representation together with the prompt and four \texttt{[REL]} tokens discussed in Section~\ref{sec:rel_main}. At the decoder midpoint, \texttt{[REL]} selects $K_p$ image tokens that the LLM still needs the most. The budget splitting over depth follows $K_m=4K_p$. Optional adaptive halting, described in Section~\ref{sec:adaptive_halting_main}, can remove all image tokens at any point after layer 8 once triggered; if this occurs before the midpoint, the necessity pruning is skipped.

\subsection{Semantic Variable-Resolution Tokenization}
\label{sec:varest_main}

To maximize semantic fungibility, we prefer merging within semantic regions and preserve distinctions at transitions. We use a lightweight object localizer to supply semantic prior. The semantic granularity that decides whether we further divide a car into fine-grained components like car doors is dialled by the desired token budget. Because literal reading of text is sensitive to any compression, we use a separate text localizer to pin all text cells at native resolution, if the budget allows. Algorithm~\ref{alg:varest} provides an overview of this tokenization and explains the corresponding part in Figure~\ref{fig:method}.

\begin{algorithm}[t]
\caption{VAriable-RESolution Tokenization (\method)}
\label{alg:varest}
\small
\begin{algorithmic}[1]
\Require Image $I$, projected features $X$, token budget $K_m$
\State Predict object boundaries and textness from $I$
\State Separate text and non-text cells
\State Divide $K_m$ between both groups, prioritizing text detail
\For{each group}
    \State Initialize one region per cell
    \State Repeatedly merge regions with the weakest shared boundaries until the group's quota is reached
\EndFor
\State \Return Mean-pooled region features with position tags
\end{algorithmic}
\end{algorithm}

\paragraph{Semantic localization.}
Let $\mathbf{x}_i$ denote the projected image token of grid cell $i$. The object localizer predicts a boundary strength $B_{ij}$ between neighbour cells, and the text localizer predicts a textness probability $T_i\in[0,1]$ for each cell. They are distilled from the cosine distance form of DINOv3~\cite{simeoni2025dinov3} features and EasyOCR~\cite{jaidedai2020easyocr} bounding boxes, respectively. They are trained separately and remain frozen during VLM adaptation and inference. Since their outputs are continuous fields rather than masks, the same evidence can support different granularities. We augment object boundaries with text-border evidence through $B^*_{ij}=\max\{B_{ij},\lambda|T_i-T_j|\}$, with $\lambda=0.25$. This discourages merging across text transitions even within an otherwise uniform object.

\paragraph{Allocating the token budget.}
We divide cells into text and non-text regions using a textness threshold $\tau(r_m)=1-\frac{1}{2}\mathrm{clip}(r_m,0,1)$, where $r_m$ is the merging-stage token retention ratio. From $B^*$, we estimate the minimum number of non-text regions $k_{min}$. Cells with $T_i>\tau(r_m)$ remain uncompressed if the number of qualified text tokens $t$ is smaller than or equal to $K_m-k_{min}$; otherwise, we merge them into text regions too such that $t$ after merging allows the remaining non-text tokens to be partitioned into $k=K_m-t\ge k_{min}$ regions. Larger budgets therefore preserve more local distinctions, whereas smaller budgets summarize broader regions, which is semantic granularity.

\paragraph{Partitioning.}
Within each group, we use constrained mean-boundary-linkage agglomeration~\cite{arbelaez2011contour}: start from individual cells and keep merging the neighbours with the weakest shared boundary. For regions $R$ and $S$, the merge cost is
\begin{equation}
    d(R,S)=\frac{1}{|E(R,S)|}\sum_{(i,j)\in E(R,S)}B^*_{i,j},
    \label{eq:varest_partition}
\end{equation}
where $E(R,S)$ contains the grid edges joining the two regions. Averaging over their shared boundary reduces sensitivity to isolated weak edges. Agglomeration stops when each group reaches its allocated region count. This procedure prefers coarse representations within weakly separated content and distinction around strong boundaries.

\paragraph{Region representation.}
Each region becomes one image token by
\begin{equation}
    \mathbf{z}_R=\frac{1}{|R|}\sum_{i\in R}\mathbf{x}_i+\operatorname{MLP}\!\left(c_x(R),c_y(R),\sqrt{\frac{|R|}{N}}\right),
    \label{eq:varest_merge}
\end{equation}
where $N$ is the number of cells in the grid and $(c_x,c_y)$ is the normalized region centroid. The learned position tag supplies the location and extent of the pooled content. Regions are emitted in raster order of their first cell.

For LLaVA-NeXT-8B, we partition the base view and the stitched, unpadded high-resolution view separately, allocating tokens proportionally to cell counts. On Qwen3.8-27B, we partition one grid after the native visual merger. Region coordinates and extent are normalized within each grid.

\begin{figure}
    \centering
    \includegraphics[width=\linewidth]{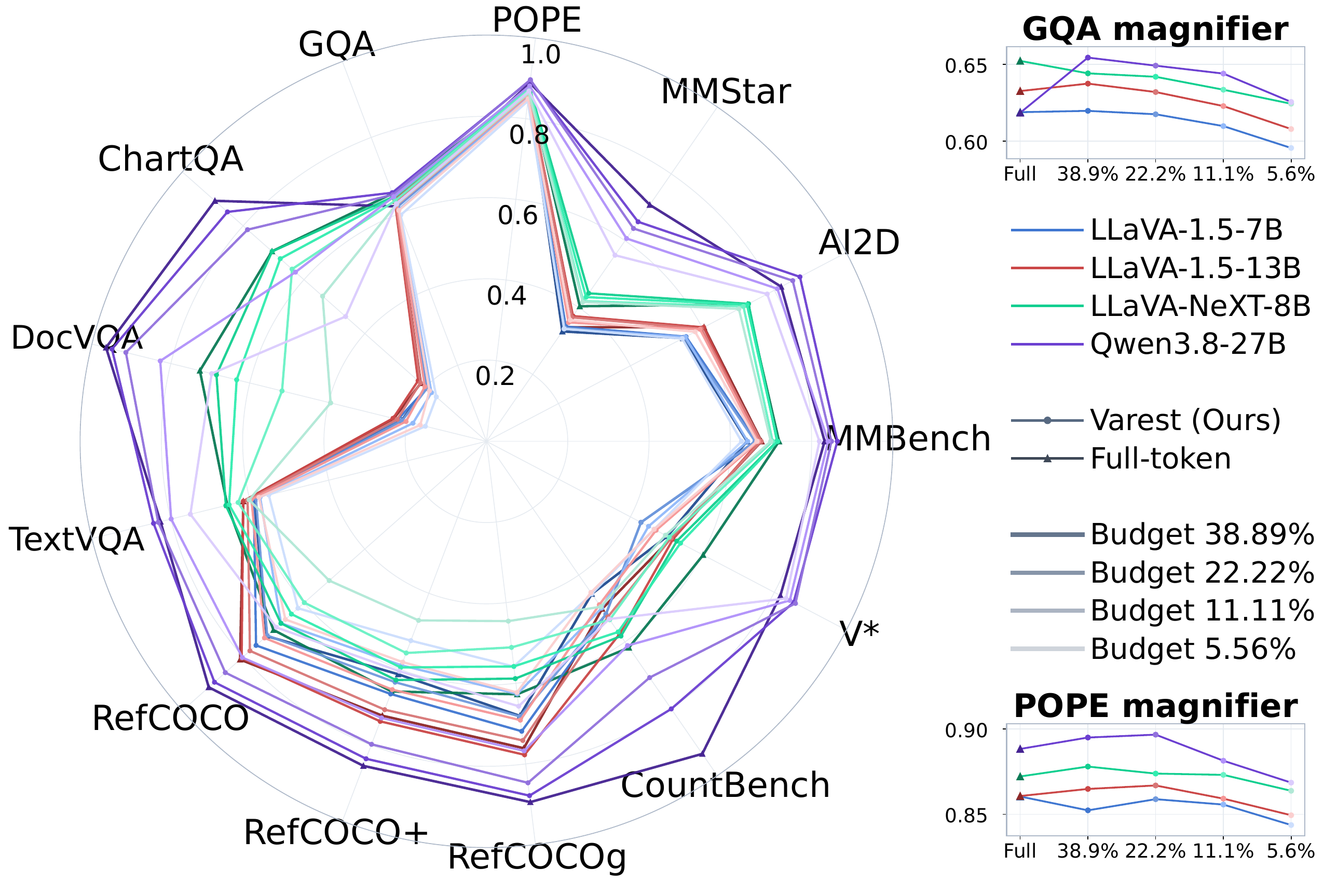}
    \caption{\textbf{Per-benchmark performance of \method{} across backbones and token budgets.} Text-based and grounding benchmarks benefit more from larger budgets compared to general VQA.}
    \label{fig:radar}
\end{figure}

\subsection{\texttt{[REL]} Necessity Pruning}
\label{sec:rel_main}

Deeper LLM layers need fewer visual tokens~\cite{chen2024fastv}. Inspired by PyramidDrop~\cite{xing2025pyramiddrop}, we retain more tokens before the midpoint and prune aggressively at layer $L/2$ to keep the tokens that the LLM still needs. The \texttt{[REL]} registers and scoring head learn relative necessity scores for image tokens, ranking them by their estimated need in decoder layers after $L/2$. We place four learned \texttt{[REL]} registers after the user prompt and before the assistant turn. Their causal attention can access both the image and the prompt without accessing the target answer. After the first $L/2$ layers of an $L$-layer LLM, let $\mathbf{h}^{\mathrm{rel}}_a$ denote register $a$'s hidden state and $\mathbf{h}_j$ the state of image token $j$. A bilinear head computes
\begin{equation}
    \begin{aligned}
        \mathbf{q}=W_q\!\left(
            \frac{1}{4}\sum_{a=1}^{4}
            \operatorname{LN}(\mathbf{h}^{\mathrm{rel}}_a)
        \right),
        s_j=\frac{
            \mathbf{q}^{\top}W_k\operatorname{LN}(\mathbf{h}_j)
        }{\sqrt{d_{\mathrm{rel}}}},
    \end{aligned}
    \label{eq:rel_scores}
\end{equation}
where $\operatorname{LN}$ is non-affine LayerNorm, $W_q$ and $W_k$
are learned projections, and $d_{\mathrm{rel}}=256$. For inference, we retain the $K_p$ highest-scoring image tokens using hard \texttt{top-k} and remove the rest together with the \texttt{[REL]} registers. We implement pruning as binary attention masking with a straight-through sigmoid surrogate~\cite{bengio2013estimating} during training, allowing gradients to flow through and train the selector in the joint adaptation.

\subsection{Optional Adaptive Halting}
\label{sec:adaptive_halting_main}

Wang et al.~\cite{wang2026horizon} show that VLMs have an \emph{information horizon}: a layer beyond which image tokens contribute little to the output. HiDrop~\cite{wu2026hidrop} exploits this horizon by removing all image tokens at a fixed layer, which is layer 25 for LLaVA-1.5-7B. Yet the horizon varies across tasks. Appendix~\ref{sec:supp_information_horizon} shows that text-heavy benchmarks have the latest horizon among benchmarks, later than layer 25; therefore, a fixed cutoff can retain unnecessary image tokens for some inputs while removing critical evidence too early for others. We thus introduce adaptive halting, an optional, training-free, plug-and-play mechanism that decides when to drop all remaining image tokens for each input. We further discuss its design, algorithm, and implementation in Appendix~\ref{sec:supp_adaptive_halting_implementation}.

\begin{figure}
    \centering
    \includegraphics[width=1\linewidth]{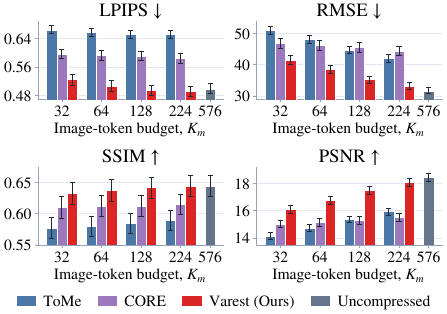}
    \caption{\textbf{Approximated fungibility at different merging-stage budgets.} We simulate merging by replacing each region with its mean colour on the raw image pixels and approximate fungibility by reconstruction quality of the merged view. Error bars are 95\% CIs. \method{} achieves the best quality across all token budgets and approaches the uncompressed quality at the highest budget.}
    \label{fig:reconstruction_quality}
\end{figure}

\subsection{Training and Budget Control}

\paragraph{Joint adaptation.}
We optimize the \texttt{[REL]} registers, scoring head, multimodal projector, position tags, and LoRA~\cite{hu2022lora} adapters in the vision and language stacks and vocabulary head using teacher-forced answer cross-entropy loss and AdamW~\cite{loshchilov2019adamw} on the VLM's own tuning dataset, so the VLM sees no additional data. Qwen3.8-27B uses mix665k~\cite{liu2024llava15} due to the lack of corresponding data. The localizers remain frozen, and no token-level selection labels are required. A straight-through sigmoid surrogate supplies gradients to the selector~\cite{bengio2013estimating}. We vary the merging budget during training and sample the pruning count from zero through the realized merging count. See Appendix~\ref{sec:supp_training} for more details.

\paragraph{Budget accounting.}
We measure the layer-averaged image-token budget $K=\frac{1}{L}\sum_{\ell=1}^{L}K_\ell=\frac{K_m+K_p}{2}$, where the second equality holds for midpoint pruning without adaptive halting, excluding text and \texttt{[REL]} tokens. For LLaVA-1.5, the retained ratio is $r=K/576$. For LLaVA-NeXT and Qwen3.8, we apply the requested ratio to each image's native sequence length, so actual counts vary with input resolution.
\section{Related Work}
\label{sec:related_work}

\begin{figure}
    \centering
    \includegraphics[width=1\linewidth]{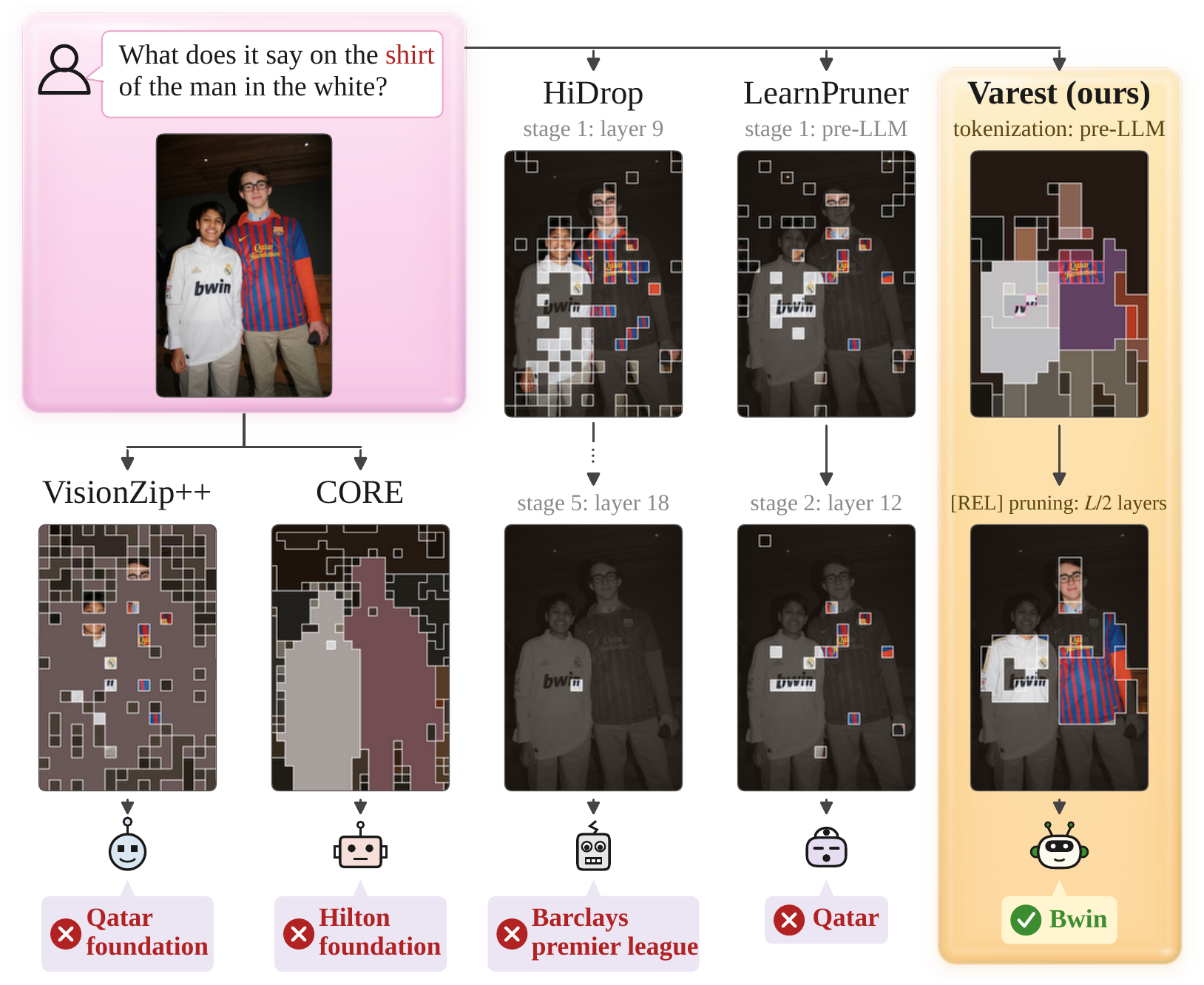}
    \caption{\textbf{Visualized comparison of the baselines and \method{} on LLaVA-1.5-7B.} HiDrop and LearnPruner retain the named text patches but lose contextual semantics, while CORE's coarse regions discard text and waste many tokens on the background. \method{} keeps text cells at native resolution and preserves more fine-grained details than CORE at the same token budget $K=32$.}
    \label{fig:baseline_visualization}
\end{figure}

\begin{figure*}[t]
    \centering
    \includegraphics[width=1\linewidth]{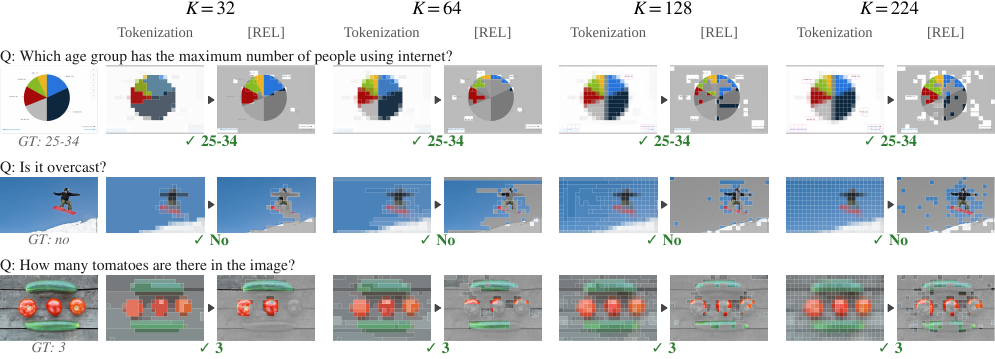}
    \caption{\textbf{Visualization of \method{} across budgets on LLaVA-1.5-7B.} The left panel shows the tokenization and the right panel shows \texttt{[REL]} necessity pruning. Across budgets, key text remains identified and protected, and semantic granularity becomes finer as $K$ increases.}
    \label{fig:qualitative_analysis}
\end{figure*}

\begin{figure}
    \centering
    \includegraphics[width=1\linewidth]{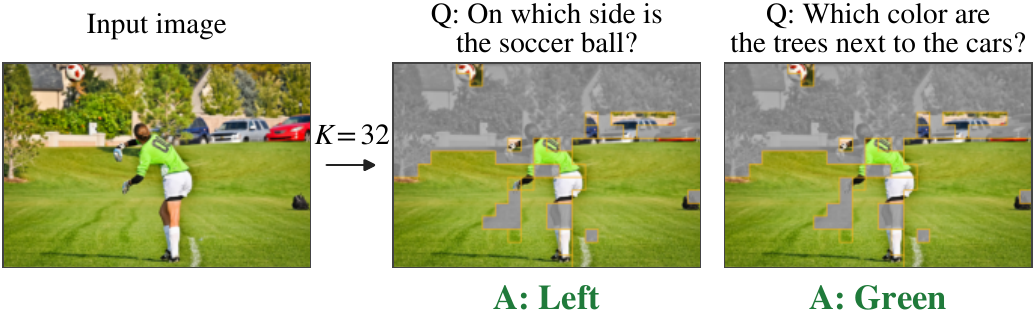}
    \caption{\textbf{Shared visual tokens support different questions.} On LLaVA-1.5-7B, \texttt{[REL]} keeps the same image tokens for two unrelated questions, and the VLM answers both correctly. The named objects are not necessarily needed after early decoder processing. The mechanism behind this finding is investigated in Section~\ref{sec:ablation}.}
    \label{fig:prompt_sensitivity}
\end{figure}

\paragraph{Compression signals.}
We categorize compression signals into semantic~\cite{lei2026core}, saliency~\cite{bolya2023tome,yang2025visionzip,alvar2025divprune,yang2025vflowopt,chen2026otprune,ma2026apet}, and query cues~\cite{chen2024fastv,zhang2025sparsevlm,xing2025pyramiddrop,huang2026nuwa,singh2026duet,takezoe2026learnpruner,wu2026hidrop}. Saliency cues estimate importance or redundancy from the image alone. ToMe~\cite{bolya2023tome} merges similar features, while VisionZip~\cite{yang2025visionzip} retains attention-ranked tokens and aggregates the remaining context. These image-only criteria permit compression before the prompt is processed, but do not explicitly account for semantics. Query cues use language context to prioritize content needed for a particular question, as in SparseVLM~\cite{zhang2025sparsevlm}. This directs the budget toward task-relevant evidence. However, when derived from decoder attention, these cues rely on decoder processing, and attention reliability can vary across layers~\cite{takezoe2026learnpruner}. \method{} combines within-image and depth-wise budget allocations, using variable-resolution tokenization and learned relative necessity pruning, guided by semantic cues that specify which features can be summarized together.

\paragraph{Allocating the token budget according to semantics.}
Within an image, semantic priors guide nonuniform token-budget distribution across regions. CORE~\cite{lei2026core} guides pooling with object masks and orders the resulting tokens by their centroids, preserving object membership and spatial layout. Limitations of CORE's object-based pooling include erasing text details within an object and an inability to control the granularity. \method{}'s variable-resolution tokenizer addresses this through maximizing semantic fungibility, using boundary and textness cues to allocate resolution. It separates detected text from its surrounding object, allowing coarse object summaries alongside finer text. Detected text patches remain at native resolution when the budget permits.

\paragraph{Allocating the token budget across decoder depth.}
Across decoder depth, the complementary question is how long a visual token should remain available. Multi-stage methods~\cite{xing2025pyramiddrop,singh2026duet,takezoe2026learnpruner} use larger budgets early and fewer tokens later, allowing contextual processing before pruning. DUET-VLM~\cite{singh2026duet} and LearnPruner~\cite{takezoe2026learnpruner} combine this with pre-LLM reduction. \method{} follows this depth-wise allocation overall but differs in selection. LearnPruner learns early patch selection but uses text-to-image attention to prune visual tokens in a middle decoder layer. Attention does not necessarily reflect a token’s influence on the prediction. In NLP models, Jain and Wallace~\cite{jain2019attention} find that attention weights can correlate poorly with gradient-based importance and that different attention distributions can yield similar predictions. Rather than using attention weights as an importance proxy, \method{} learns a relative necessity score for each semantic region through \texttt{[REL]} registers and a scoring head. This score naturally reflects how much the LLM still needs a token.
\section{Experiments}

We first visually inspect \method{}'s qualitative behaviours in Section~\ref{sec:qualitative_analysis}. Across three LLaVA backbones, Section~\ref{sec:quantitative_analysis} evaluates \method{} against 15 baselines on MMBench~\cite{liu2024mmbench}, AI2D~\cite{kembhavi2016ai2d}, MMStar~\cite{chen2024mmstar}, POPE~\cite{li2023pope} (three metrics), GQA~\cite{hudson2019gqa}, ChartQA~\cite{masry2022chartqa}, DocVQA~\cite{mathew2021docvqa}, TextVQA~\cite{singh2019textvqa}, RefCOCO/RefCOCO+~\cite{yu2016refcoco}, RefCOCOg~\cite{mao2016refcocog}, CountBench~\cite{paiss2023countbench}, and V*~\cite{wu2024vstar}. In addition, on Qwen3.8-27B~\cite{qwen38}, we compare with VisionZip++~\cite{yang2025visionzip} and LearnPruner~\cite{takezoe2026learnpruner}, which are both trained on the same external dataset, so the comparison is fair. We omit DUET-VLM~\cite{singh2026duet}, HiDrop~\cite{wu2026hidrop}, and CORE~\cite{lei2026core} because reproducing their two-stage training requires data that are unavailable. Adaptive halting is disabled except in Sections~\ref{sec:adaptive_halting_results} and \ref{sec:efficiency}, the latter of which analyzes the efficiency of \method{} on all four backbones; Section~\ref{sec:ablation} isolates the contributions of each component.

\begin{figure*}
    \centering
    \includegraphics[width=1\linewidth]{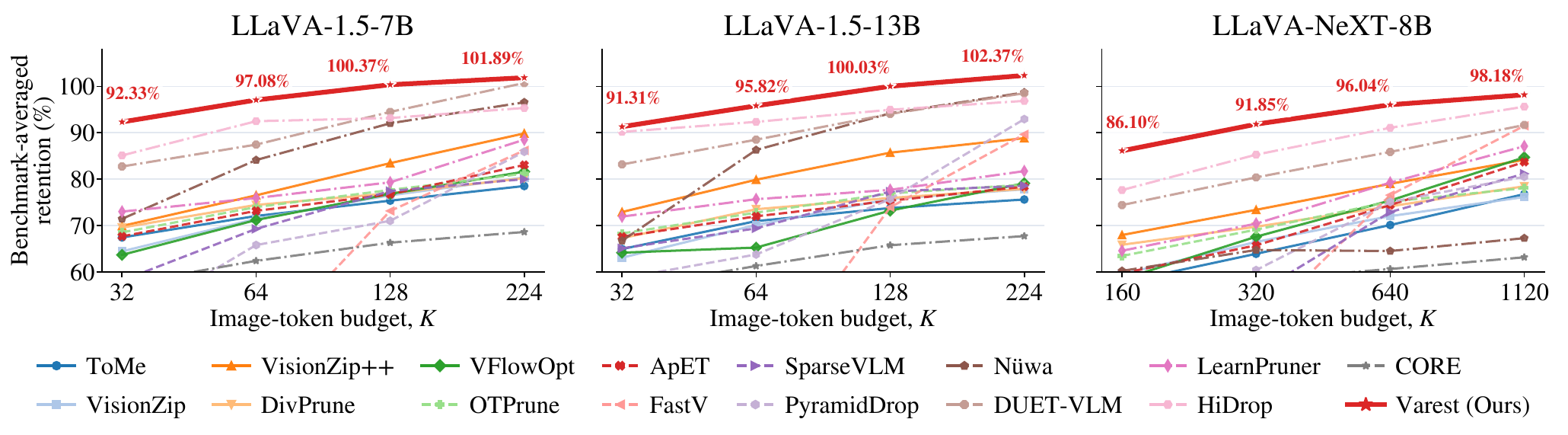}
    \caption{\textbf{Benchmark-averaged retention rates.} 100\% denotes uncompressed-backbone performance. \method~achieves the highest performance retention at every evaluated budget across all three LLaVA backbones, preserving over 85\% performance at $r=5.56\%$.}
    \label{fig:retention}
\end{figure*}

\begin{table*}
    \centering
    \caption{\textbf{Benchmark-averaged retention rates.} We evaluate the average retention rate with respect to the uncompressed model of each method on 13 benchmarks and 15 metrics (3 POPE metrics) across VQA, OCR, and grounding~\cite{liu2024mmbench,kembhavi2016ai2d,chen2024mmstar,li2023pope,hudson2019gqa,masry2022chartqa,mathew2021docvqa,singh2019textvqa,yu2016refcoco,mao2016refcocog,paiss2023countbench,wu2024vstar}. \method~outperforms all 15 baselines, of which the ones with * are also trained, at every budget across all three LLaVA backbones.}
    \resizebox{\linewidth}{!}{%
        \begin{tabular}{l|cccccccccccccccc}
            \toprule
            & \multirow{2}{*}{ToMe} & \multirow{2}{*}{VisionZip} & VisionZip & \multirow{2}{*}{DivPrune} & \multirow{2}{*}{VFlowOpt} & \multirow{2}{*}{OTPrune} & \multirow{2}{*}{ApET} & \multirow{2}{*}{FastV} & Sparse & Pyramid & \multirow{2}{*}{Nüwa} & DUET- & Learn & \multirow{2}{*}{HiDrop*} & \multirow{2}{*}{CORE*} & \multirow{2}{*}{\textbf{\method}*}\\
            & & & ++* & & & & & & VLM & Drop & & VLM* & Pruner*\\
            $K$ & \scriptsize\cite{bolya2023tome}[ICLR'23] & \scriptsize\cite{yang2025visionzip}[CVPR'25] & \scriptsize\cite{yang2025visionzip}[CVPR'25] & \scriptsize\cite{alvar2025divprune}[CVPR'25] & \scriptsize\cite{yang2025vflowopt}[ICCV'25] & \scriptsize\cite{chen2026otprune}[CVPRF'26] & \scriptsize\cite{ma2026apet}[CVPR'26] & \scriptsize\cite{chen2024fastv}[ECCV'24] & \scriptsize\cite{zhang2025sparsevlm}[ICML'25] & \scriptsize\cite{xing2025pyramiddrop}[CVPR'25] & \scriptsize\cite{huang2026nuwa}[ICLR'26] & \scriptsize\cite{singh2026duet}[CVPR'26] & \scriptsize\cite{takezoe2026learnpruner}[ICLR'26] & \scriptsize\cite{wu2026hidrop}[ICLR'26] & \scriptsize\cite{lei2026core}[CVPR'26] & \textbf{(Ours)}\\

            \midrule
            \rowcolor{gray!12}\multicolumn{17}{c}{\textit{LLaVA-1.5-7B (up to 576 image tokens)}}\\
            \midrule
            224 & 78.50\% & 80.36\% & 89.86\% & 80.19\% & 81.60\% & 81.17\% & 83.05\% & 86.19\% & 79.99\% & 85.90\% & 96.58\% & \colorbox{lightgreen}{100.74\%} & 88.51\% & 95.36\% & 68.58\% & \colorbox{green}{\textbf{101.89\%}}\\
            128 & 75.34\% & 76.49\% & 83.44\% & 76.93\% & 76.80\% & 77.65\% & 76.72\% & 73.12\% & 77.42\% & 70.99\% & 92.04\% & \colorbox{lightgreen}{94.50\%} & 79.31\% & 93.18\% & 66.27\% & \colorbox{green}{\textbf{100.37\%}}\\
            64 & 72.03\% & 71.41\% & 76.49\% & 74.42\% & 71.16\% & 73.95\% & 73.14\% & 42.39\% & 69.25\% & 65.76\% & 84.11\% & 87.44\% & 75.91\% & \colorbox{lightgreen}{92.49\%} & 62.35\% & \colorbox{green}{\textbf{97.08\%}}\\
            32 & 67.40\% & 64.40\% & 69.87\% & 69.53\% & 63.64\% & 68.49\% & 67.63\% & 39.67\% & 58.11\% & 48.74\% & 71.41\% & 82.72\% & 72.98\% & \colorbox{lightgreen}{85.10\%} & 57.22\% & \colorbox{green}{\textbf{92.33\%}}\\
            \midrule
            \rowcolor{gray!12}\multicolumn{17}{c}{\textit{LLaVA-1.5-13B (up to 576 image tokens)}}\\
            \midrule
            224 & 75.59\% & 77.79\% & 88.83\% & 77.88\% & 79.10\% & 78.80\% & 78.36\% & 89.65\% & 78.57\% & 92.94\% & \colorbox{lightgreen}{98.70\%} & 98.53\% & 81.72\% & 96.88\% & 67.69\% & \colorbox{green}{\textbf{102.37\%}}\\
            128 & 73.75\% & 75.72\% & 85.70\% & 75.93\% & 73.19\% & 76.90\% & 75.32\% & 73.90\% & 77.31\% & 75.67\% & 94.19\% & 94.06\% & 77.68\% & \colorbox{lightgreen}{94.96\%} & 65.72\% & \colorbox{green}{\textbf{100.03\%}}\\
            64 & 70.93\% & 70.09\% & 79.86\% & 73.49\% & 65.21\% & 72.75\% & 71.96\% & 34.39\% & 69.36\% & 63.72\% & 86.28\% & 88.50\% & 75.66\% & \colorbox{lightgreen}{92.35\%} & 61.24\% & \colorbox{green}{\textbf{95.82\%}}\\
            32 & 64.90\% & 63.03\% & 72.89\% & 67.24\% & 64.11\% & 68.27\% & 67.57\% & 37.29\% & 65.05\% & 58.57\% & 66.70\% & 83.16\% & 71.90\% & \colorbox{lightgreen}{90.19\%} & 56.29\% & \colorbox{green}{\textbf{91.31\%}}\\
            \midrule
            \rowcolor{gray!12}\multicolumn{17}{c}{\textit{LLaVA-NeXT-8B (up to 2880 image tokens)}}\\
            \midrule
            1120 & 76.75\% & 76.13\% & 84.16\% & 78.44\% & 84.69\% & 78.06\% & 83.68\% & 91.50\% & 81.16\% & 80.50\% & 67.24\% & 91.67\% & 87.10\% & \colorbox{lightgreen}{95.62\%} & 63.12\% & \colorbox{green}{\textbf{98.18\%}}\\
            640 & 70.08\% & 72.00\% & 78.98\% & 74.13\% & 75.37\% & 75.13\% & 74.40\% & 76.53\% & 72.87\% & 75.05\% & 64.47\% & 85.86\% & 79.24\% & \colorbox{lightgreen}{91.07\%} & 60.59\% & \colorbox{green}{\textbf{96.04\%}}\\
            320 & 63.87\% & 66.43\% & 73.36\% & 69.71\% & 67.56\% & 69.08\% & 65.70\% & 48.34\% & 55.30\% & 60.35\% & 64.69\% & 80.36\% & 70.38\% & \colorbox{lightgreen}{85.27\%} & 58.20\% & \colorbox{green}{\textbf{91.85\%}}\\
            160 & 57.66\% & 59.46\% & 67.92\% & 65.82\% & 58.35\% & 63.41\% & 59.64\% & 33.65\% & 58.45\% & 46.71\% & 60.18\% & 74.39\% & 64.53\% & \colorbox{lightgreen}{77.63\%} & 55.19\% & \colorbox{green}{\textbf{86.10\%}}\\
            \bottomrule
        \end{tabular}%
    }
    \label{tab:average_retention_rates}
\end{table*}

\subsection{Qualitative Analysis}
\label{sec:qualitative_analysis}

In Figure~\ref{fig:baseline_visualization}, the baselines lose the shirt text or the wearer's context, while \method{} preserves both and answers correctly.

Figure~\ref{fig:qualitative_analysis} shows how the \method{} tokenizer allocates resolution across budgets, merging the most fungible regions first. For instance, \method{} saves tokens by grouping the fungible wooden background into fewer, larger regions as the budget shrinks, while each tomato and cucumber stays within its boundaries. As a result, even at $K=32$, the diverse images in Figure~\ref{fig:teaser} remain recognizable after tokenization.

In Figure~\ref{fig:qualitative_analysis}, \texttt{[REL]} prunes the rightmost tomato at the lowest budget, yet the tomato count stays correct. In Figure~\ref{fig:prompt_sensitivity}, \texttt{[REL]} prunes the queried trees and keeps the same tokens for two unrelated questions, yet both answers are correct. Pruning after the first $L/2$ layers can thus drop even queried content, as the kept tokens suffice to form nearly the same hidden states as without necessity pruning (see Section~\ref{sec:ablation}).

\subsection{Quantitative Analysis}
\label{sec:quantitative_analysis}

\begin{table}[t]
    \centering
    \caption{\textbf{Benchmark-averaged retention rates on Qwen3.8-27B.} \method{} outperforms both trained (*) baselines at every token budget.}
    \resizebox{\linewidth}{!}{%
        \begin{tabular}{lrrrr}
            \toprule
            Method & $r=38.89\%$ & 22.22\% & 11.11\% & 5.56\%\\
            \midrule
            VisionZip++*~\cite{yang2025visionzip} & \colorbox{lightgreen}{95.76\%} & \colorbox{lightgreen}{88.50\%} & \colorbox{lightgreen}{77.34\%} & \colorbox{lightgreen}{67.61\%}\\
            LearnPruner*~\cite{takezoe2026learnpruner} & 79.43\% & 71.74\% & 64.16\% & 53.57\%\\
            \textbf{Varest* (Ours)} & \colorbox{green}{\textbf{99.40\%}} & \colorbox{green}{\textbf{96.66\%}} & \colorbox{green}{\textbf{91.17\%}} & \colorbox{green}{\textbf{84.04\%}}\\
            \bottomrule
        \end{tabular}%
    }
    \label{tab:average_retention_rates_qwen}
\end{table}

\begin{figure*}
    \centering
    \includegraphics[width=1\linewidth]{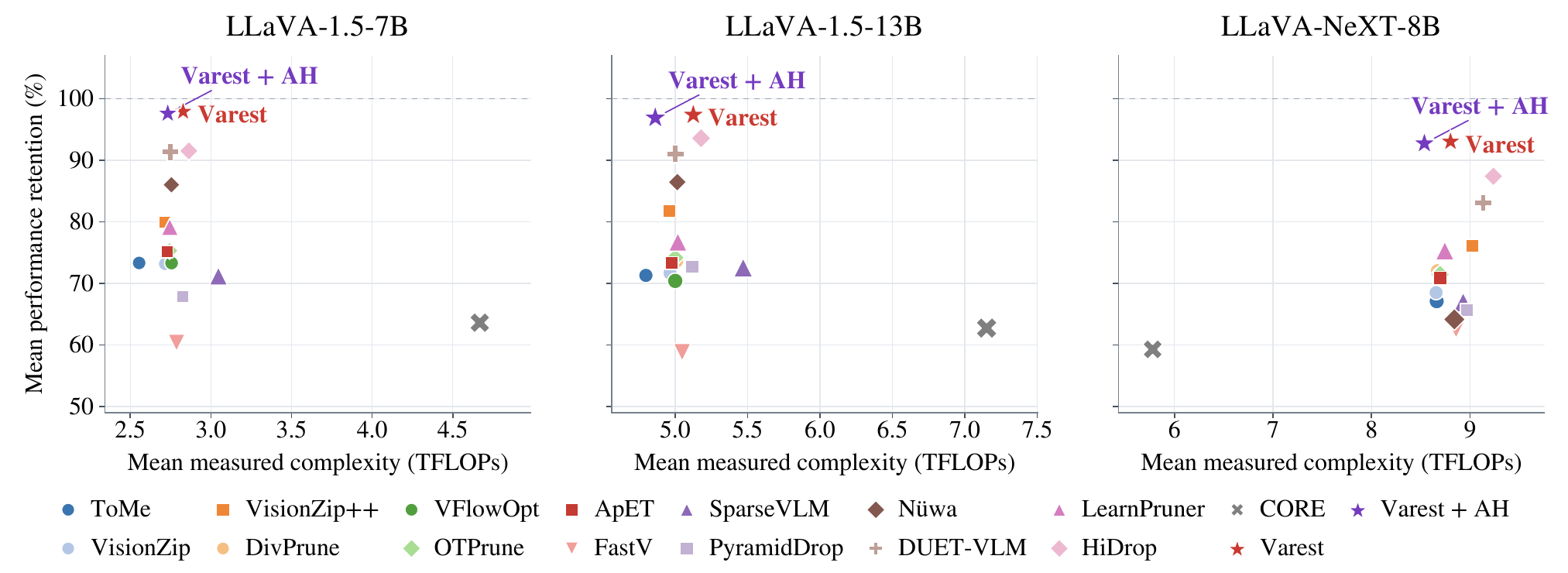}
    \caption{\textbf{Performance retention versus measured inference FLOPs.} Each point averages both axes independently over four budgets; size corresponds to latency; smaller, higher, and further left is better. \method~achieves higher retention with slightly less computation than HiDrop, the strongest baseline by mean retention. Adaptive halting (AH) further reduces computation with a small retention loss.}
    \label{fig:efficiency}
\end{figure*}

We first examine whether \method's tokenization achieves the intended fungibility. Figure~\ref{fig:reconstruction_quality} shows the approximated fungibility of ToMe, CORE, and our method, through four reconstruction-quality metrics comparing the merged view (unlike the visualizations in the figures, text patches are not protected as raw pixels for fairness) with the original image, on 260 images (20 from each benchmark). \method{} consistently achieves higher fungibility than the two merging-based baselines and approaches the intrinsic uncompressed quality (bounded by the $24\times24$ resolution) as the budget increases.

\method{} achieves the highest mean performance retention at every evaluated budget across all four backbones, as shown in Tables~\ref{tab:average_retention_rates}-\ref{tab:average_retention_rates_qwen} and Figure~\ref{fig:retention}. Notably, at an 11.11\% token-retention ratio, \method{} outperforms every baseline evaluated at 22.22\% on all four backbones. At larger budgets, mean retention exceeds 100\% on both LLaVA-1.5 backbones, showing that \method{} can also improve upon the uncompressed model. By grouping fungible tokens into semantic regions, \method{} acts as a VLM \emph{catalyst}, providing the decoder with key information while lowering token budgets. Per-benchmark results across the 15 metrics are discussed in Appendix~\ref{sec:supp_quantitative_results}, where we find that \method{} achieves 190 first-place wins summed across 15 metrics and 4 backbones.

Figure~\ref{fig:radar} shows how performance responds to budgets varies across tasks. General VQA remains relatively stable as the budget decreases, whereas text- and chart-heavy and grounding benchmarks benefit more from larger budgets.

\subsection{Adaptive Halting}
\label{sec:adaptive_halting_results}

\begin{table}[t]
    \centering
    \caption{\textbf{Optional halting on three LLaVA backbones.} The score costs are relatively small compared to the additional savings.}
    \resizebox{\linewidth}{!}{%
        \begin{tabular}{lrrrr}
            \toprule
            Backbone & $\Delta$ Score & Trigger Rate
            & Skips \texttt{[REL]} & Tokens Saved \\
            \midrule
            LLaVA-1.5-7B
            & \cellcolor{retPink!52.86}$-0.10$
            & 73.0\% & 2.3\%
            & \cellcolor{retGreen!76.98}9.6\% \\
            LLaVA-1.5-13B
            & \cellcolor{retPink!100.00}$-0.21$
            & 84.6\% & 28.8\%
            & \cellcolor{retGreen!100.00}12.9\% \\
            LLaVA-NeXT-8B
            & \cellcolor{retPink!100.00}$-0.21$
            & 86.9\% & 9.2\%
            & \cellcolor{retGreen!70.00}8.6\% \\
            \bottomrule
        \end{tabular}%
    }
    \label{tab:adaptive_halting}
\end{table}

We enable adaptive halting using the same checkpoints and budgets, without additional training or backbone-specific tuning. Table~\ref{tab:adaptive_halting} reports the mean score changes, halting rates, and image-token savings across 13 benchmarks and four budgets. We observe an additional 8.6--12.9\% reduction in visual-token propagation, measured by image-token counts summed across decoder layers. Mean score losses are 0.10--0.21 points on the benchmarks' native 0--100 scales, averaged equally over 15 metrics and four budgets. Appendix~\ref{sec:supp_adaptive_halting} provides more detailed results about adaptive halting.

Halting predominantly follows \texttt{[REL]} necessity pruning on LLaVA-1.5-7B and LLaVA-NeXT-8B, withdrawing the retained tokens from later layers. On LLaVA-1.5-13B, 28.8\% of inputs halt before selection runs, including halts at the selection boundary, contributing to its larger savings.

\subsection{Efficiency Analysis}
\label{sec:efficiency}

Figure~\ref{fig:efficiency} compares performance retention and measured computation, each averaged over four budgets. Across the three LLaVA backbones, \method{} improves retention over HiDrop, the strongest baseline, by 3.79--6.39 percentage points while using 1.0--4.7\% fewer FLOPs and 1.27--1.66 times faster (without adaptive halting). Moreover, \method{} is 1.26--1.45 times faster with up to a 74.36\% reduction in FLOPs compared to the uncompressed Qwen3.8-27B. Compared to VisionZip++, \method{} has comparable measured FLOPs and latency but achieves much better performance retention rates in Table~\ref{tab:average_retention_rates_qwen}. Its accuracy advantage therefore persists when accounting for measured computation. Appendix~\ref{sec:supp_efficiency} includes detailed numerical measurements for all four backbones.

Adaptive halting further reduces FLOPs by 3.0--5.1\%, with retention losses of 0.30--0.48 percentage points. It provides a plug-and-play additional tradeoff between compute and accuracy. However, its frequent detection outweighs the compute savings, making it slightly slower than vanilla \method{} at batch size 1, unless there are many more tokens.

\subsection{Ablation Studies}
\label{sec:ablation}

\begin{table*}[t]
    \centering
    \caption{\textbf{Effect of removing the text localizer on selected benchmarks.} Entries are changes in native 0--100 scores relative to full \method{} at each token budget, with textness set to zero. The absence of the text localizer damages TextVQA most consistently on all three backbones and DocVQA on the two LLaVA-1.5 models. Its effect is small and mixed when the input image is mostly text.}
    \resizebox{1\linewidth}{!}{%
        \begin{tabular}{lrrrr@{\qquad}lrrrr@{\qquad}lrrrr}
            \toprule
            \multicolumn{5}{c}{LLaVA-1.5-7B (up to 576 image tokens)}
            & \multicolumn{5}{c}{LLaVA-1.5-13B (up to 576 image tokens)}
            & \multicolumn{5}{c}{LLaVA-NeXT-8B (up to 2880 image tokens)} \\
            \cmidrule(lr){1-5}\cmidrule(lr){6-10}\cmidrule(lr){11-15}
            $K$ & TextVQA & DocVQA & ChartQA & CountBench
            & $K$ & TextVQA & DocVQA & ChartQA & CountBench
            & $K$ & TextVQA & DocVQA & ChartQA & CountBench \\
            \midrule
            224 & \cellcolor{retPink!10.80}$-0.02$ & \cellcolor{retPink!19.24}$-0.23$ & \cellcolor{retPink!13.21}$-0.08$ & \cellcolor{retGreen!18.04}$+0.20$ & 224 & \cellcolor{retGreen!10.80}$+0.02$ & \cellcolor{retPink!17.23}$-0.18$ & \cellcolor{retGreen!21.25}$+0.28$ & $+0.00$ & 1120 & \cellcolor{retPink!20.45}$-0.26$ & \cellcolor{retPink!14.42}$-0.11$ & \cellcolor{retPink!40.54}$-0.76$ & \cellcolor{retPink!18.44}$-0.21$\\
            128 & \cellcolor{retPink!36.52}$-0.66$ & \cellcolor{retPink!41.34}$-0.78$ & \cellcolor{retGreen!11.61}$+0.04$ & \cellcolor{retPink!59.02}$-1.22$ & 128 & \cellcolor{retPink!16.83}$-0.17$ & \cellcolor{retPink!26.47}$-0.41$ & \cellcolor{retPink!11.61}$-0.04$ & \cellcolor{retPink!34.51}$-0.61$ & 640 & \cellcolor{retPink!52.99}$-1.07$ & \cellcolor{retGreen!13.21}$+0.08$ & \cellcolor{retGreen!48.57}$+0.96$ & \cellcolor{retPink!34.51}$-0.61$\\
            64 & \cellcolor{retPink!61.43}$-1.28$ & \cellcolor{retPink!38.93}$-0.72$ & \cellcolor{retPink!11.61}$-0.04$ & \cellcolor{retPink!91.56}$-2.03$ & 64 & \cellcolor{retPink!60.22}$-1.25$ & \cellcolor{retPink!49.78}$-0.99$ & \cellcolor{retGreen!11.61}$+0.04$ & \cellcolor{retGreen!67.46}$+1.43$ & 320 & \cellcolor{retPink!65.45}$-1.38$ & \cellcolor{retGreen!18.44}$+0.21$ & \cellcolor{retGreen!11.61}$+0.04$ & \cellcolor{retPink!34.51}$-0.61$\\
            32 & \cellcolor{retPink!65.85}$-1.39$ & \cellcolor{retPink!36.92}$-0.67$ & \cellcolor{retPink!11.61}$-0.04$ & \cellcolor{retPink!26.07}$-0.40$ & 32 & \cellcolor{retPink!58.62}$-1.21$ & \cellcolor{retPink!37.72}$-0.69$ & \cellcolor{retGreen!32.50}$+0.56$ & $+0.00$ & 160 & \cellcolor{retPink!100.00}$-2.24$ & \cellcolor{retGreen!42.54}$+0.81$ & \cellcolor{retGreen!74.29}$+1.60$ & \cellcolor{retGreen!26.47}$+0.41$\\
            \bottomrule
        \end{tabular}%
    }
    \label{tab:no_text_localizer}
\end{table*}

\begin{table}
    \centering
    \caption{\textbf{Retention of \method{} variants at matched budgets.} $^\dagger$ denotes an inference-time ablation of the joint model \method. Localization and necessity pruning provide complementary benefits.}
    \resizebox{1\linewidth}{!}{%
        \begin{tabular}{lrrrrrrr}
            \toprule
            \multirow{2}{*}{$K$} & \multirow{2}{*}{\method} & \multirow{2}{*}{-\texttt{[REL]}}
            & Random
            & Oracle
            & -Object
            & -Text
            & No \\
            & & & Selection$^\dagger$ & Selection$^\dagger$ & Localizer & Localizer & Localizer\\
            \midrule
            \rowcolor{gray!12}\multicolumn{8}{c}{\textit{LLaVA-1.5-7B (up to 576 image tokens)}}\\
            \midrule
            224 & 101.89\% & \cellcolor{retPink!10.57}101.80\% & \cellcolor{retPink!29.29}98.83\% & \cellcolor{retGreen!10.50}101.97\% & \cellcolor{retPink!13.34}101.36\% & \cellcolor{retPink!11.26}101.69\% & \cellcolor{retPink!13.40}101.35\%\\
            128 & 100.37\% & \cellcolor{retPink!14.79}99.61\% & \cellcolor{retPink!31.30}96.99\% & \cellcolor{retGreen!12.21}100.72\% & \cellcolor{retPink!18.82}98.97\% & \cellcolor{retPink!11.51}100.13\% & \cellcolor{retPink!23.61}98.21\%\\
            64 & 97.08\% & \cellcolor{retPink!20.27}95.45\% & \cellcolor{retPink!27.14}94.36\% & \cellcolor{retGreen!13.28}97.60\% & \cellcolor{retPink!33.76}93.31\% & \cellcolor{retPink!13.47}96.53\% & \cellcolor{retPink!37.79}92.67\%\\
            32 & 92.33\% & \cellcolor{retPink!29.85}89.18\% & \cellcolor{retPink!23.80}90.14\% & \cellcolor{retGreen!13.03}92.81\% & \cellcolor{retPink!49.77}86.02\% & \cellcolor{retPink!13.53}91.77\% & \cellcolor{retPink!52.73}85.55\%\\
            \midrule
            \rowcolor{gray!12}\multicolumn{8}{c}{\textit{LLaVA-1.5-13B (up to 576 image tokens)}}\\
            \midrule
            224 & 102.37\% & \cellcolor{retPink!20.84}100.65\% & \cellcolor{retPink!23.30}100.26\% & \cellcolor{retPink!12.46}101.98\% & \cellcolor{retPink!13.28}101.85\% & \cellcolor{retPink!13.09}101.88\% & \cellcolor{retPink!13.47}101.82\%\\
            128 & 100.03\% & \cellcolor{retPink!24.31}97.76\% & \cellcolor{retPink!19.08}98.59\% & \cellcolor{retPink!10.19}100.00\% & \cellcolor{retPink!13.97}99.40\% & \cellcolor{retPink!13.40}99.49\% & \cellcolor{retPink!17.06}98.91\%\\
            64 & 95.82\% & \cellcolor{retPink!21.03}94.07\% & \cellcolor{retPink!19.14}94.37\% & \cellcolor{retGreen!10.25}95.86\% & \cellcolor{retPink!24.37}93.54\% & \cellcolor{retPink!12.02}95.50\% & \cellcolor{retPink!29.22}92.77\%\\
            32 & 91.31\% & \cellcolor{retPink!25.19}88.90\% & \cellcolor{retPink!15.42}90.45\% & \cellcolor{retGreen!12.08}91.64\% & \cellcolor{retPink!46.30}85.55\% & \cellcolor{retGreen!10.38}91.37\% & \cellcolor{retPink!47.75}85.32\%\\
            \midrule
            \rowcolor{gray!12}\multicolumn{8}{c}{\textit{LLaVA-NeXT-8B (up to 2880 image tokens)}}\\
            \midrule
            1120 & 98.18\% & \cellcolor{retPink!15.92}97.24\% & \cellcolor{retPink!36.97}93.90\% & \cellcolor{retGreen!10.25}98.22\% & \cellcolor{retPink!10.95}98.03\% & \cellcolor{retPink!11.45}97.95\% & \cellcolor{retPink!13.03}97.70\%\\
            640 & 96.04\% & \cellcolor{retPink!21.09}94.28\% & \cellcolor{retPink!38.11}91.58\% & \cellcolor{retPink!10.95}95.89\% & \cellcolor{retPink!13.47}95.49\% & \cellcolor{retPink!12.27}95.68\% & \cellcolor{retPink!18.19}94.74\%\\
            320 & 91.85\% & \cellcolor{retPink!23.80}89.66\% & \cellcolor{retPink!35.21}87.85\% & \cellcolor{retGreen!12.90}92.31\% & \cellcolor{retPink!42.02}86.77\% & \cellcolor{retPink!11.89}91.55\% & \cellcolor{retPink!53.11}85.01\%\\
            160 & 86.10\% & \cellcolor{retPink!30.55}82.84\% & \cellcolor{retPink!31.24}82.73\% & \cellcolor{retGreen!14.47}86.81\% & \cellcolor{retPink!85.57}74.11\% & \cellcolor{retPink!10.63}86.00\% & \cellcolor{retPink!100.00}71.82\%\\
            \bottomrule
        \end{tabular}%
    }
    \label{tab:ablation}
\end{table}

Table~\ref{tab:ablation} compares variants at matched layer-averaged token budgets. Ablations are retrained unless denoted with $^\dagger$. $-$\texttt{[REL]} is a tokenizer-only arm, where $K_m=K$ and $r_m=r$. Random and Oracle Selections replace \texttt{[REL]} with random selection and gold-answer attention, respectively, keeping the joint model's pruning layer and token counts. $-$Object Localizer reorders the original boundary strengths according to adjacent CLIP-feature~\cite{radford2021clip} distances; $-$Text Localizer sets all textness values to zero; and No Localizer combines both ablations. Shuffling ablations are in Appendix~\ref{sec:supp_spatial_information}. To connect task performance with the resulting partitions, Table~\ref{tab:partition_preservation} evaluates structural preservation at $K_m=64$. Region IoU measures object-mask precision over 3,811 referred objects from RefCOCO. Native text measures the fraction of individual OCR-matched answer cells retained, averaged over 2,688 TextVQA questions with available answer cells. More ablation studies are in Appendix~\ref{sec:supp_ablation_studies}.

\paragraph{Object Localizer.}
Replacing the learned boundary ordering reduces Region IoU from 68.43\% to 45.01\% (Table~\ref{tab:partition_preservation}) and lowers accuracy (Table~\ref{tab:ablation}). The widening performance gaps as token budgets shrink show the importance of benchmarking on small budgets, where the impacts of differences in compression quality are amplified. Under such tight compression, even the object localizer's accuracy becomes particularly important for preserving downstream performance.

\paragraph{Text localization.}
Object alignment alone does not ensure text preservation. Removing the text localizer reduces unmerged text cells from 43.2\% to 17.1\% (Table~\ref{tab:partition_preservation}). The impact of the text localizer is clearest in TextVQA, where it increases accuracy in 11 of 12 budgets across the three LLaVA backbones (Table~\ref{tab:no_text_localizer}). The limited effect on DocVQA and ChartQA is expected: the text localizer prevents text cells from merging with the objects, but on text-only images, where the number of text patches exceeds the budget, it needs to merge the text patches without the ability to rank them by importance. Appendix~\ref{sec:supp_budget_dependent_text_threshold} also isolates the contribution of the budget-dependent text threshold. Overall, the two localizers support complementary aspects of tokenization.

\paragraph{\texttt{[REL]} necessity pruning.}
The joint model exceeds the separately trained $-$\texttt{[REL]} variant at every budget, gaining 2.41--3.26 retention points at the smallest budgets. This comparison evaluates the joint design, including training and budget allocation across decoder depth. To examine the significance of selection quality within this design, we hold the pruning location and token counts fixed, and our learned selection outperforms random selection at every budget on all three LLaVA backbones by 0.86--4.46 percentage points.

We find that learned selection is largely stable across prompts (Figure~\ref{fig:prompt_sensitivity}): a shared subset of visual tokens supports different questions. The gain over random selection shows that prompt insensitivity coexists with successfully learning which tokens the LLM needs. Meanwhile, Oracle Selection remains within 0.71 retention points of \method{} at every budget, suggesting limited headroom for greater prompt sensitivity at the evaluated pruning layers. Appendix~\ref{sec:supp_prompt_sensitivity} explains this counterintuitive finding by revealing the discrepancy between when prompt awareness becomes available and when it remains beneficial. Appendix~\ref{sec:supp_rel_hidden_state_preservation} then further analyzes why the advantage becomes small. We find that at the decoder midpoint, the tokens kept by \texttt{[REL]} are the ones that allow the following layers to construct the same hidden states as without pruning, using a different basis. The linear combination coefficients, known as the attention weights, describe how tokens are combined rather than directly measuring whether a token is necessary to form the same answer.

\begin{table}
    \centering
    \caption{\textbf{Structural preservation on LLaVA-1.5 grid.} At $K_m=64$, removing the text localizer slightly improves region IoU but destroys native text capture, showing its necessity.}
    \resizebox{\linewidth}{!}{%
        \begin{tabular}{lrrrr}
            \toprule
            & Full tokenizer & $-$Object Localizer & $-$Text Localizer & No Localizer \\
            \midrule
            Region IoU $\uparrow$
            & 68.43\%
            & \cellcolor{retPink!69.71}45.01\%
            & \cellcolor{retGreen!12.07}69.24\%
            & \cellcolor{retPink!67.90}45.72\%\\
            Native text $\uparrow$
            & 43.2\%
            & \cellcolor{retPink!23.77}37.8\%
            & \cellcolor{retPink!76.54}17.1\%
            & \cellcolor{retPink!100.00}7.9\%\\
            \bottomrule
        \end{tabular}%
    }
    \label{tab:partition_preservation}
\end{table}
\section{Discussion}
\label{sec:discussion}

Aggressive visual-token compression makes it difficult to retain both global image context and fine visual details at a small token budget. \method{} addresses this challenge by grouping coherent content while preserving object distinctions and text details. It maintains full spatial coverage at a budget-controlled granularity, then applies relative necessity pruning after early decoder processing to retain visual tokens still needed by later layers. Moreover, \method{} remains compatible with FlashAttention~\cite{dao2022flashattention} and shortens the sequence, freeing the context window for longer prompts or more images. Natural extensions include ranking text importance to improve text-only benchmarks. More broadly, the semantic fungibility framework has transformational potential for compression beyond 2D natural images, such as videos, 3D volumes, and medical images, where an efficient use of visual tokens matters more for the limited context window. For example, we could model anatomical regions as more fungible than small lesions and tissue boundaries, offering a better alternative to uniformly downsampling the entire scan.

\clearpage

\section*{LLM Usage Statement}

\paragraph{Manuscript writing and visualization.}
The human authors wrote the entire manuscript manually. Grammarly was used to help with syntax and grammar corrections. Codex, Claude Code, and Gemini occasionally provided feedback on writing clarity and served as simulated reviewers, but no generated content was directly incorporated into the manuscript.

All figures in the main text were programmatically (as opposed to generatively) created by Codex and Claude Code with human guidance and necessary data provided, except Figure~\ref{fig:method}, which the authors designed manually in Affinity with the 3D tokenizer part modelled and rendered in Blender.

\paragraph{Technical implementation.}
We utilized Codex and Claude Code to assist with baseline implementations following a standardized protocol: (1) The authors manually read the reference paper, inspected the official codebase (where available), and designed a reimplementation plan; (2) the authors reimplemented the method with the help of coding agents within our evaluation framework; (3) the authors validated the reimplementation against the original reported results using the author-provided evaluation protocols; (4) upon matching original performance, the authors benchmarked the baseline under our evaluation setup; and (5) in cases of discrepancies, the authors conducted debugging and error analysis until the reimplementation was verified faithful. 

Additionally, Claude Code was employed during early-stage brainstorming to suggest exploratory hypotheses and experiments. All final concept formulations, experiment designs, and method refinements were independently proposed, directed, and validated entirely by the human authors.

\makeatletter
\@ifpackagewith{cvpr}{review}{}{%
\section*{Contribution and Acknowledgement}

TF led the project, proposed the \method{} method, designed core experiments, and oversaw all changes. TF, AW, and CC wrote the manuscript and created the primary visualizations. AW implemented and trained CORE, HiDrop, DUET-VLM, VFlowOpt, and SparseVLM. CC implemented and trained ToMe, LearnPruner, VisionZip/VisionZip++, and ApET. RA established the initial codebase with the full-token pipeline and implemented FastV. AW, CC, RA, and VX conducted and analyzed experiments for the method prototype during early development. VX managed experimental results and developed visualization utilities. At the end, RA and VX read through and checked the manuscript, flagged mistakes, and reported them to TF, AW, and CC for correction.

We thank Xinggang Wang, Jun Ma, Ruihang Zhang, Yuanye Liu, and Tianyi Xu for their advice and feedback on the manuscript; Philippe Tremblay for making the 3D tokenization view in Figure~\ref{fig:method} in Blender; Emile Petit for his help on the Information Horizon experiments (Appendix~\ref{sec:supp_information_horizon}); Xiyue Jiao and Yucheng Chen for assistance with baseline implementations; Martin Tin for contributions to selecting the samples for efficiency measurement. In addition, we thank Lambda.ai for sponsored cloud compute credits.
}
\makeatother

{
    \small
    \bibliographystyle{ieeenat_fullname}
    \bibliography{main}
}

\clearpage
\setcounter{page}{1}
\renewcommand{\thepage}{A\arabic{page}}
\maketitlesupplementary
\appendix
  
\begin{strip}
    \centering
    \captionsetup{type=figure}
    \includegraphics[width=1\linewidth]{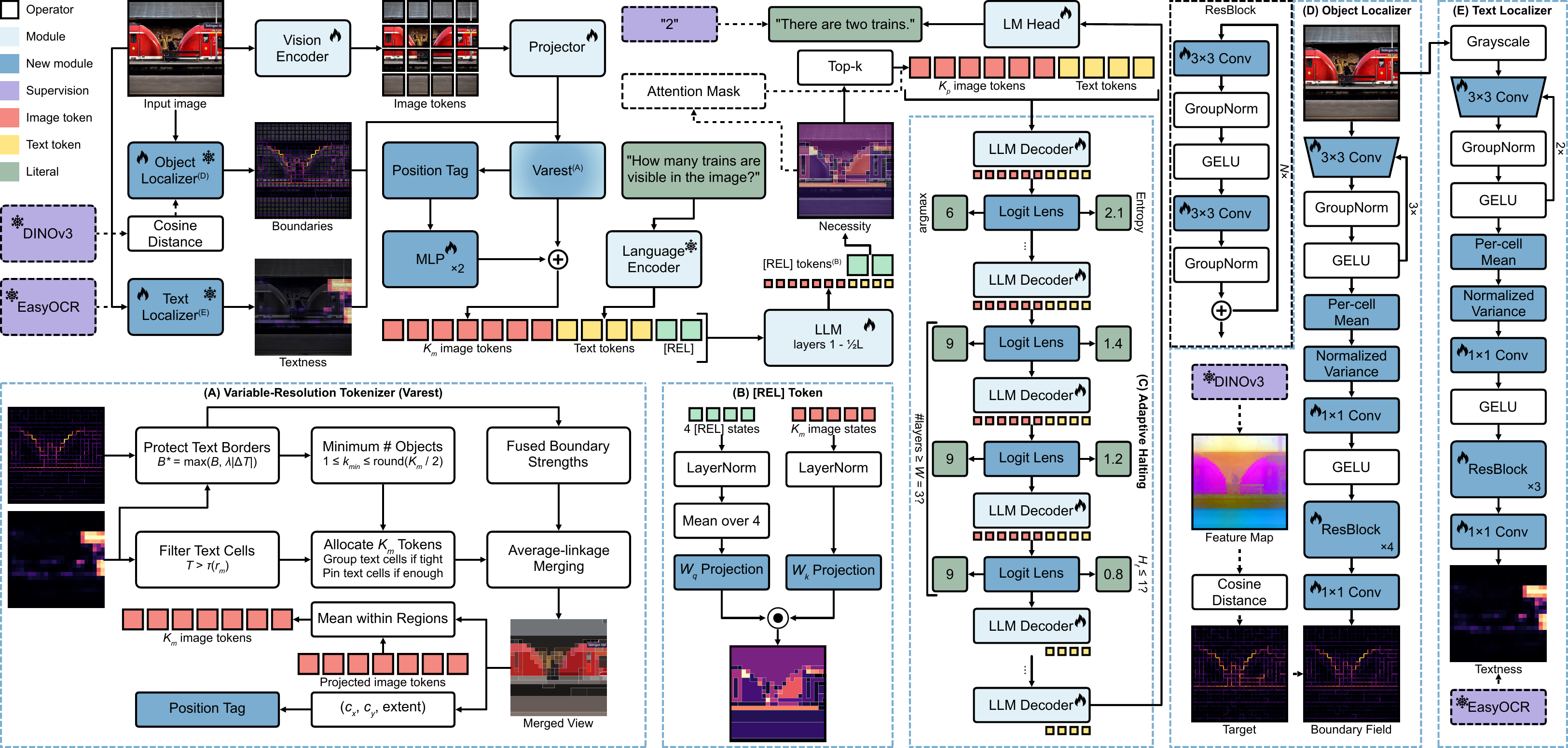}
    \caption{\textbf{Detailed architecture of \method.} (A) Object boundaries and textness guide the variable-resolution tokenization at a budget-controlled granularity that produces $K_m$ image tokens, each augmented with an additive learned position tag. (B) Four learned \texttt{[REL]} registers assign relative necessity scores at the decoder midpoint and retain the $K_p$ highest-scoring visual tokens. Training uses binary attention masking with a straight-through sigmoid surrogate (Qwen training compacts inputs to DeltaNet token mixers). Inference uses hard top-$K_p$ selection. (C) Optional adaptive halting removes all remaining visual tokens once the intermediate answer predictions become stable and confident. (D--E) Lightweight object and text localizers predict boundary strengths and textness from RGB and grayscale pixels, respectively. DINOv3 feature distances and EasyOCR boxes provide their training supervision. The foundational model supplies a semantic prior learned from a large amount of data, which the object localizer distills. We choose DINOv3 for its outstanding spatial alignment.}
    \label{fig:method_detailed}
\end{strip}

\section{Method and Training Details}
\label{sec:supp_method_training}

We describe the method architecture, components, and training design in detail. The architecture components and their positions are described in Section~\ref{sec:supp_method_details}. We then describe training protocols in Section~\ref{sec:supp_training}. Finally, Section~\ref{sec:supp_evaluation} documents baseline adjustments and evaluation protocols.

\subsection{Method}
\label{sec:supp_method_details}

\subsubsection{Localizer Architecture and Field Alignment}
\label{sec:localizer_details}

\begin{figure*}
    \centering
    \includegraphics[width=1\linewidth]{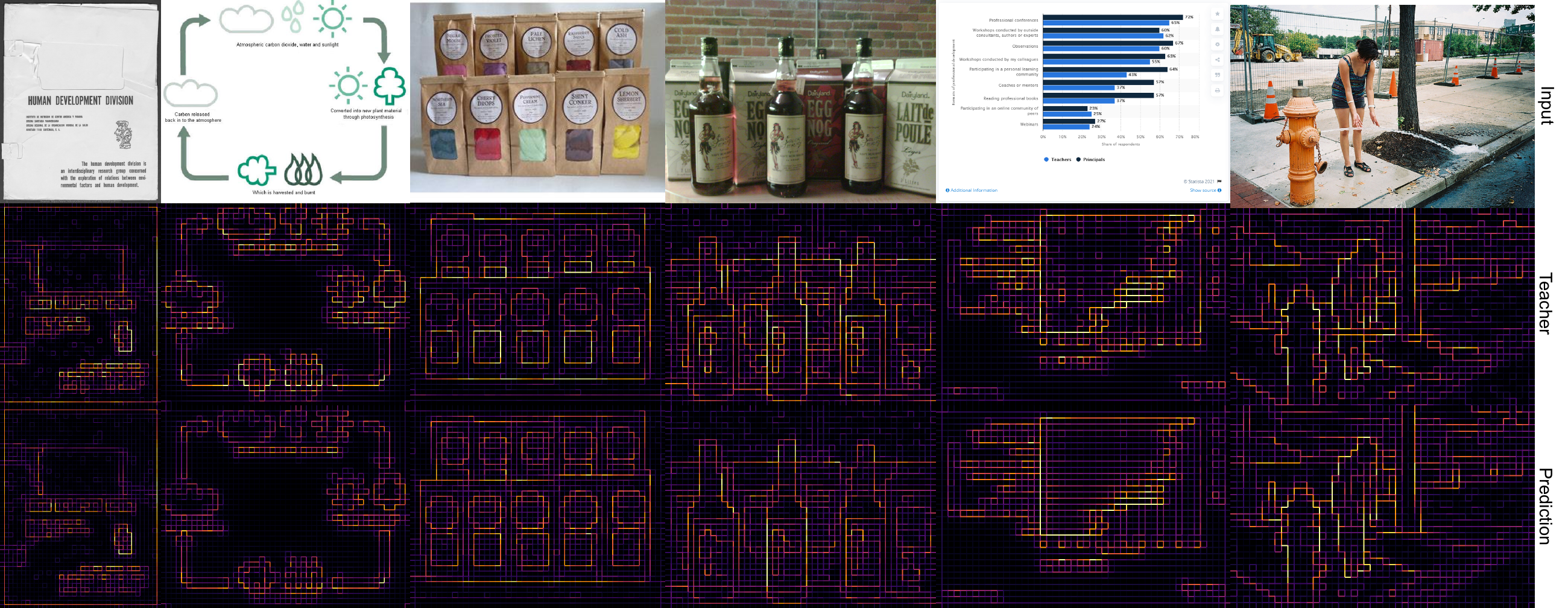}
    \caption{\textbf{Visualization of the object localizer.} Top: input images. Middle: teacher boundary strengths computed as cosine distances between neighbour DINOv3 features. Bottom: predictions from the lightweight RGB localizer. Brighter edges indicate stronger boundaries. The localizer captures the teacher's main boundary structure across documents, diagrams, product images, charts, and natural scenes, providing continuous evidence for region merging without running DINOv3 during inference.}
    \label{fig:object_localizer}
\end{figure*}

\begin{figure}
    \centering
    \includegraphics[width=1\linewidth]{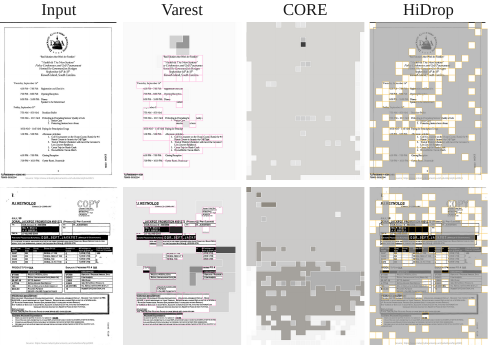}
    \caption{\textbf{\method{} applied to document images.} Compared with baselines, \method{} is able to consistently identify text, while wasting little budget on whitespace.}
    \label{fig:document_case}
\end{figure}

Figure~\ref{fig:method_detailed}(D--E) details the two localizers that both operate on pixels scaled to $[0,1]$, independently of the vision encoder, prompt, and token budget. DINOv3 and EasyOCR provide supervision targets only. After both localizers are separately trained, they remain frozen during adaptation and inference.

\paragraph{Shared architecture.}
Each stem layer consists of a bias-free $3\times3$ convolution with stride 2 and padding 1, followed by eight-group GroupNorm~\cite{wu2018groupnorm} and GELU~\cite{hendrycks2016gelu}. Within each cell, we compute the channel-wise feature mean $\boldsymbol{\mu}_i$ and population variance $\mathbf{v}_i$. The variance is standardized across cells of the same image, independently for each channel, and concatenated with the mean:
\begin{equation}
    \mathbf{u}_i=\left[\boldsymbol{\mu}_i,\frac{\mathbf{v}_i-\bar{\mathbf{v}}}{\max\{\operatorname{Std}_{j}(\mathbf{v}_j),10^{-9}\}}\right],
\end{equation}
where $\operatorname{Std}$ uses the sample standard deviation. A $1\times1$ projection and GELU precede the residual blocks~\cite{he2016resnet}. Each block applies a dilated $3\times3$ convolution~\cite{yu2016dilated}, GroupNorm, GELU, another convolution, and GroupNorm, then adds the input without a further activation. Padding equals the dilation. GroupNorm uses eight groups, affine parameters, and $\epsilon=10^{-5}$. Convolutions after the stem are biased, and no additional normalization precedes the $1\times1$ projection.

\paragraph{Object localizer.}
The RGB stem has three layers with channel widths $3\rightarrow96\rightarrow192\rightarrow384$. Each $16\times16$ input cell becomes a $2\times2$ stem block. Concatenating its moments yields 768 channels, which we project to 288 channels and process with four residual blocks with dilations (1, 2, 4, 8). A final $1\times1$ convolution predicts two unconstrained boundary strengths per cell, corresponding to its right and lower neighbours. No sigmoid is applied. Figure~\ref{fig:object_localizer} visualizes the object localizer's predictions on selected examples.

\begin{figure*}
    \centering
    \includegraphics[width=1\linewidth]{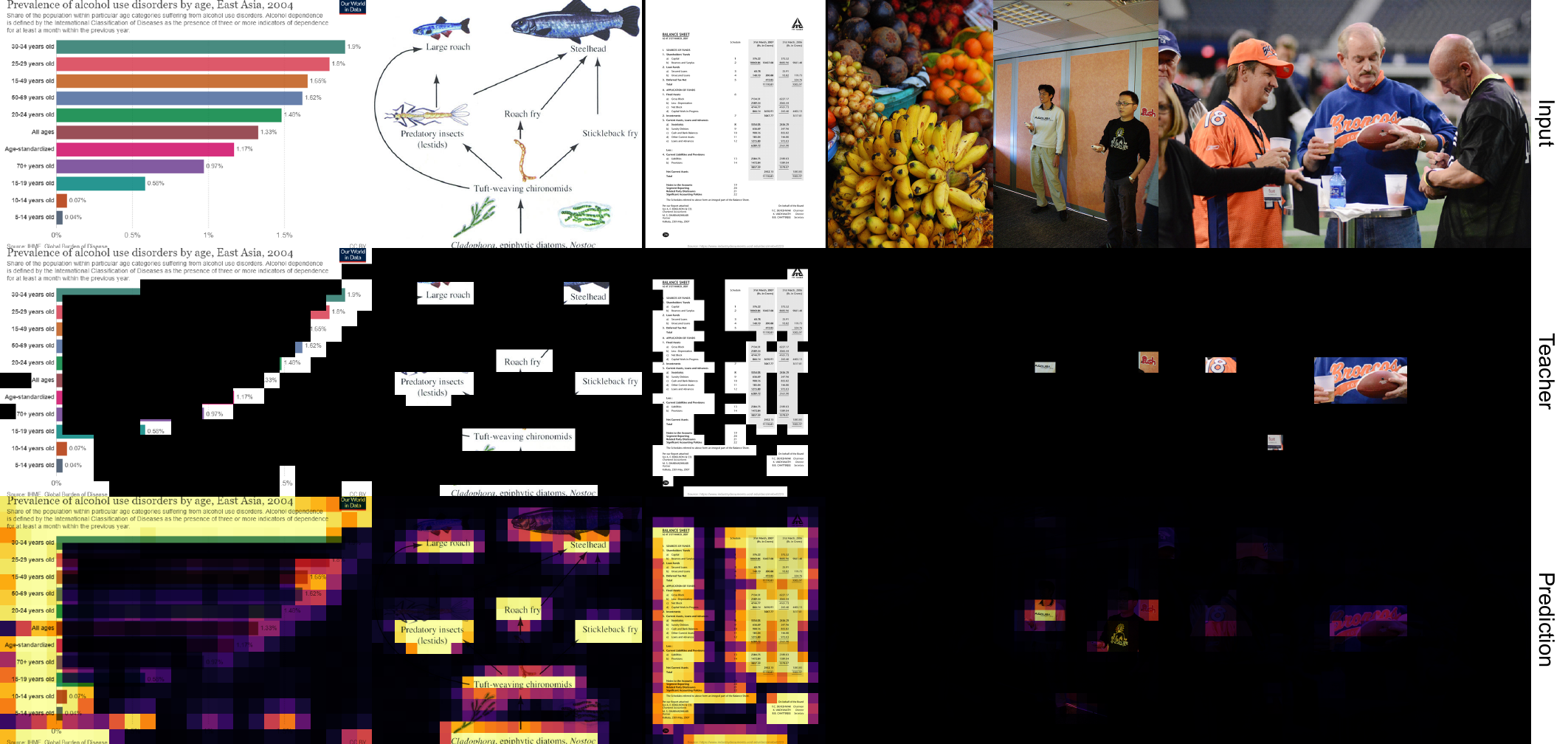}
    \caption{\textbf{Visualization of the text localizer.} Top: input images. Middle: EasyOCR-derived text labels, visualized by masking unlabeled cells. Bottom: per-cell textness predicted by the lightweight grayscale localizer, with brighter colors indicating higher probabilities. The examples span charts, diagrams, documents, and natural scenes,  illustrating localization of text at different scales and low responses to non-text textures in the fruit image.}
    \label{fig:text_localizer}
\end{figure*}

\paragraph{Text localizer.}
The grayscale stem has two layers with channel widths $1\rightarrow32\rightarrow64$. The image is bicubic-resized to $28n_c\times28n_r$ for a target grid of $n_r\times n_c$ cells. It yields a $7\times7$ stem block per cell. Moments are computed by average pooling, with variance $\max\{\operatorname{AvgPool}(\mathbf{y}^2)-\operatorname{AvgPool}(\mathbf{y})^2,0\}$. The concatenated 128 channels pass through a 128-channel projection and three residual blocks with dilations (1, 2, 4). A final $1\times1$ convolution produces one logit per cell, converted to textness by sigmoid. Figure~\ref{fig:text_localizer} visualizes the text localizer's predictions on selected examples. More importantly, the text localizer enables our method to perform well on no-object images as shown in Figure~\ref{fig:document_case}, unlike CORE, which only has a conceptually similar object localizer.

\paragraph{Alignment with image tokens.}
For LLaVA-1.5~\cite{liu2024llava15}, the object branch bicubic-resizes the image with aspect ratio preserved up to grid rounding, using a 768-pixel long side and 16-pixel cells. Its predictions are mapped to the visual grid by taking the maximum fine-grid boundary strength along each coarse boundary. Both fields occupy the cells whose centers lie inside the original image in the padded $24\times24$ grid. Padding has zero textness and zero internal boundary strengths. Its boundaries are assigned strength 9.

For LLaVA-NeXT~\cite{li2024llavanext-strong}, fields are aligned separately to the $24\times24$ base view and the stitched, unpadded $n_r\times n_c$ high-resolution view. The object branch runs once with long side $\max\{768,16\max(n_r,n_c)\}$, and its boundary field is max-pooled onto both grids. The text branch runs separately at each view's grid shape. Nonexistent outer edges are zero.

For Qwen3.8-27B, both object boundary and textness fields align to a single $n_r\times n_c$ grid after the native visual merger. Each cell covers $32\times32$ pixels of the resized image.

\subsubsection{Token-Budget Allocation}

\paragraph{Stage budgets.}
Without adaptive halting, midpoint pruning targets a layer-averaged image-token budget $K$, excluding prompt tokens and \texttt{[REL]} registers. The stage budgets are
\begin{equation}
    K_m=\operatorname{round}\!\left(\frac{8K}{5}\right), \qquad K_p=2K-K_m.
\end{equation}
For $K\in\{32,64,128,224\}$ on LLaVA-1.5, this gives $(K_m,K_p)=(51,13)$, $(102,26)$, $(205,51)$, and $(358,90)$, respectively. The 4:1 split is approximate after rounding.

\paragraph{Image-dependent NeXT budgets.}
For an unpadded high-resolution grid of $n_r\times n_c$ cells, the native sequence length is $N_{\mathrm{native}}=576+n_rn_c+n_r$, including one newline token per high-resolution row. A requested retention ratio $r$ determines
$K_{\mathrm{img}}=\max\{1,\lfloor rN_{\mathrm{native}}\rfloor\}$.
We compute
\begin{equation}
    \begin{aligned}
        K_m &=
        \operatorname{clip}\!\left(
            \operatorname{round}(8K_{\mathrm{img}}/5),
            6,576+n_rn_c
        \right),\\
        K_p &= \operatorname{clip}(2K_{\mathrm{img}}-K_m,0,K_m).
    \end{aligned}
\end{equation}
The merging budget is divided between the base and high-resolution views in proportion to their cell counts. We round the base-view share and assign the remainder to the high-resolution view, enforcing at least three tokens per view and each view's cell capacity. Both views use the same merging-stage ratio $r_m=K_m/N_{\mathrm{native}}$ to determine the textness threshold. Newline tokens enter the native-length accounting but are not emitted by our tokenizer. Capacity constraints can make the realized mean $(K_m+K_p)/2$ differ from $K_{\mathrm{img}}$.

\paragraph{Image-dependent Qwen budgets.}
The native grid contains $N=n_rn_c$ image tokens. We set $K_{\mathrm{img}}=\max\{1,\lfloor rN\rfloor\}$ and apply the stage-budget split above with $K=K_{\mathrm{img}}$, capping stage counts at the available image tokens. The textness threshold uses $r_m=K_m/N$.

\begin{figure*}
    \centering
    \includegraphics[width=1\linewidth]{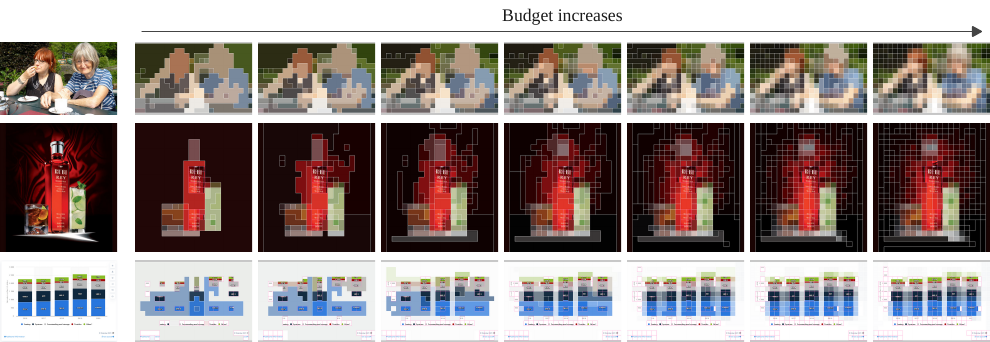}
    \caption{\textbf{\method{} tokenization across token budgets.} Text is kept at native resolution consistently, while extra budget splits objects into finer semantic parts, e.g., the head in the top row separates into face and hair.}
    \label{fig:tokenization_sweep}
\end{figure*}

\begin{figure}
    \centering
    \includegraphics[width=1\linewidth]{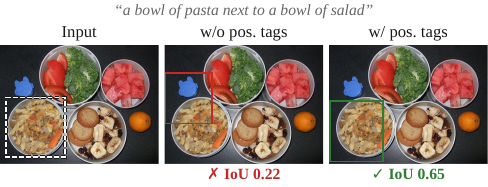}
    \caption{\textbf{\method{} compared with and without position tags.} Position tags enable more precise grounding coordinates.}
    \label{fig:supp_position_tag}
\end{figure}

\paragraph{Text and non-text groups.}
Let $M$ denote the merging budget assigned to one grid: $M=K_m$ for LLaVA-1.5 and Qwen3.8-27B, or the allocated view budget for NeXT. If letterbox padding is present, its cells form one reserved token, leaving a content budget $C=M-1$; otherwise, $C=M$. The builder requires $C\geq2$. Among image-content cells, we identify text by $T_i>\tau(r_m)$, where
\begin{equation}
    \tau(r_m)=1-\frac{1}{2}\operatorname{clip}(r_m,0,1).
\end{equation}
For LLaVA-1.5, $r_m=K_m/576$. Let $n_t$ and $n_o$ be the numbers of text and non-text cells.

\paragraph{Reserved non-text regions.}
To prevent text cells from consuming the entire budget, we reserve at least $k_{\min}$ regions for non-text content, estimated from the fused boundary field $B^*$ before separating text and non-text cells. Starting from individual cells on the complete aligned grid, including padding if present, we run weakest-mean-boundary agglomeration down to two regions. The reference threshold is
\begin{equation}
    q_B=\operatorname{Quantile}_{0.75}
    \left(\left\{
        \max(B^*_{i,\rightarrow},B^*_{i,\downarrow})
    \right\}_{i}\right).
\end{equation}
Let $k_0$ be the number of regions immediately after the first merge whose cost strictly exceeds $q_B$; if no crossing occurs, we set $k_0=1$. The budget-dependent reserve is
\begin{equation}
    k_{\min}=
    \max\!\left\{1,\min\!\left(
        k_0,\operatorname{round}(M/2)
    \right)\right\}.
\end{equation}
The cap uses the grid's total budget $M$, before reserving the padding token. Each NeXT view computes its own reserve. All rounding uses nearest-integer rounding with ties to even.

\paragraph{Quota assignment.}
When both groups are nonempty and $C-n_t\geq k_{\min}$, we preserve every text cell individually and allocate $C-n_t$ regions to non-text content. Otherwise, we initially allocate $\min\{k_{\min},C-1\}$ non-text regions and assign the remaining budget to text. We cap the non-text quota at $n_o$, recompute the remaining text quota, and cap it at $n_t$. Any unused capacity is reassigned first to non-text cells and then to text cells. If either group is empty, the other receives the content budget up to its cell count. Thus, every content cell remains represented, and the requested region count is attained whenever sufficient cells are available. Figure~\ref{fig:tokenization_sweep} illustrates how the resolution evolves as the budget grows.

\subsubsection{Region-Merging Procedure}

\paragraph{Graph construction.}
For each nonempty text or non-text group, each cell starts as its own region. Adjacent cells within the same group are connected by edges weighted by the fused boundary strength $B^*_{ij}$. Groups are processed independently, so merging never crosses the text/non-text split. Padding is excluded from the graph construction.

\paragraph{Merge updates.}
For each connected region pair $(R,S)$, we maintain the sum $A_{RS}$ of its edge weights and the number $n_{RS}$ of those edges. The merge priority is $d(R,S)=A_{RS}/n_{RS}$, implementing the mean-boundary rule in Equation~\ref{eq:varest_partition}. We repeatedly merge the pair with the smallest priority. After merging $R$ and $S$ into $U$, its connection to any neighbouring region $Q$ is updated by
\begin{equation}
    A_{UQ}=A_{RQ}+A_{SQ},
    \qquad
    n_{UQ}=n_{RQ}+n_{SQ},
\end{equation}
with absent connections contributing zero to both terms. The new priority is $A_{UQ}/n_{UQ}$. Thus, updates preserve the mean over constituent edges rather than averaging region-pair costs equally. A min-heap stores candidate merges, and a union-find structure tracks region membership; outdated heap entries are discarded or refreshed when encountered.

\begin{table*}
    \centering
    \captionsetup{justification=raggedright,singlelinecheck=false}
    \begingroup
    \small
    \setlength{\tabcolsep}{5pt}
    
    % Remove booktabs default white padding to fix the header gaps
    \setlength{\aboverulesep}{0pt}
    \setlength{\belowrulesep}{0pt}
    
    % Increase row height to add more margin inside rows
    \renewcommand{\arraystretch}{1.5} 
    
    \newcommand{\baselinecell}[1]{\raggedright #1}
    \newcommand{\iconcell}[1]{\centering #1}
    \newcommand{\baselinerule}{\arrayrulecolor{black!20}\specialrule{0.3pt}{0pt}{0pt}\arrayrulecolor{black}}
    
    % Custom icon command to shift the image down and stop it from hitting the top border
    \newcommand{\icon}[1]{\raisebox{-0.25\height}{\includegraphics[scale=0.06]{#1}}}

    \caption{\textbf{Localizer training.} Both localizers are trained once, separately from the VLM. Localizers were shared across LLaVA backbones but trained separately for Qwen to match its native resolution. They stay frozen during VLM adaptation and inference.}
    \begin{tabular}{
        p{\dimexpr0.03\linewidth-2\tabcolsep\relax}
        p{\dimexpr0.11\linewidth-2\tabcolsep\relax}
        p{\dimexpr0.43\linewidth-2\tabcolsep\relax}
        p{\dimexpr0.43\linewidth-2\tabcolsep\relax}
    }
        \toprule
        \rowcolor{green}
        \baselinecell{} & \baselinecell{} & \baselinecell{\textbf{Text localizer}} & \baselinecell{\textbf{Object localizer}} \tabularnewline
        \midrule
        
        \iconcell{\icon{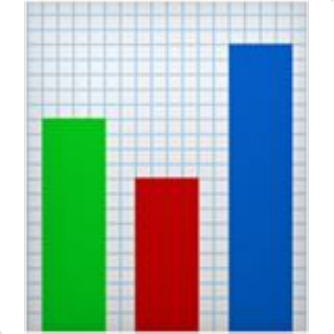}} 
            & \baselinecell{Data} 
            & \baselinecell{20{,}000 text-bearing images from the OCR-VQA and TextVQA subsets of LLaVA-665K, plus text-free negatives from all of LLaVA-665K at the corpus text-free:text ratio. We take a 3{,}000-image held-out split at the same ratio.}
            & \baselinecell{341{,}000 unique images from LLaVA-665K, 5\% held out for validation.} \tabularnewline
        \baselinerule
        
        \rowcolor{lightgreen!30}
        \iconcell{\icon{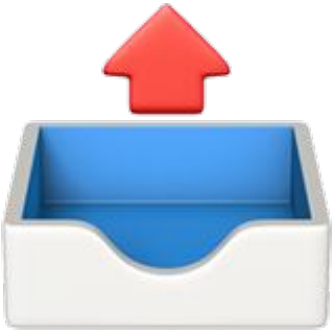}} 
            & \baselinecell{Input} 
            & \baselinecell{Grayscale, at the image's own aspect ratio, resized to 28\,px per cell of the $24{\times}24$ token grid. (32\,px on Qwen3.8-27B)}
            & \baselinecell{RGB, at the image's own aspect ratio, long side 768\,px, 16\,px cells. On Qwen3.8-27B, the image matches Qwen's dynamic resolution, with one 32\,px cell per visual token.} \tabularnewline
        \baselinerule
        
        \iconcell{\icon{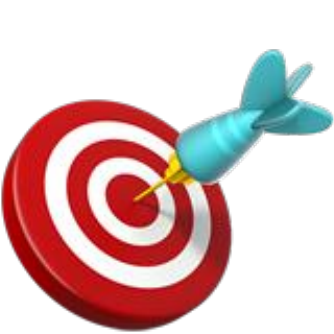}} 
            & \baselinecell{Target} 
            & \baselinecell{EasyOCR boxes, rasterized to cells; a cell is positive if any box touches it.}
            & \baselinecell{Cosine distance between adjacent DINOv3 ViT-L/16 patch features (right and lower neighbour), on the teacher's native patch grid.} \tabularnewline
        \baselinerule
        
        \rowcolor{lightgreen!30}
        \iconcell{\icon{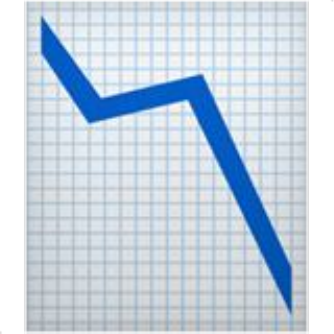}} 
            & \baselinecell{Loss} 
            & \baselinecell{Per-cell binary cross-entropy, positive weight $(1-p)/p$, where $p$ is the mean positive-cell fraction of the text-bearing training images.}
            & \baselinecell{L1 regression, masked on the non-existent last-column/last-row edges.} \tabularnewline
        \baselinerule
        
        \iconcell{\icon{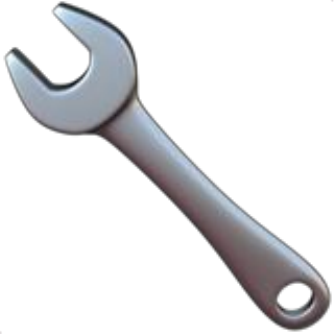}} 
            & \baselinecell{Optimizer} 
            & \baselinecell{AdamW, LR $10^{-3}$, weight decay 0.01, gradient clipping 1.0.}
            & \baselinecell{AdamW, LR $10^{-3}$, weight decay 0.01, gradient clipping 1.0.} \tabularnewline
        \baselinerule   
        
        \rowcolor{lightgreen!30}
        \iconcell{\icon{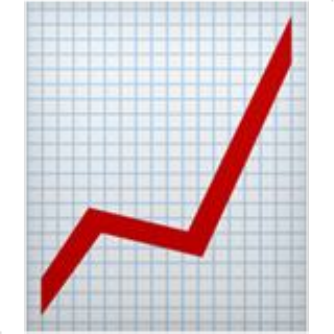}} 
            & \baselinecell{LR schedule} 
            & \baselinecell{Constant.}
            & \baselinecell{Cosine annealing to 0 over 100 epochs.} \tabularnewline
        \baselinerule
        
        \iconcell{\icon{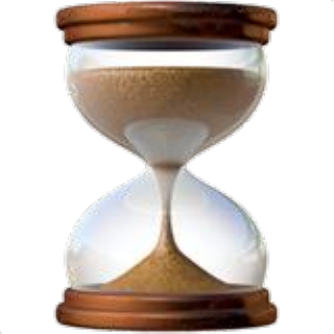}} 
            & \baselinecell{Batch / duration} 
            & \baselinecell{16 images, 30{,}000 steps. Batches group images with the same grid shape.}
            & \baselinecell{16 images, 100 epochs. Batches group images with the same grid shape.} \tabularnewline
        \baselinerule
        
        \rowcolor{lightgreen!30}
        \iconcell{\icon{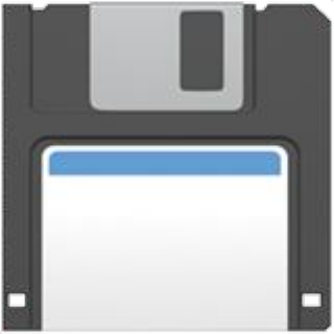}} 
            & \baselinecell{Checkpoint selection} 
            & \baselinecell{None; final weights.}
            & \baselinecell{Best validation Spearman $\rho$ to the teacher edges (the final epoch)} \tabularnewline
        \bottomrule
    \end{tabular}
    \label{tab:localizer_training}
    \endgroup
\end{table*}

\paragraph{Disconnected components.}
Separating text and non-text cells can produce disconnected components whose number exceeds the allocated quota. Before agglomeration, we connect these components with a sparse set of auxiliary edges. Starting from the first component in raster order, we repeatedly connect the nearest pair consisting of one component already in the tree and one outside it. Distances are measured between component centroids in cell coordinates. For components $C_a$ and $C_b$, the auxiliary edge joins their first cells in raster order and receives weight
\begin{equation}
    w_{\mathrm{bridge}}(C_a,C_b)
    =10^3+\|\mathbf{c}(C_a)-\mathbf{c}(C_b)\|_2.
\end{equation}
These weights are fixed before merging and enter the same sum-and-count updates as ordinary edges. Their large penalty defers merging across disconnected components, while allowing the requested quota to be reached. For such a merge, the resulting region can be spatially disconnected.

\paragraph{Termination and coverage.}
Agglomeration stops when the group's allocated region count is reached. If its quota equals its cell count, all cells remain individual regions; empty groups are skipped. Combining the two partitions and the reserved padding region assigns every grid cell to exactly one output region. The procedure therefore reduces token count through aggregation without discarding spatial coverage.

\begin{table*}
    % Audited against /workspace/input and saved llmoe_job FullToken results
    % for LLaVA-1.5-7B, LLaVA-1.5-13B, and LLaVA-NeXT-8B.
    \centering
    \captionsetup{justification=raggedright,singlelinecheck=false,hypcap=true}
    \begingroup
    \small
    \setlength{\tabcolsep}{5pt}
    \setlength{\aboverulesep}{0pt}
    \setlength{\belowrulesep}{0pt}
    \renewcommand{\arraystretch}{1.5}
    \definecolor{tableBanner}{HTML}{79B8E8}
    \definecolor{tableStripe}{HTML}{F2F8FD}
    \newcommand{\baselinecell}[1]{\raggedright #1}
    % Use the supplied vector emoji exports directly; no extra packages needed.
    \newcommand{\protocolicon}[1]{\raisebox{-0.25\height}{\includegraphics[scale=0.06]{images/icons/#1.pdf}}\hspace{0.6em}}
    \newcommand{\baselinerule}{\arrayrulecolor{black!20}\specialrule{0.3pt}{0pt}{0pt}\arrayrulecolor{black}}

    \caption{\textbf{Benchmark evaluation protocols.} Generation uses greedy decoding, one beam, and KV caching. Output limits are 1,024 tokens for MMBench, 16 for CountBenchQA and V*, and 128 otherwise, except that Qwen3.8-27B also uses 1,024 for MMStar and AI2D. All listed evaluation sets are used without an additional sample limit.}
    \begin{tabular}{
        p{\dimexpr0.22\linewidth-2\tabcolsep\relax}
        p{\dimexpr0.38\linewidth-2\tabcolsep\relax}
        p{\dimexpr0.40\linewidth-2\tabcolsep\relax}
    }
        \toprule
        \rowcolor{tableBanner}
        \baselinecell{\textbf{Benchmark}} & \baselinecell{\textbf{Evaluation set}} & \baselinecell{\textbf{Reported metric}} \tabularnewline
        \midrule
        \baselinecell{\protocolicon{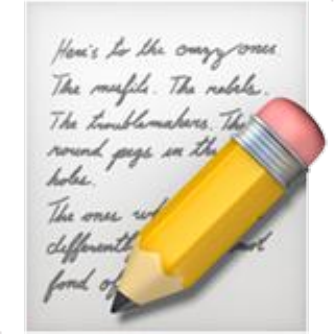}MMBench~\cite{liu2024mmbench}} & \baselinecell{English dev, 20230712; 4,377 questions} & \baselinecell{Circular accuracy} \tabularnewline
        \baselinerule
        \rowcolor{tableStripe}
        \baselinecell{\protocolicon{vqa}AI2D~\cite{kembhavi2016ai2d}} & \baselinecell{Test; 1,904 text-option and 1,184 letter-label questions} & \baselinecell{Accuracy over all 3,088 questions} \tabularnewline
        \baselinerule
        \baselinecell{\protocolicon{vqa}MMStar~\cite{chen2024mmstar}} & \baselinecell{Validation; 1,500 questions} & \baselinecell{Accuracy} \tabularnewline
        \baselinerule
        \rowcolor{tableStripe}
        \baselinecell{\protocolicon{vqa}POPE~\cite{li2023pope}} & \baselinecell{COCO val2014; random: 2,910; popular/adversarial: 3,000 each} & \baselinecell{Positive-class (yes) F1, separately per variant} \tabularnewline
        \baselinerule
        \baselinecell{\protocolicon{vqa}GQA~\cite{hudson2019gqa}} & \baselinecell{Testdev-balanced; 12,578 questions} & \baselinecell{Normalized exact-match accuracy} \tabularnewline
        \baselinerule
        \rowcolor{tableStripe}
        \baselinecell{\protocolicon{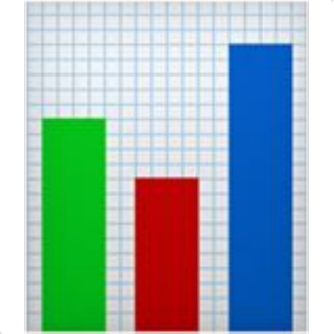}ChartQA~\cite{masry2022chartqa}} & \baselinecell{Test; 1,250 human and 1,250 augmented questions} & \baselinecell{Relaxed accuracy; mean of the two subset accuracies} \tabularnewline
        \baselinerule
        \baselinecell{\protocolicon{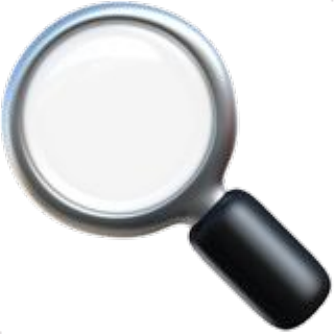}DocVQA~\cite{mathew2021docvqa}} & \baselinecell{SP-DocVQA v1.0 validation; 5,349 questions} & \baselinecell{ANLS; similarities below 0.5 score zero} \tabularnewline
        \baselinerule
        \rowcolor{tableStripe}
        \baselinecell{\protocolicon{ocr}TextVQA~\cite{singh2019textvqa}} & \baselinecell{v0.5.1 validation; 5,000 questions matched to LLaVA's OCR prompt file} & \baselinecell{Mean $\min(m/3,1)$; $m$ is the number of matching normalized answers among ten references} \tabularnewline
        \baselinerule
        \baselinecell{\protocolicon{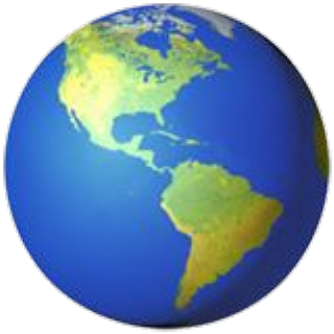}RefCOCO~\cite{yu2016refcoco}} & \baselinecell{UNC validation; 10,834 expressions} & \baselinecell{Accuracy at IoU $\geq0.5$} \tabularnewline
        \baselinerule
        \rowcolor{tableStripe}
        \baselinecell{\protocolicon{grounding}RefCOCO+~\cite{yu2016refcoco}} & \baselinecell{UNC validation; 10,758 expressions} & \baselinecell{Accuracy at IoU $\geq0.5$} \tabularnewline
        \baselinerule
        \baselinecell{\protocolicon{grounding}RefCOCOg~\cite{mao2016refcocog}} & \baselinecell{UMD validation; 4,896 expressions} & \baselinecell{Accuracy at IoU $\geq0.5$} \tabularnewline
        \baselinerule
        \rowcolor{tableStripe}
        \baselinecell{\protocolicon{counting}CountBenchQA \cite{paiss2023countbench}} & \baselinecell{Test; 491 question--image pairs} & \baselinecell{Exact count accuracy} \tabularnewline
        \baselinerule
        \baselinecell{\protocolicon{grounding}V*~\cite{wu2024vstar}} & \baselinecell{Test; 115 direct-attribute and 76 relative-position questions} & \baselinecell{Accuracy over all 191 questions} \tabularnewline
        \bottomrule
    \end{tabular}
    \label{tab:supp_evaluation_protocol}
    \endgroup
\end{table*}

\subsubsection{Region Encoding and Backbone Integration}

\paragraph{Feature pooling.}
For all three LLaVA backbones, we extract patch features from the backbone's configured vision layer, discard the classification token, and apply the multimodal projector to each patch. For Qwen3.8-27B, we use the output of the native visual merger. Region pooling therefore operates in the language-model embedding space, after vision encoding and projection. For each region, all constituent cells contribute equally to the mean in Equation~\ref{eq:varest_merge}.

\paragraph{Position tags.}
For a region $R$ on an $n_r\times n_c$ grid, let $(r_i,c_i)$ be the zero-based row and column indices of cell $i$. Its normalized centroid and extent are
\begin{equation}
    \begin{aligned}
        c_x(R)&=\frac{1}{|R|n_c}\sum_{i\in R}(c_i+\tfrac12),\\
        c_y(R)&=\frac{1}{|R|n_r}\sum_{i\in R}(r_i+\tfrac12),\\
        a(R)&=\sqrt{\frac{|R|}{n_rn_c}}.
    \end{aligned}
\end{equation}
The half-cell offset places each coordinate at the cell center. The extent measures the square root of the region's fraction of the grid, using all member cells even when the region is disconnected. The position-tag MLP has the architecture of
\begin{equation}
    \operatorname{Linear}(3,256)
    \rightarrow \operatorname{GELU}
    \rightarrow \operatorname{Linear}(256,d_{\mathrm{LLM}}),
\end{equation}
with biases in both linear layers. Its output is added to the pooled feature. The MLP is trained separately for each backbone and shared across its regions and image views. Figure~\ref{fig:supp_position_tag} and Table~\ref{tab:supp_position_tag_ablation} show their effect on grounding.

\paragraph{Backbone geometry.}
For the two LLaVA-1.5 models, pooling and coordinate normalization use the complete $24\times24$ padded grid. The reserved padding region is encoded by the same mean-and-tag operation as other regions. For LLaVA-NeXT, we retain the AnyRes resolution selection and tiling. The base image produces a $24\times24$ feature grid, while each high-resolution tile produces its own $24\times24$ grid. We stitch the projected tile features into their spatial layout and remove padding using the backbone's unpadding rule. Region encoding then operates separately on the base grid and the stitched high-resolution grid, with coordinates and extent normalized within each view. Regions may span tile boundaries within the stitched view. For Qwen3.8-27B, we retain the backbone's native image resizing and
operate on a single $n_r\times n_c$ grid whose dimensions depend on the resized image. The native visual merger combines each $2\times2$ group of $16\times16$ patches into one token, so each grid cell corresponds to a $32\times32$ block of the resized image. We take the mean token in each region and add a position tag computed from the region centroid and extent normalized over the full grid.

\paragraph{Sequence construction.}
Within each view, regions are ordered by the raster index of their first cell. LLaVA-NeXT concatenates the base-view regions before the high-resolution regions and emits no image-newline tokens. The resulting visual sequence occupies the image position in the backbone's chat template, before the user prompt. On LLaVA, the assembled sequence receives consecutive decoder position IDs; position tags supply the region geometry in addition to this sequence-position information. On Qwen3.8-27B, each region instead receives the M-RoPE height and width position of its region centre.

\begin{figure*}
    \centering
    \includegraphics[width=1\linewidth]{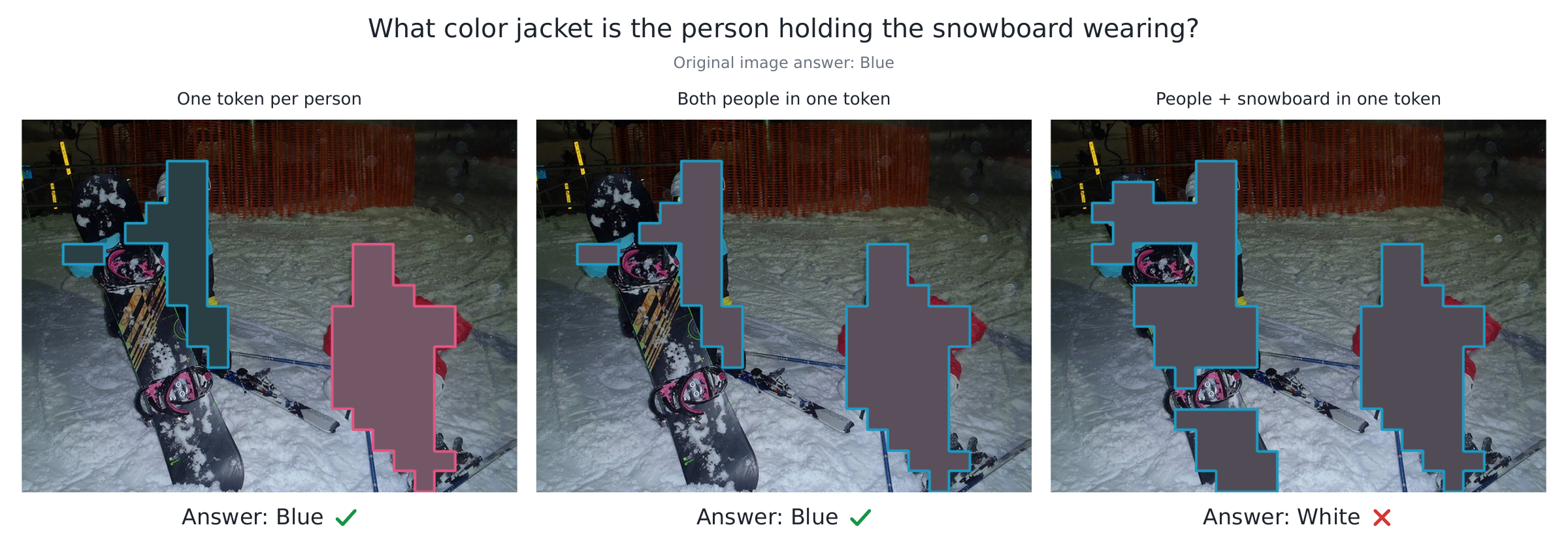}
    \caption{\textbf{Semantic compatibility affects token merging.} Highlighted patches are merged into one token per person (left), one token shared by both people (middle), or one token combining both people and the snowboard (right). The first two configurations preserve the original answer, ``Blue,'' while including the snowboard changes the answer to ``White.'' This example illustrates how merging semantically different content can disrupt an attribute needed for answering, even when merging multiple instances of the same category preserves it.}
    \label{fig:object_merging}
\end{figure*}

\begin{figure*}
    \centering
    \includegraphics[width=1\linewidth]{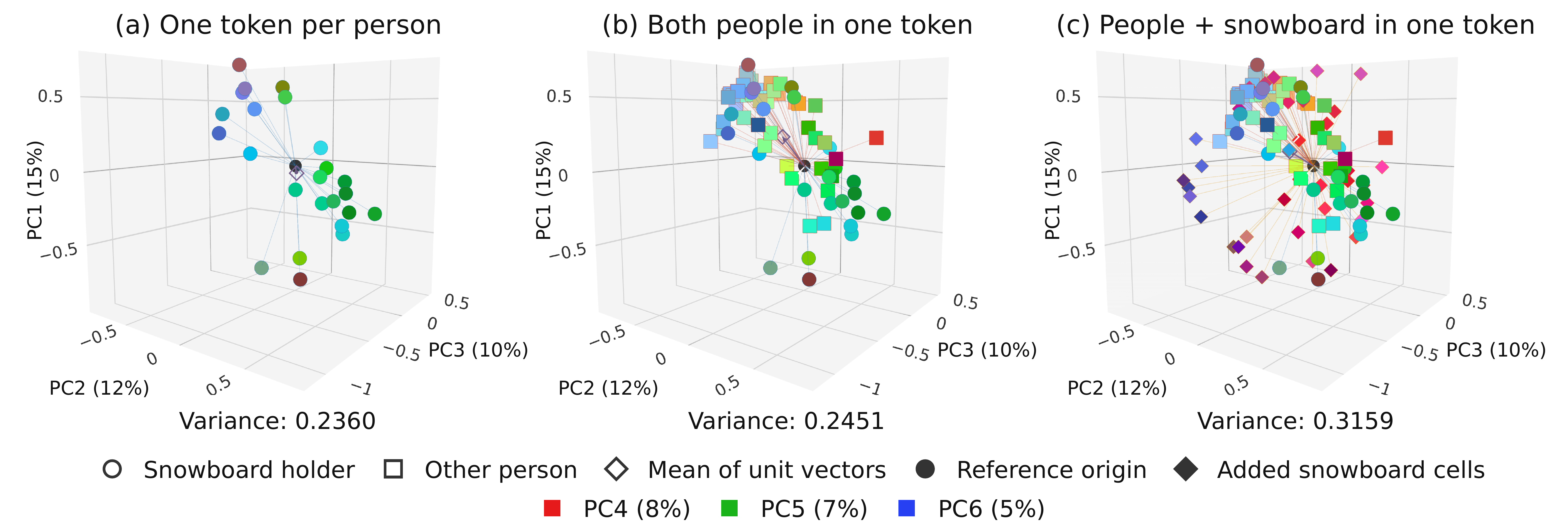}
    \caption{\textbf{Semantic compatibility and token feature dispersion.} Adding another person changes feature variance only slightly, whereas including the snowboard produces a larger variance increase. The people + snowboard group is too complex to summarize it in one token.}
    \label{fig:object_merging_pca}
\end{figure*}

\subsubsection{\texttt{[REL]} Necessity Pruning}

\paragraph{Registers and scoring.}
Four learned \texttt{[REL]} embeddings are inserted after the user turn, directly after the prompt. Selection occurs at the midpoint, which is after 16 completed decoder layers for LLaVA-1.5-7B and LLaVA-NeXT-8B, after 20 for LLaVA-1.5-13B, and after 32 for Qwen3.8-27B. The scoring head follows Equation~\ref{eq:rel_scores}, using non-affine LayerNorm and bias-free projections to 256 dimensions. Each register state is normalized before averaging.

\paragraph{Differentiable selection.}
Let $n_v$ be the number of image tokens at the selection boundary. During training, we standardize their scores as
\begin{equation}
    \bar{s}_j=
    \frac{s_j-\operatorname{mean}_k(s_k)}
    {\max\{\operatorname{Std}_k(s_k),10^{-3}\}},
\end{equation}
using the sample standard deviation and retaining gradients through this operation. Standardization is skipped when $n_v=1$. Let $m_j\in\{0,1\}$ indicate membership in the top-$K_p$ set, computed from detached scores. For $0<K_p<n_v$, the detached threshold $\theta$ is the midpoint between the $K_p$-th and $(K_p+1)$-th highest standardized scores.
At the endpoints, we use $\theta=\max_j\bar{s}_j+\gamma$ for $K_p=0$ and $\theta=\min_j\bar{s}_j-\gamma$ for $K_p=n_v$. A straight-through surrogate~\cite{bengio2013estimating} defines
\begin{equation}
    \begin{aligned}
        u_j&=\sigma\!\left(\frac{\bar{s}_j-\theta}{\gamma}\right),
        \qquad \gamma=0.1,\\
        \widetilde{m}_j&=m_j+u_j-\operatorname{sg}(u_j),
    \end{aligned}
    \label{eq:rel_mask_details}
\end{equation}
where $\operatorname{sg}$ stops gradients. The forward pass uses binary selection, while the backward pass follows the sigmoid surrogate.

\paragraph{Attention masking during training.}
A single selection mask is reused in every subsequent decoder layer and attention head. It multiplies the exponentiated attention logits of visual keys before renormalization:
\begin{equation}
    \alpha_{ij}
    =
    \frac{w_j\exp(e_{ij})}
         {\sum_{k\leq i}w_k\exp(e_{ik})},
    \qquad j\leq i,
\end{equation}
where $e_{ij}$ is the scaled query-key logit, $w_j=\widetilde{m}_j$ for image tokens, and $w_j=1$ for text tokens. After selection, register's keys have weight 1 on LLaVA and 0 on Qwen3.8-27B. Causal masking remains in effect. For numerical stability, logits are shifted by a detached row maximum, and the denominator is clamped below at $10^{-30}$. Pruned visual tokens are kept during training; however, keys are masked so no later token attends to them. The answer cross-entropy loss supplies gradients to the selector without token-level selection labels. Qwen3.8-27B applies this mask only in its full-attention layers. Its 48 Gated DeltaNet layers instead mix only text and retained image tokens after selection.

\begin{figure*}[t]
    \centering
    \includegraphics[width=\linewidth]{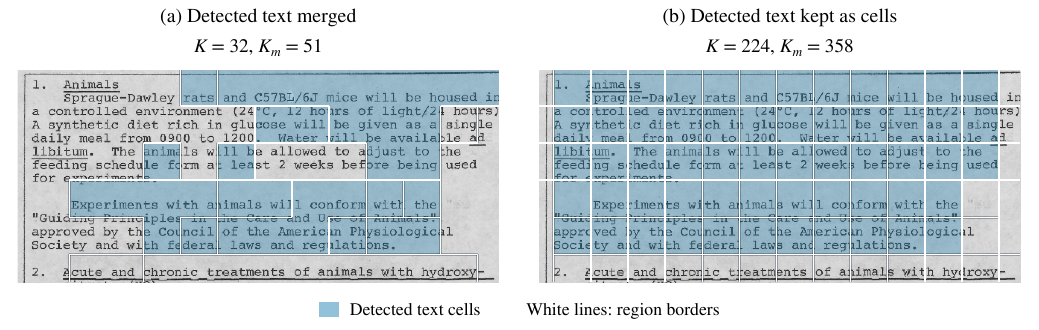}
    \caption{\textbf{Text resolution across budgets.} Varest partitions of the same document on LLaVA-1.5-7B, before decoder pruning. Blue shading marks detected text cells; white lines are region borders. At the smaller budget, detected text cells are grouped into fewer regions. At the larger budget, each detected text cell receives its own region.}
    \label{fig:supp_tokenizer_diagnostics}
\end{figure*}

\paragraph{Physical removal at inference.}
At the midpoint, we retain the $K_p$ highest-scoring image tokens, restore their original sequence order, and physically remove the remaining image tokens together with the registers.
Surviving tokens preserve their original position tags. Selection is performed only during prefill; thus, the KV cache contains only the retained image and text tokens after the midpoint.

\subsubsection{Adaptive Halting Implementation}
\label{sec:supp_adaptive_halting_implementation}

Rather than keeping image tokens to a fixed depth, adaptive halting predicts each input’s information horizon from the LLM’s intermediate predictions and withdraws the image once the horizon is detected. Algorithm~\ref{alg:adaptive_halting} implements the criterion during prefill, independently for each example.

\begin{algorithm}[t]
\caption{Prefill with Adaptive Halting}
\label{alg:adaptive_halting}
\small
\begin{algorithmic}[1]
\Require Prompt states $\mathbf{H}$, decoder depth $L$,
         pruning budget $K_p$
\State $\mathrm{halted}\gets\mathrm{False}$
\For{$\ell=0,\ldots,L-1$}
    \If{not $\mathrm{halted}$}
        \State $\mathbf{p}_\ell\gets
            \operatorname{Lens}(\mathbf{H}_{\mathrm{last}})$
        \State $\hat{y}_\ell\gets\arg\max_v p_\ell(v)$
        \If{$\ell\geq8$ and
            $\operatorname{Stable}_3(\hat{y}_{0:\ell})$ and
            $H(\mathbf{p}_\ell)\leq1$}
            \State Remove all image tokens and registers
                   from $\mathbf{H}$
            \State $\mathrm{halted}\gets\mathrm{True}$
        \EndIf
    \EndIf
    \If{$\ell=L/2$ and not $\mathrm{halted}$}
        \State Retain the $K_p$ image tokens selected by
               \texttt{[REL]}
        \State Remove the other image tokens and registers
    \EndIf
    \State $\mathbf{H}\gets
        \operatorname{Decoder}_{\ell+1}(\mathbf{H})$,
        storing its KV cache
\EndFor
\State Generate the answer using the final states and KV caches
\end{algorithmic}
\end{algorithm}

During prefill, after each decoder layer $\ell$, we apply a logit lens~\cite{nostalgebraist2020logitlens} to the hidden
state $\mathbf{h}^{(\ell)}_{T}$ at the last prompt position $T$, the final token of the assistant-turn prefix:
\begin{equation}
    \begin{aligned}
        \mathbf{p}_\ell &= \operatorname{softmax}\!\bigl(W_U\operatorname{Norm}(\mathbf{h}^{(\ell)}_{T})\bigr),\\
        \hat{y}_\ell &= \arg\max_{v}\,p_\ell(v),\quad
        H_\ell = -\sum_{v}p_\ell(v)\log p_\ell(v),
    \end{aligned}
    \label{eq:halt_lens}
\end{equation}
where $\operatorname{Norm}$ is the LLM's final normalization, and
$W_U$ is its frozen base vocabulary head, with LoRA updates bypassed. This readout is $\operatorname{Lens}$ in Algorithm~\ref{alg:adaptive_halting}. The distribution $\mathbf{p}_\ell$ estimates the first answer token from the intermediate hidden state. We use prediction stability and confidence as heuristics. Halting occurs at the first layer satisfying all three conditions:
\begin{enumerate}
    \item \textbf{Warm-up:} $\ell \geq 8$, excluding shallow
          readouts from triggering premature withdrawal.
    \item \textbf{Stability:}
          $\hat{y}_{\ell-2}=\hat{y}_{\ell-1}=\hat{y}_{\ell}$,
          so the top-1 prediction agrees across three
          consecutive layers ($\operatorname{Stable}_3$ in Algorithm~\ref{alg:adaptive_halting}).
    \item \textbf{Confidence:} $H_\ell \leq 1$ nat.
\end{enumerate}

Once triggered, we remove all remaining image tokens and \texttt{[REL]} registers. Here, $\ell$ counts completed layers, so halting removes tokens before layer $\ell+1$ and takes precedence over midpoint pruning. If no trigger occurs by Layer $L/2$, standard \texttt{[REL]} necessity pruning proceeds, and halting can still trigger afterwards. Later layers cache only the shortened prompt. Answer generation continues through all decoder layers using the adapted vocabulary head, with position IDs continuing from the original prompt length.

\begin{figure*}
    \includegraphics[width=\linewidth]{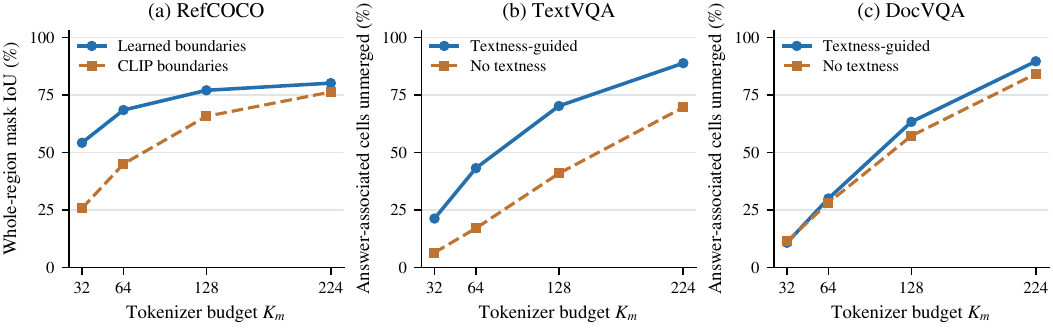}
    \caption{\textbf{Spatial detail retained by the tokenizer.} (a) Whole-region mask IoU on RefCOCO with learned or CLIP-derived boundaries. (b--c) Answer-associated OCR cells kept unmerged, with predicted or zero textness. Each comparison fixes the other field. $K_m$ counts regions before decoder pruning, not the layer-averaged budget.}
    \label{fig:supp_partition_detail}
\end{figure*}

\subsection{Training Details}
\label{sec:supp_training}

The object localizer is trained through distillation on DINOv3 ViT-L/16, while the text localizer learns from EasyOCR~\cite{jaidedai2020easyocr}. Training settings are described in Table~\ref{tab:localizer_training}. We describe the loss formulation below; both are averaged across the batch, where each batch contains only images with the same aspect ratio (if it divides into the same grid shape).

\paragraph{Object localizer.}
Consider a training image on the teacher's $n_r\times n_c$ patch grid. Let $\mathbf{f}_i$ be the DINOv3 feature of cell $i$, and let $o\in\{\rightarrow,\downarrow\}$ index its right and lower neighbour $i_o$. DINOv3 features change sharply across object boundaries. We define teacher edge strength by the cosine distance:
\begin{equation}
    D_{i,o}=1-\frac{\mathbf{f}_i^{\top}\mathbf{f}_{i_o}}{\|\mathbf{f}_i\|_2\,\|\mathbf{f}_{i_o}\|_2}.
\end{equation}

Let $\mathcal{E}$ be the set of edges, excluding the right edges of the last column and the lower edges of the last row, so $|\mathcal{E}|=n_r(n_c-1)+(n_r-1)n_c$. Then, the loss is the average difference from the teacher's edge strength across all edges:
\begin{equation}
    \mathcal{L}_{\mathrm{obj}}=\frac{1}{|\mathcal{E}|}\sum_{(i,o)\in\mathcal{E}}\bigl|\hat{B}_{i,o}-D_{i,o}\bigr|.
\end{equation}

\paragraph{Text localizer.}
For an image with $N$ cells, let $y_i\in\{0,1\}$ indicate that an EasyOCR box touches cell $i$, and let $\hat{T}_i=\sigma(z_i)$ be the predicted textness from logit $z_i$. As text is sparse in most images, an unweighted loss would favour predicting no text everywhere. Thus the positive weight is:
\begin{equation}
    w_{+}=\frac{1-p}{p},\qquad
    p=\frac{1}{|\mathcal{I}_{\mathrm{text}}|}\sum_{m\in\mathcal{I}_{\mathrm{text}}}
      \frac{1}{N_m}\sum_{i=1}^{N_m}y_{m,i},
\end{equation}
where $\mathcal{I}_{\mathrm{text}}$ is the set of text-bearing training images and image $m$ has $N_m$ cells with labels $y_{m,i}$. This balances the classes. Text-free negatives are excluded from $p$ so that they do not inflate $w_{+}$. For each image, the loss is the positive-weighted binary cross-entropy averaged over its cells:
\begin{equation}
    \mathcal{L}_{\mathrm{text}}=-\frac{1}{N}\sum_{i=1}^{N}
    \Bigl[w_{+}\,y_i\log\hat{T}_i+(1-y_i)\log\bigl(1-\hat{T}_i\bigr)\Bigr].
\end{equation}

\begin{table}
    \centering
    \captionsetup{justification=raggedright,singlelinecheck=false}
    \begingroup
    \footnotesize
    \setlength{\tabcolsep}{3pt}
    \setlength{\aboverulesep}{0pt}
    \setlength{\belowrulesep}{0pt}
    \renewcommand{\arraystretch}{1.3}

    \newcommand{\baselinecell}[1]{\raggedright #1}
    \newcommand{\iconcell}[1]{\centering #1}
    \newcommand{\baselinerule}{\arrayrulecolor{black!20}\specialrule{0.3pt}{0pt}{0pt}\arrayrulecolor{black}}
    \newcommand{\icon}[1]{\raisebox{-0.25\height}{\includegraphics[scale=0.06]{#1}}}
    \newcommand{\sharedcell}[1]{\multicolumn{2}{p{\dimexpr0.72\linewidth-2\tabcolsep\relax}}{\raggedright #1}}

    \caption{\textbf{Joint adaptation.} Training hyperparameters per backbone. Full-width entries are shared.}
    \begin{tabular}{
        p{\dimexpr0.08\linewidth-2\tabcolsep\relax}
        p{\dimexpr0.20\linewidth-2\tabcolsep\relax}
        p{\dimexpr0.36\linewidth-2\tabcolsep\relax}
        p{\dimexpr0.36\linewidth-2\tabcolsep\relax}
    }
        \toprule
        \rowcolor{green}
        \baselinecell{} & \baselinecell{} & \baselinecell{\textbf{LLaVA-1.5 / NeXT}} & \baselinecell{\textbf{Qwen3.8-27B}} \tabularnewline
        \midrule

        \iconcell{\icon{images/icons/data.pdf}}
            & \baselinecell{Data}
            & \baselinecell{665K (624{,}609) / NeXT-Data (738{,}590)}
            & \baselinecell{665K (624{,}609) with Qwen-format boxes} \tabularnewline
        \baselinerule

        \rowcolor{lightgreen!30}
        \iconcell{\icon{images/icons/downchart.pdf}}
            & \baselinecell{Loss}
            & \sharedcell{Teacher-forced cross-entropy on answer tokens} \tabularnewline
        \baselinerule

        \iconcell{\icon{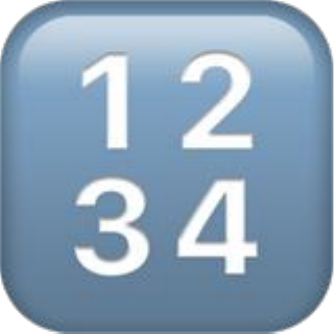}}
            & \baselinecell{LoRA Params}
            & \sharedcell{$r{=}32$, $\alpha{=}64$ \newline dropout=0.05 for LLaVA; no dropout for Qwen} \tabularnewline
        \baselinerule

        \iconcell{\icon{images/icons/target.pdf}}
            & \baselinecell{LoRA Scope}
            & \sharedcell{LLM (attention for LLaVA; token mixers for Qwen), vision attention, and LM head} \tabularnewline
        \baselinerule

        \rowcolor{lightgreen!30}
        \iconcell{\icon{images/icons/input.pdf}}
            & \baselinecell{Full training}
            & \sharedcell{Projector (merger for Qwen) fine-tuned\newline Position tag and \texttt{[REL]} head trained from scratch} \tabularnewline
        \baselinerule

        \iconcell{\icon{images/icons/wrench.pdf}}
            & \baselinecell{Optimizer}
            & \sharedcell{AdamW, constant LR $2{\times}10^{-5}$ and gradient clipping 1.0 for all parameters except \texttt{[REL]}} \tabularnewline
        \baselinerule

        \rowcolor{lightgreen!30}
        \iconcell{\icon{images/icons/upchart.pdf}}
            & \baselinecell{Warm-up, decay}
            & \baselinecell{No warm-up, weight decay 0.01}
            & \baselinecell{3\% linear warm-up, no weight decay} \tabularnewline
        \baselinerule

        \iconcell{\icon{images/icons/target.pdf}}
            & \baselinecell{\texttt{[REL]} head}
            & \baselinecell{LR $10^{-3}$ with 2k-step warm-up, no weight decay or clipping}
            & \baselinecell{LR $10^{-3}$ with the shared warm-up, clipped separately at 1.0} \tabularnewline
        \baselinerule

        \rowcolor{lightgreen!30}
        \iconcell{\icon{images/icons/hourglass.pdf}}
            & \baselinecell{Batch size}
            & \baselinecell{1}
            & \baselinecell{4 (global)} \tabularnewline
        \baselinerule

        \iconcell{\icon{images/icons/floppydisk.pdf}}
            & \baselinecell{Checkpoint SWA}
            & \sharedcell{Final weights averaged with last four snapshots, 32k samples apart (32k steps on LLaVA, 8k on Qwen)} \tabularnewline
        \bottomrule
    \end{tabular}
    \label{tab:training_hparams}
    \endgroup
\end{table}

\paragraph{Joint adaptation.}
Each backbone is trained for one epoch over the image rows of its instruction-tuning data, using the first turn of each conversation. To train a model that works at all budgets, each training step samples a merging budget $K_m$ and a pruning budget $K_p$. On the two LLaVA-1.5 backbones, $K_m$ is drawn uniformly from 32 to 576 tokens, while on LLaVA-NeXT and Qwen3.8-27B, it is a uniform fraction of $N$ between $32/576$ and 1. $K_p$ is then drawn uniformly from 0 to $K_m$, so the model also sees cases with no image tokens after the midpoint. Table~\ref{tab:training_hparams} lists the remaining settings. It takes 54, 61, and 92 hours on one H100 for LLaVA-1.5-7B, LLaVA-1.5-13B, and LLaVA-NeXT-8B, respectively, and 37 hours on four H100 GPUs for Qwen3.8-27B.

\subsection{Evaluation Settings}
\label{sec:supp_evaluation}

\paragraph{Baseline configurations.}
Table~\ref{tab:supp_baseline_adaptations} summarizes the baseline training requirements on all four backbones.

\paragraph{Budget matching.}
For a decoder with $L$ layers, a schedule's realized image-token budget is $\bar K=L^{-1}\sum_{\ell=1}^{L}n_\ell$, where $n_\ell$ is the number of image tokens processed at layer $\ell$. This includes early unpruned stages and assigns zero to layers before visual injection or after visual exit. On NeXT, the labels $1120/640/320/160$ express retention ratios relative to a 2,880-token reference; per-image targets depend on the native sequence length. On Qwen3.8-27B, the listed ratios apply to each image's native token count $N$. Integer rounding and capacity limits can cause deviations from the requested budget. When transferring LLaVA-1.5 schedules from 7B to 13B, we scale pruning, injection, and exit positions proportionally to decoder depth and adjust retained counts where needed to preserve the layer average.

\paragraph{Benchmark protocols.}
Table~\ref{tab:supp_evaluation_protocol} specifies the splits and scoring rules. Qwen3.8-27B uses its native chat template with thinking disabled and scores multiple-choice responses by the final answer letter. Grounding responses use JSON boxes with coordinates on a 0--1000 scale of the unpadded image.

\subsection{Compute Resources}

During development, we used an NVIDIA RTX 4090 (24 GB), an NVIDIA RTX 5090 (32 GB), 8 NVIDIA H100 (80 GB), and 8 NVIDIA H200 (141 GB) GPUs for about 6000 GPU hours in total to run experiments such as full fine-tuning and hyperparameter searches to determine the ceiling and directions. Inference and evaluation utilized 32 NVIDIA H100 (80 GB) GPUs concurrently, whereas other experiments were conducted on a single NVIDIA H100 (80 GB) GPU, unless otherwise specified.

\begin{figure}
    \centering
    \includegraphics[width=\linewidth]{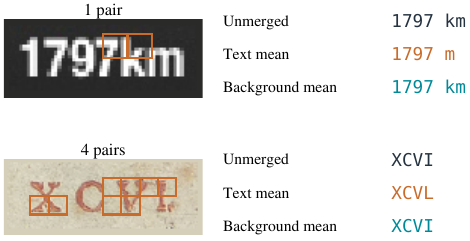}
    \caption{\textbf{Reading errors without token deletion.} Orange boxes mark patches whose features are averaged. All 576 tokens remain at their original positions. Background averaging preserves the original readings. These are selected error examples; Table~\ref{tab:supp_text_averaging} reports the full cohort.}
    \label{fig:supp_text_averaging}
\end{figure}

\begin{figure*}
    \centering
    \begin{subfigure}[t]{0.48\textwidth}
        \centering
        \includegraphics[width=1\linewidth]{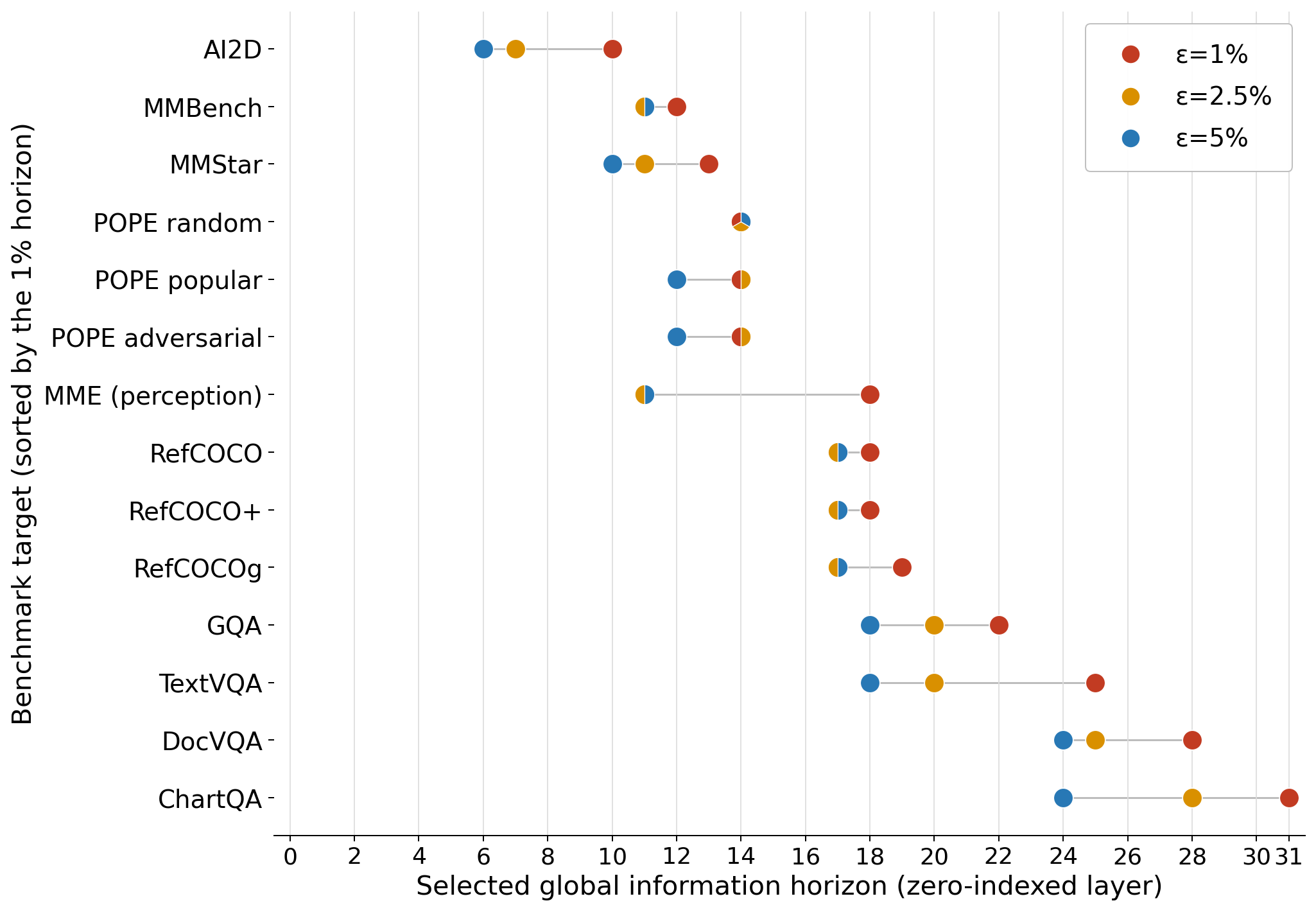}
        \caption{LLaVA-1.5-7B}
        \label{fig:information_horizon_llava_7}
    \end{subfigure}
    \begin{subfigure}[t]{0.48\textwidth}
        \centering
        \includegraphics[width=1\linewidth]{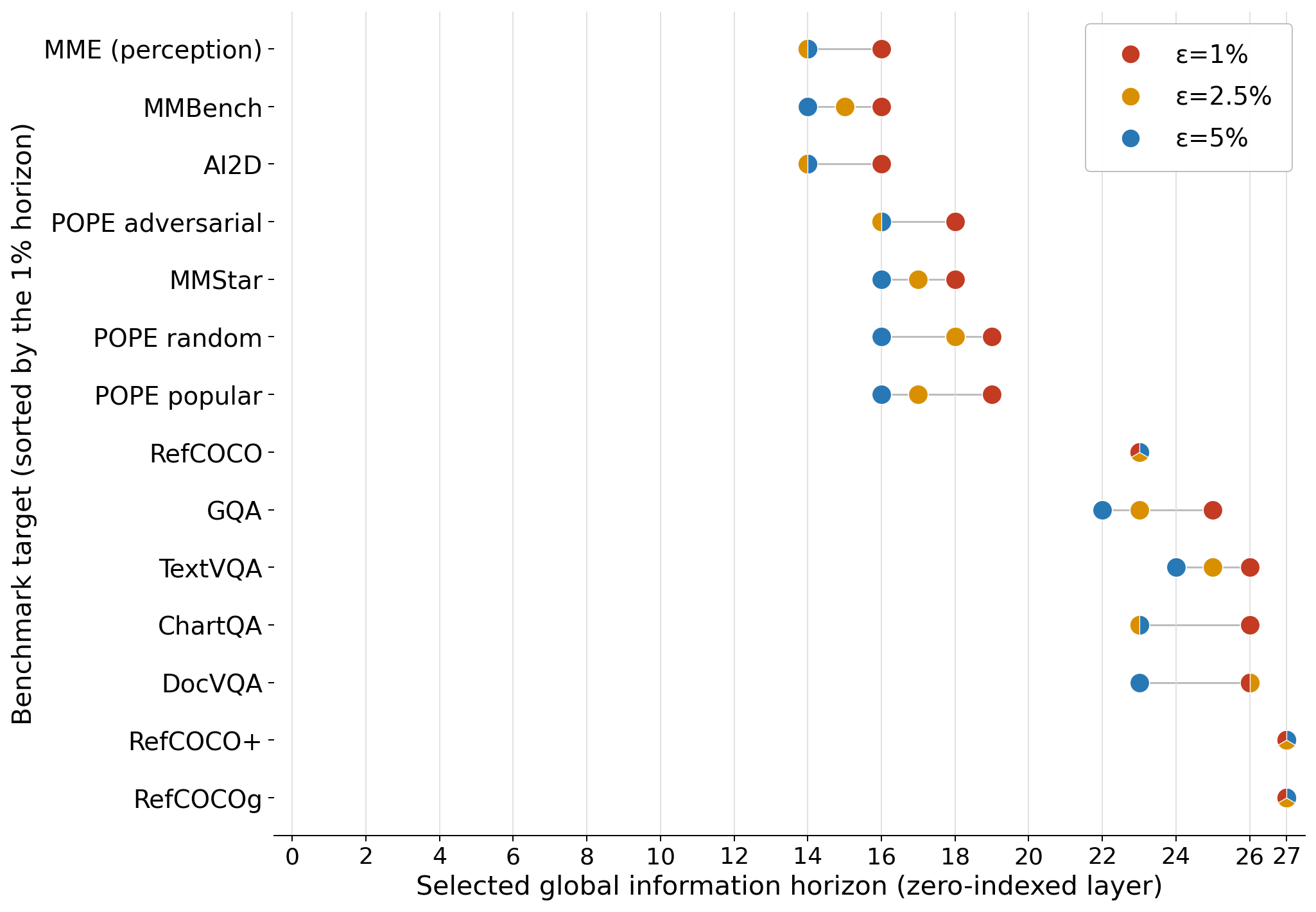}
        \caption{Qwen2.5-VL-7B}
        \label{fig:information_horizon_qwen}
    \end{subfigure}
    \caption{\textbf{Information horizon by benchmark and tolerance for LLaVA-1.5-7B and Qwen2.5-VL-7B.} Each marker indicates the earliest zero-indexed decoder layer after which all visual tokens can be removed while keeping the relative performance drop within $\epsilon\in\{1\%,2.5\%,5\%\}$ of the full-token baseline. Lower horizons indicate earlier opportunities to terminate visual-token processing. Horizons vary substantially across benchmarks and backbones: text-, chart-, and document-heavy tasks require late visual access in LLaVA, while grounding tasks reach the final decoder layer in Qwen. Relaxing the tolerance generally permits earlier withdrawal, but some tasks remain sensitive even at 5\%. This variation motivates adaptive selection of the visual-token withdrawal depth.}
    \label{fig:supp_information_horizons}
\end{figure*}

\section{Additional Experiments}
\label{sec:supp_additional_experiments}

These experiments examine what merging preserves, beyond the answer scores in the main text. Section~\ref{sec:supp_cross_boundary} examines how merging affects answers. Section~\ref{sec:supp_text_resolution} tests how averaging affects word reading and measures the text resolution retained across budgets.

\subsection{Cross-Boundary Merging Behaviour}
\label{sec:supp_cross_boundary}

% Generated by scripts/plot_controlled_merging.py; do not edit the numbers here.
\begin{table}[!tp]
    \centering
    \small
    \renewcommand{\arraystretch}{1.1}
    \caption{\textbf{Correct word readings after averaging.} The unmerged model reads 35/45 words correctly. Text and background interventions average equally many pairs per crop. ``All'' varies by crop. Mean replacement retains 576 tokens; mean-and-delete removes one token per pair. Exact match ignores case and non-alphanumeric characters. All 45 words enter each result.}
    \begin{tabular}{@{}lcrr@{}}
        \toprule
        Averaging operation & Pairs & Text & Background \\
        \midrule
        \multirow{3}{*}{Mean replacement} & 1 & 34/45 & 35/45 \\
         & 4 & 33/45 & 35/45 \\
         & All & 34/45 & 35/45 \\
        \midrule
        \multirow{3}{*}{Mean-and-delete} & 1 & 34/45 & 35/45 \\
         & 4 & 32/45 & 35/45 \\
         & All & 30/45 & 35/45 \\
        \bottomrule
    \end{tabular}
    \label{tab:supp_text_averaging}
\end{table}

\paragraph{Semantic compatibility and answering.}
Figure~\ref{fig:object_merging} compares three ways of merging highlighted image patches. Representing each person separately preserves the original answer, ``Blue,'' as does combining both people into one token. Including the snowboard in the same token changes the answer to ``White.'' In this example, merging across instances of the same category preserves the queried attribute, whereas mixing people with the snowboard disrupts it. To examine the feature complexity that each mean token must summarize, we visualize the token groups in a shared six-dimensional PCA (Figure~\ref{fig:object_merging_pca}). Adding another person changes the variance (also referred to as complexity) only slightly, whereas including the snowboard produces a larger increase. The people + snowboard group combines distinct semantic content, and its mean token fails to preserve the jacket-colour attribute needed for answering in Figure~\ref{fig:object_merging}.

\paragraph{Measuring object alignment.}
We evaluate tokenizer partitions on 3,811 referred objects from 1,500 RefCOCO validation images~\cite{yu2016refcoco}, before any decoder processing. Object polygons are rasterized on a padded $240\times240$ canvas and averaged into a $24\times24$ coverage map $c_i\in[0,1]$. We reconstruct each object using complete tokenizer regions: $u_i=1$ when the region containing cell $i$ has mean object coverage greater than 0.5, and zero otherwise. We then compute
\begin{equation}
    \operatorname{IoU}_{\mathrm{region}}
    = \frac{\sum_i \min(u_i,c_i)}
           {\sum_i \max(u_i,c_i)},
\end{equation}
averaged equally over objects. This fixed reconstruction rule measures how tokenizer regions match object masks.

\paragraph{Boundary guidance across budgets.}
Figure~\ref{fig:supp_partition_detail}(a) compares learned boundaries with CLIP-$B$, which replaces their spatial ordering using adjacent-cell CLIP cosine distances while preserving the boundary-value histogram (Section~\ref{sec:supp_spatial_information}). Textness and region counts are held fixed. At $K_m=32$, learned boundaries achieve 54.21\% region IoU, compared with 25.74\% for CLIP-$B$. At $K_m=224$, the corresponding values are 80.13\% and 76.28\%. The larger advantage under tight compression indicates that boundary guidance matters most when each region must summarize larger image areas.

\subsection{Text Resolution Across Budgets}
\label{sec:supp_text_resolution}

\paragraph{Controlled text-token averaging.}
Object boundaries alone do not protect the distinctions needed for reading. We evaluate 45 of 48 TextOCR~\cite{singh2021textocr} validation word crops from distinct TextVQA images, excluding three visually identified annotation errors. Crops are centred on $336\times336$ canvases without resizing. Frozen LLaVA-1.5-7B transcribes them with at most 16 output tokens. We average disjoint horizontal pairs of pre-projector features whose patches have at least 50\% word-polygon coverage. Controls average an equal number of background pairs outside the crop and its margin.

\emph{Mean replacement} writes each pair's mean into both original positions, keeping 576 tokens. \emph{Mean-and-delete} keeps one mean token and removes the other, also changing later positions. Table~\ref{tab:supp_text_averaging} separates these effects.

Figure~\ref{fig:supp_text_averaging} shows that averaging alone can change a unit or letter. Across the evaluated crops, averaging text features reduces correct word readings, while averaging equally many background pairs preserves the unmerged score (Table~\ref{tab:supp_text_averaging}).

\paragraph{Answer-text preservation across budgets.}
We mark patch cells touched by OCR boxes matching normalized answer words in 2,688 TextVQA questions from 2,044 images~\cite{singh2019textvqa} and 4,817 DocVQA questions from 1,260 images~\cite{mathew2021docvqa}. For each question, we measure the fraction kept in single-cell regions, then average over questions. Predicted and zero textness use identical boundaries and region counts.

Figure~\ref{fig:supp_partition_detail}(b--c) shows how textness guidance preserves finer text resolution as the budget grows. Its benefit is larger on TextVQA. On DocVQA, the curves remain close at low budgets. This agrees with the ablation studies in Section~\ref{sec:ablation}.
\section{Further Quantitative Results}
\label{sec:supp_quantitative_results}

We discuss the performance of \method{}~compared with other baselines per backbone in \ref{sec:supp_per_benchmark_performance}, as well as the predictive performance of the object and text localizers. 

\subsection{Per-Benchmark Performance}
\label{sec:supp_per_benchmark_performance}

\begin{figure*}
    \centering
    \includegraphics[width=1\linewidth]{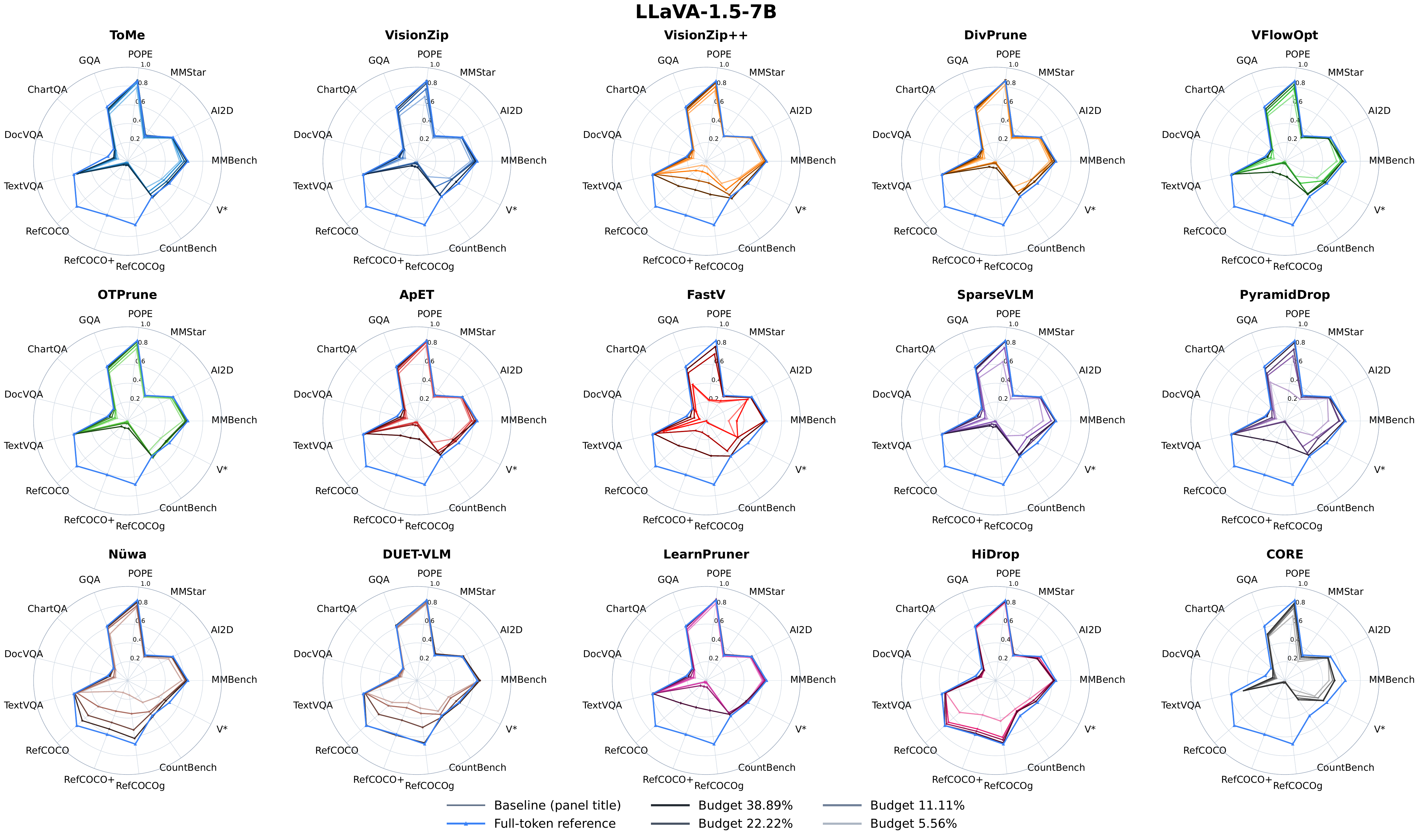}
    \caption{\textbf{Per-benchmark results for LLaVA-1.5 7B.}}
    \label{fig:llava_1_5_7b_all_baselines_radar}
\end{figure*}

\begin{figure*}
    \centering
    \includegraphics[width=1\linewidth]{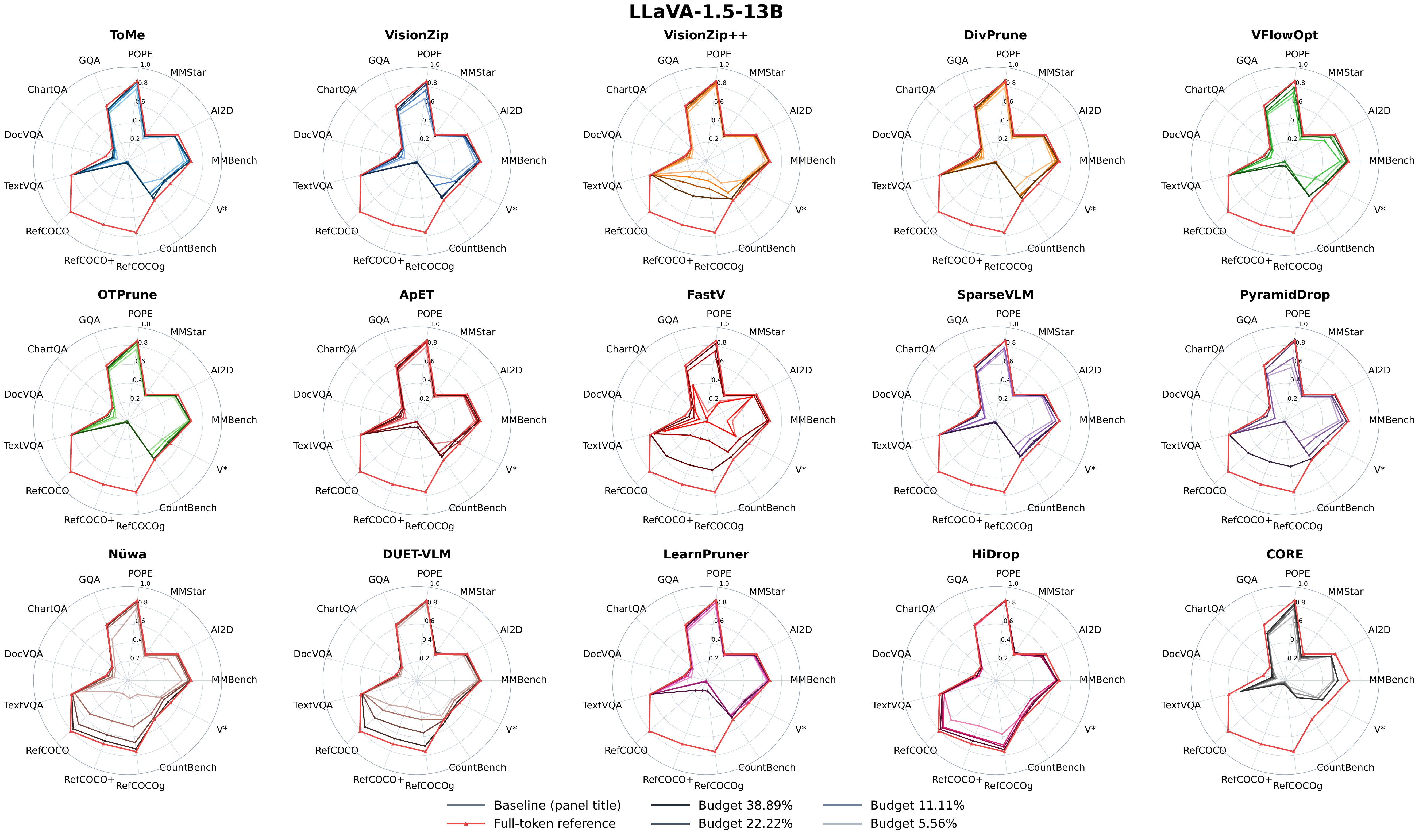}
    \caption{\textbf{Per-benchmark results for LLaVA-1.5 13B.}}
    \label{fig:llava_1_5_13b_all_baselines_radar}
\end{figure*}

\begin{figure*}
    \centering
    \includegraphics[width=1\linewidth]{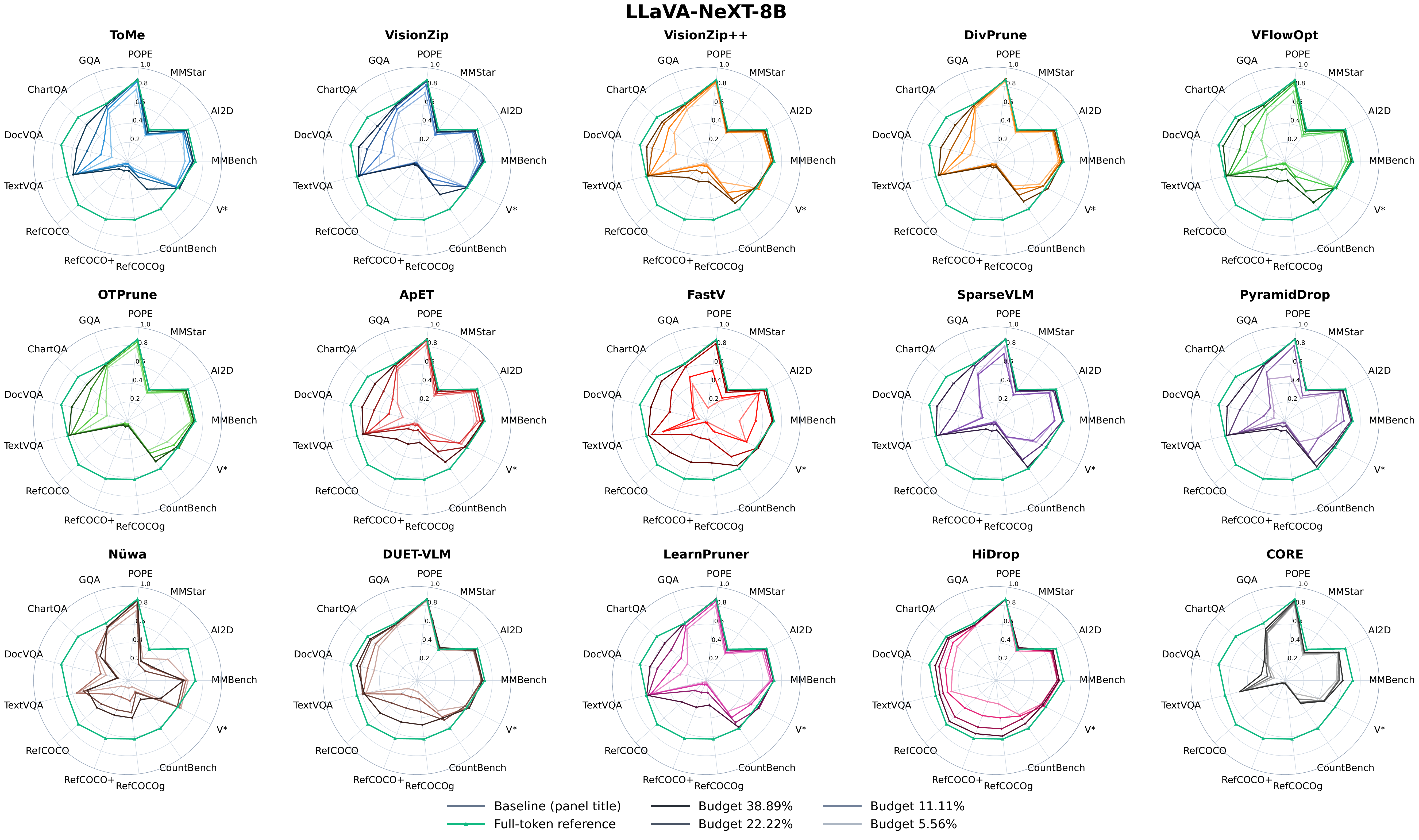}
    \caption{\textbf{Per-benchmark results for LLaVA-NeXT 8B.}}
    \label{fig:llava_next_8b_all_baselines_radar}
\end{figure*}

\begin{figure*}
    \centering
    \includegraphics[width=\linewidth]{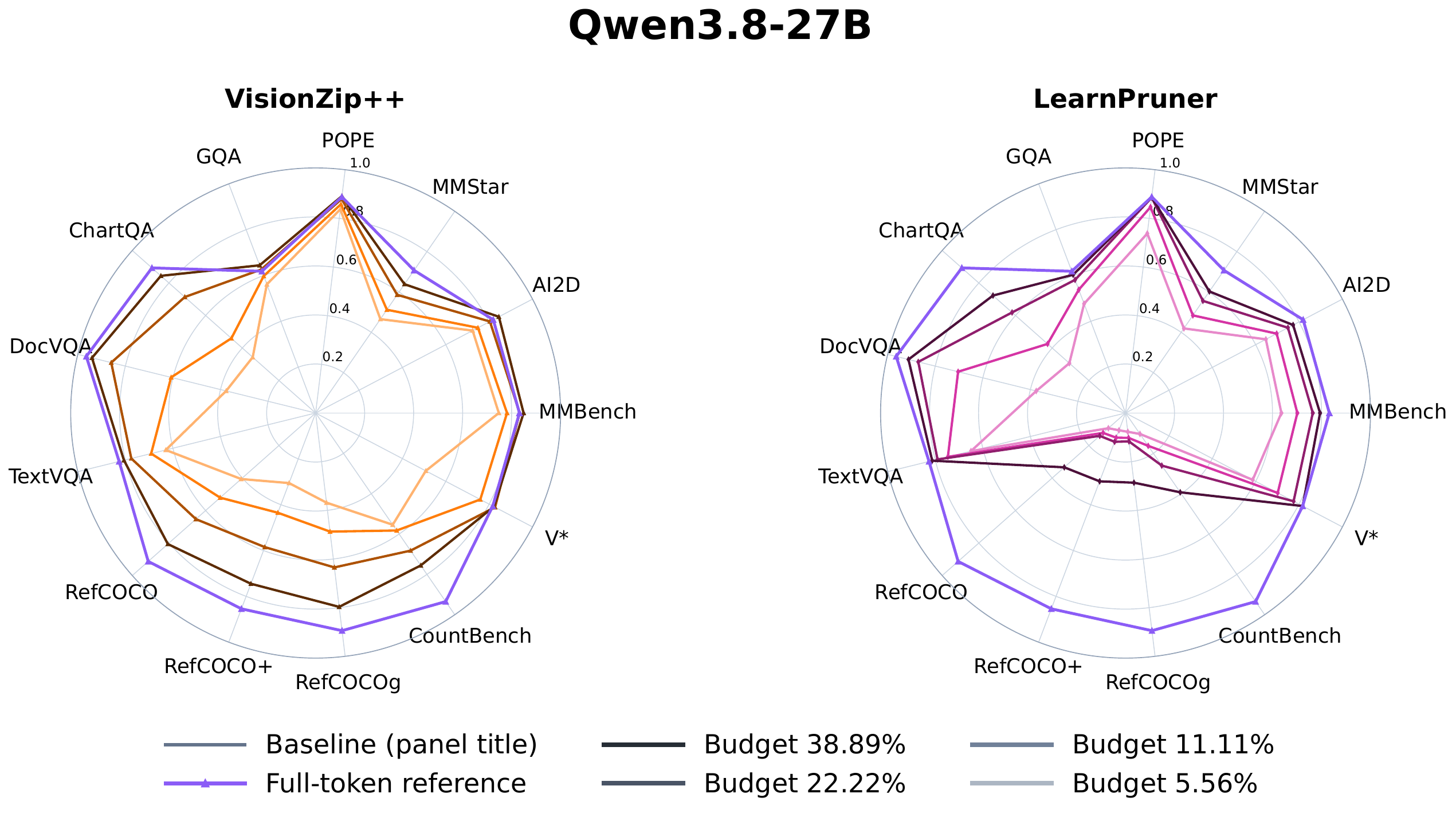}
    \caption{\textbf{Per-benchmark results for Qwen3.8-27B.}}
    \label{fig:qwen3_8_27b_all_baselines_radar}
\end{figure*}

We include radar charts for performance of each baseline for LLaVA-1.5 7B in Figure~\ref{fig:llava_1_5_7b_all_baselines_radar}, 13B in Figure~\ref{fig:llava_1_5_13b_all_baselines_radar}, LLaVA-NeXT 8B in Figure~\ref{fig:llava_next_8b_all_baselines_radar}, and Qwen3.8-27B in Figure~\ref{fig:qwen3_8_27b_all_baselines_radar}. Furthermore, we highlight the per-benchmark numeric results for the four baselines selected by the highest overall retention rate across backbones: LLaVA-1.5 7B in Table~\ref{tab:baselines_llava7b}, LLaVA-1.5 13B in Table~\ref{tab:baselines_llava13b}, and LLaVA-NeXT-8B in Table~\ref{tab:baselines_llavanext}. We also include per-benchmark results for the two baselines implemented on Qwen3.8-27B in Table \ref{tab:baselines_qwen38_27b}.

We count wins compared with the selected baselines in Tables~\ref{tab:baselines_llava7b}--\ref{tab:baselines_qwen38_27b}. On LLaVA-1.5-7B, \method{} has first place on 11 metrics, 10 at $K=64$, 10 at $K=128$, and 8 at $K=224$, summing to 39 in total. The runner-up, DUET-VLM, has 19 wins. On LLaVA-1.5-13B, \method{} has 7 at $K=32$, 11 at $K=64$, 14 at $K=128$, and 13 wins at $K=224$, totalling 45. The second-highest HiDrop has 10 first-place wins. On LLaVA-NeXT-8B, \method{} has 10 wins at $K=160$, 12 at $K=320$, 14 at $K=640$, and 11 at $K=1120$, totalling 47, compared to the next-best HiDrop at 6 wins. Finally, \method{} ranks first on Qwen3.8-27B in 14 metrics at $r=5.56\%$ and all 15 at all other ratios, totalling 59 of 60 wins across budgets; VisionZip++ is the runner-up with one win.

\subsection{Object-Localizer Accuracy}
\label{sec:supp_object_localizer_accuracy}

\begin{table}[t]
    \centering
    \small
    \caption{\textbf{Object-boundary accuracy on 1,493 COCO images from the object localizer's held-out validation set.} AP and ODS-F (\%) are computed over pooled grid edges, with ODS-F at the best single global threshold. Boundaries make up 8.2\% of edges on the native grid and 14.1\% on the tokenizer grid.}
    \begin{tabular}{llrr}
        \toprule
        Grid & Method & AP & ODS-F \\
        \midrule
        \multirow{3}{*}{Native 16-pixel} & Sobel gradient & 14.1 & 22.1 \\
         & Object localizer & 47.2 & 48.4 \\
         & DINOv3-L/16 & 47.3 & 49.2 \\
        \midrule
        \multirow{3}{*}{Tokenizer cells} & Sobel gradient & 20.8 & 30.1 \\
         & Object localizer & 51.0 & 51.1 \\
         & DINOv3-L/16 & 50.5 & 51.1 \\
        \bottomrule
    \end{tabular}
    \label{tab:object_localizer_accuracy}
\end{table}

We evaluate our object localizer accuracy against Sobel gradient and DINOv3 on LLaVA-1.5 in Table \ref{tab:object_localizer_accuracy}. The object localizer nearly matches its DINOv3-L/16 teacher: on the native 16-pixel grid it is within 0.1 AP and 0.8 ODS-F, and on the tokenizer cells, where the partition actually operates, it is on par (51.0 vs.\ 50.5 AP, identical 51.1 ODS-F). Both learned fields far outperform the Sobel gradient, by about 33 AP on the native grid and 30 AP on tokenizer cells, indicating that the boundaries reflect semantic object structure.

\subsection{Text-Localizer Accuracy}
\label{sec:supp_text_localizer_accuracy}

\begin{table}[t]
    \centering
    \small
    \caption{\textbf{Text localization on 1,432 held-out text-bearing images.} Labels come from OCR. AUC (\%) is averaged per image; AP and ODS-F (\%) pool cells across images.}
    \begin{tabular}{lrrr}
        \toprule
        Detector & AUC & AP & ODS-F \\
        \midrule
        Sobel edge density & 72.8 & 49.5 & 54.9 \\
        Text localizer (text-only version) & 97.1 & 96.2 & 89.7 \\
        Text localizer & 97.1 & 96.1 & 89.1 \\
        \bottomrule
    \end{tabular}
    \label{tab:text_localizer_accuracy}
\end{table}

\begin{table}[t]
    \centering
    \small
    \caption{\textbf{Text localizer at different thresholds.} A cell counts as text if $p_i>\tau$. All values are percentages. Recall and precision use the 1,432 text-bearing images. False cells are the share of cells detected as text, and false images are the share of images with at least one detected cell, on 1,568 text-free images.}
    \resizebox{\linewidth}{!}{%
        \begin{tabular}{rrrrrrr}
            \toprule
            & \multicolumn{4}{c}{Text localizer} & \multicolumn{2}{c}{Text-only version} \\
            \cmidrule(lr){2-5}\cmidrule(lr){6-7}
            $\tau(K_m)$ & Recall & Prec. & \begin{tabular}[b]{@{}c@{}}False\\cells\end{tabular} & \begin{tabular}[b]{@{}c@{}}False\\images\end{tabular} & Recall & \begin{tabular}[b]{@{}c@{}}False\\images\end{tabular} \\
            \midrule
            0.500 & 85.5 & 88.9 & 0.174 & 3.4 & 89.0 & 80.5 \\
            0.689 & 80.7 & 92.9 & 0.101 & 2.2 & 85.2 & 62.4 \\
            0.822 & 75.6 & 95.6 & 0.062 & 1.6 & 81.0 & 44.3 \\
            0.911 & 69.5 & 97.5 & 0.033 & 0.8 & 75.9 & 25.7 \\
            0.956 & 63.5 & 98.5 & 0.019 & 0.4 & 70.8 & 14.7 \\
            \bottomrule
        \end{tabular}%
    }
    \label{tab:text_localizer_thresholds}
\end{table}

In Table~\ref {tab:text_localizer_accuracy}, we compare the text localizer with a version trained only on images containing text and with Sobel edge density on LLaVA-1.5. We evaluate on a held-out set of 1,432 text-bearing images. Both the localizer and the text-only version far outperform Sobel edge density (97.1 vs.\ 72.8 AUC). We also evaluate text detection across thresholds in Table~\ref{tab:text_localizer_thresholds}. Training with text-free negatives sharply reduces false detections on text-free images. At the thresholds $\tau(K_m)\in\{0.689, 0.822, 0.911, 0.956\}$ for $K=224$ to $32$, only 0.4--2.2\% of text-free images receive any false text cell, against 14.7--62.4\% for the text-only version.
\section{Adaptive Halting}
\label{sec:supp_adaptive_halting}

Section~\ref{sec:supp_information_horizon} measures the information horizon of LLaVA-1.5-7B and Qwen2.5-VL-7B across benchmarks. Section~\ref{sec:supp_halting_outcomes} then examines where the adaptive rule triggers in \method{} on three backbones and how its decisions affect answer quality.

% Terry's original information-horizon section, retained from the older worktree.
% Only heading levels, a figure label, and the marked transitions are new.
\subsection{Information Horizon Analysis}
\label{sec:supp_information_horizon}

% Added transition: distinguish the reference experiment from the adaptive rule.
We first measure when complete visual-token withdrawal becomes tolerable on LLaVA-1.5-7B and Qwen2.5-VL-7B.

Wang et al.~\cite{wang2026horizon} observe that visual-token information
gradually vanishes with decoder depth and term the transition point the
\emph{information horizon}. Motivated by this observation, we study a
performance-based horizon under complete visual-token withdrawal. Let $A_0$
denote the full-token baseline score and $A_l$ the score obtained after
dropping all visual tokens following zero-indexed decoder layer $l$. We define
the relative performance drop as $\Delta(l)=\frac{A_0-A_l}{A_0}$, and the information horizon at tolerance $\epsilon$ as $h_{\epsilon}=\min\left\{l\mid\Delta(l)\leq\epsilon\right\}$.
Dropping after layer $l$ preserves visual-token access through layer $l$ and removes it from layers $l+1$ onward. Thus, a lower $h_{\epsilon}$ indicates that visual processing can terminate earlier without exceeding the prescribed performance loss.

We evaluate 12 benchmarks comprising 14 metrics, as POPE contributes separate random, popular, and adversarial F1 scores, on LLaVA-1.5-7B and Qwen2.5-VL-7B~\cite{qwen2025qwen25technicalreport}. We locate $h_{\epsilon}$ by binary search with batch size 32 for $\epsilon\in\{1\%,2.5\%,5\%\}$. On one NVIDIA RTX 6000 Ada Generation GPU with 48 GB of memory, the complete evaluation takes 41 hours, 27 minutes, and 17 seconds for LLaVA-1.5-7B, and 4 days, 14 hours, 53 minutes, and 32 seconds for Qwen2.5-VL-7B. Medians are computed over targets whose horizons occur within the evaluated decoder boundaries.

% POPE splits regenerated by section_d_analysis/recompute_halting_pope_splits.py using saved no-halting counts.
\begin{table*}[t]
    \centering
    \small
    \renewcommand{\arraystretch}{1.10}
    \caption{\textbf{Task-wise effects of adaptive halting.} Scores are changes from the same checkpoint without halting, averaged equally over four budgets. Pink and green indicate score decreases and increases, respectively, with a shared intensity scale across backbones. $h$ is the mean number of completed layers among inputs that halt. Saved is the mean per-input reduction in image-token processing summed across decoder layers, using recorded stage counts from the matching no-halting run for the three LLaVA backbones.}
    \resizebox{\linewidth}{!}{%
        \begin{tabular}{@{}lrrrr@{\hspace{10pt}}rrrr@{\hspace{10pt}}rrrr@{}}
            \toprule
            & \multicolumn{4}{c}{LLaVA-1.5-7B}
            & \multicolumn{4}{c}{LLaVA-1.5-13B}
            & \multicolumn{4}{c}{LLaVA-NeXT-8B} \\
            \cmidrule(lr){2-5}\cmidrule(lr){6-9}\cmidrule(lr){10-13}
            Task
            & \shortstack{$\Delta$ Score\\(pts)} & \shortstack{Halts\\(\%)} & \shortstack{$h$\\(layers)} & \shortstack{Saved\\(\%)}
            & \shortstack{$\Delta$ Score\\(pts)} & \shortstack{Halts\\(\%)} & \shortstack{$h$\\(layers)} & \shortstack{Saved\\(\%)}
            & \shortstack{$\Delta$ Score\\(pts)} & \shortstack{Halts\\(\%)} & \shortstack{$h$\\(layers)} & \shortstack{Saved\\(\%)} \\
            \midrule
            MMBench
            & \cellcolor{retPink!13.78}$-0.06$ & 99.7 & 20.5 & 14.6
            & \cellcolor{retGreen!14.41}$+0.07$ & 98.2 & 27.1 & 16.7
            & $+0.00$ & 98.6 & 24.2 & 9.6 \\
            AI2D
            & \cellcolor{retPink!12.52}$-0.04$ & 99.6 & 20.9 & 14.1
            & \cellcolor{retGreen!11.26}$+0.02$ & 94.4 & 32.9 & 7.8
            & \cellcolor{retPink!15.66}$-0.09$ & 93.3 & 24.5 & 8.8 \\
            MMStar
            & $+0.00$ & 96.4 & 22.5 & 11.8
            & \cellcolor{retGreen!15.03}$+0.08$ & 84.9 & 31.0 & 9.3
            & \cellcolor{retPink!11.89}$-0.03$ & 90.0 & 24.9 & 7.9 \\
            V$^*$
            & $+0.00$ & 99.7 & 21.0 & 13.9
            & \cellcolor{retPink!26.36}$-0.26$ & 96.5 & 26.4 & 17.3
            & \cellcolor{retGreen!18.18}$+0.13$ & 96.6 & 24.8 & 8.7 \\
            POPE-random
            & \cellcolor{retPink!11.26}$-0.02$ & 100.0 & 18.5 & 17.0
            & \cellcolor{retPink!50.91}$-0.65$ & 100.0 & 20.0 & 26.5
            & \cellcolor{retPink!15.03}$-0.08$ & 100.0 & 20.1 & 14.9 \\
            POPE-popular
            & \cellcolor{retPink!10.63}$-0.01$ & 100.0 & 18.6 & 16.9
            & \cellcolor{retPink!47.76}$-0.60$ & 100.0 & 19.9 & 26.3
            & \cellcolor{retPink!15.03}$-0.08$ & 100.0 & 20.0 & 15.0 \\
            POPE-adversarial
            & \cellcolor{retPink!11.26}$-0.02$ & 100.0 & 18.5 & 17.0
            & \cellcolor{retPink!29.51}$-0.31$ & 100.0 & 19.5 & 27.1
            & \cellcolor{retPink!13.15}$-0.05$ & 100.0 & 19.4 & 15.7 \\
            GQA
            & \cellcolor{retPink!20.70}$-0.17$ & 88.4 & 22.1 & 11.1
            & \cellcolor{retPink!27.62}$-0.28$ & 89.9 & 23.3 & 17.2
            & \cellcolor{retPink!15.03}$-0.08$ & 85.9 & 22.5 & 10.2 \\
            CountBench
            & \cellcolor{retPink!13.15}$-0.05$ & 26.5 & 25.2 & 2.3
            & \cellcolor{retGreen!16.29}$+0.10$ & 26.1 & 24.8 & 4.0
            & \cellcolor{retPink!22.59}$-0.20$ & 73.2 & 24.0 & 7.3 \\
            \midrule
            RefCOCO
            & \cellcolor{retPink!12.52}$-0.04$ & 22.9 & 25.4 & 1.9
            & \cellcolor{retPink!12.52}$-0.04$ & 98.1 & 34.2 & 5.8
            & $+0.00$ & 90.8 & 27.6 & 5.0 \\
            RefCOCO+
            & \cellcolor{retPink!11.26}$-0.02$ & 17.2 & 25.8 & 1.3
            & \cellcolor{retPink!11.89}$-0.03$ & 98.6 & 34.1 & 5.9
            & \cellcolor{retPink!11.26}$-0.02$ & 91.5 & 27.5 & 5.1 \\
            RefCOCOg
            & \cellcolor{retGreen!10.63}$+0.01$ & 25.4 & 25.5 & 2.1
            & \cellcolor{retGreen!10.63}$+0.01$ & 98.0 & 34.2 & 5.7
            & \cellcolor{retGreen!12.52}$+0.04$ & 94.4 & 27.5 & 5.3 \\
            \midrule
            TextVQA
            & \cellcolor{retPink!33.29}$-0.37$ & 75.3 & 23.8 & 7.8
            & \cellcolor{retPink!22.59}$-0.20$ & 69.7 & 26.8 & 9.8
            & \cellcolor{retPink!72.31}$-0.99$ & 64.2 & 24.9 & 5.8 \\
            DocVQA
            & \cellcolor{retPink!45.87}$-0.57$ & 67.1 & 25.0 & 5.9
            & \cellcolor{retPink!38.32}$-0.45$ & 53.7 & 27.6 & 6.9
            & \cellcolor{retPink!100.00}$-1.43$ & 56.2 & 25.8 & 4.3 \\
            ChartQA
            & \cellcolor{retPink!16.92}$-0.11$ & 76.7 & 25.3 & 6.5
            & \cellcolor{retPink!47.13}$-0.59$ & 60.5 & 28.0 & 7.8
            & \cellcolor{retPink!30.14}$-0.32$ & 69.0 & 25.3 & 5.8 \\
            \bottomrule
        \end{tabular}%
    }
    \label{tab:supp_halting_tasks}
\end{table*}

\subsubsection{LLaVA-1.5-7B}

On LLaVA-1.5-7B (see Figure \ref{fig:information_horizon_llava_7}), the strict 1\% horizon ranges from L10 for AI2D to L31 for ChartQA, with a median of L18 across the 14 metrics. Relaxing the tolerance moves the median horizon to L15.5 at both 2.5\% and 5\%. ChartQA and DocVQA still require late visual-token access at every tolerance. TextVQA and GQA are also relatively late under the strict tolerance. This pattern is consistent with text-, chart-, and document-heavy tasks requiring visual evidence deeper in the decoder. AI2D has the earliest horizon at all three tolerances. MMBench, MMStar, and the POPE variants also reach their horizons comparatively early. POPE has an abrupt change; performance across all three splits is impacted significantly at L13 but is within 1\% of the full-token at L14. The RefCOCO benchmarks are very similar, all at L18--L19 for 1\% and L17 otherwise.

\subsubsection{Qwen2.5-VL-7B}

The LLaVA results show high variance across tasks. We repeat the experiment on Qwen to test how the horizon shifts between backbones. As observed by Wang. et al~\cite{wang2026horizon}, in Qwen2.5-VL-7B (Figure~\ref{fig:information_horizon_qwen}) horizons are generally closer to the final layers than LLaVA. Specifically, we observe much later horizons for localization tasks, while text-heavy tasks remain similar. This motivates the need for adaptive halting, as the horizon per task varies greatly between VLMs. The earliest horizon is L16 at 1\% (MME perception~\cite{fu2025mmecomprehensiveevaluationbenchmark}, MMBench, AI2D) and L14 at 2.5\% and 5\%, while the latest is L27 (RefCOCO+, RefCOCOg) at every tolerance. Medians for 1\%, 2.5\%, and 5\% are 21, 20.5, and 19, respectively.

\subsection{Task-wise Halting Depth and Performance}
\label{sec:supp_halting_outcomes}

Table~\ref{tab:supp_halting_tasks} breaks down the main-text halting results by benchmark and backbone, listing the three POPE splits separately.

\section{Further Efficiency Analysis}
\label{sec:supp_efficiency}

We describe the efficiency measurement protocol in Section~\ref{sec:supp_efficiency_cases}, analyze computation and memory in Section~\ref{sec:supp_complexity}, and examine latency in Section~\ref{sec:supp_efficiency_latency}. Tables~\ref{tab:efficiency_7b}--\ref{tab:efficiency_next} report all methods at each budget. Table~\ref{tab:efficiency_qwen38_27b} reports implemented methods on Qwen3.8-27B.

\subsection{Efficiency Measurement Protocol}
\label{sec:supp_efficiency_cases}

We measure efficiency on 180 VQA evaluation samples spanning 151 images. These short-answer cases are drawn from AI2D, ChartQA, DocVQA, GQA, and TextVQA, organized into two workload sweeps. The first varies prompt length from 57 to 74 tokens over 16 levels, with reference answers of 1--11 tokens. The second varies reference-answer length from 1 to 21 tokens over 20 levels, with prompts of 59--63 tokens. Each level contains five cases. Counts are from LLaVA-1.5's tokenizer and prompt template.

Each case runs independently at batch size 1 on one NVIDIA H100, with greedy decoding capped at 128 new tokens. After three initial warm-up calls, each case is timed three times. Each timed call includes vision encoding, object and text localization, region construction and merging, and answer generation. FLOP counting and memory sampling use separate untimed passes. The tables report the mean of per-case median end-to-end latencies in milliseconds; cross-budget comparisons below equally average over the four budgets, and speedups are ratios of these means.

\subsection{Computational Complexity}
\label{sec:supp_complexity}

\begin{table}[t]
    \centering
    \small
    \caption{\textbf{Object-localizer cost.} The localizer and its DINOv3 teacher are measured on a $768\times576$ RGB input.}
    \begin{tabular}{lrr}
        \toprule
        Module & Params (M) & TFLOPs \\
        \midrule
        Object localizer & 7.03 & 0.0403 \\
        DINOv3-L/16 (teacher) & 303.13 & 1.3447 \\
        CLIP-L/14-336 & $\approx304$ & 0.3650 \\
        \bottomrule
    \end{tabular}
    \label{tab:object_localizer_cost}
\end{table}

Tables~\ref{tab:efficiency_7b}--\ref{tab:efficiency_qwen38_27b} report two FLOP counts. The theoretical count follows prior work~\cite{wu2026hidrop, xing2025pyramiddrop} and applies the same dense-attention formula to every VLM, including Qwen's hybrid decoder. It covers only image tokens during prefill:
\begin{equation}
    \mathrm{FLOPs} = \sum_{\ell=1}^{L} \left( 4 n_\ell d^2 + 2 n_\ell^2 d + 3 n_\ell d m \right),
\end{equation}
where $n_\ell$ is the number of image tokens entering decoder layer $\ell$, $d$ is the hidden size, and $m$ is the FFN intermediate size. For \method{}, $n_\ell=K_m$ through the selection layer, $n_\ell=K_p$ thereafter, and $n_\ell=0$ after an adaptive halt. We omit CORE due to substantial architectural differences.

The measured count uses PyTorch~\cite{Ansel_PyTorch_2_Faster_2024}'s FLOP counter mode for matrix multiplications, convolutions, and attention kernels. It includes the vision encoder, compression modules, prefill, and decoding at the actual KV-cache lengths, counting two FLOPs per multiply--accumulate. Elementwise operations and CPU control flow are excluded.

The text and object localizers cost 0.0016 and 0.0403 TFLOPs at the input sizes specified in Tables~\ref{tab:text_localizer_cost} and \ref{tab:object_localizer_cost}. Their combined cost is only 11.5\% of the CLIP encoder's 0.3650 TFLOPs and 1.0--2.3\% of \method{}'s measured total on LLaVA-1.5-7B. On LLaVA-NeXT, text localization covers both views, while object localization runs once on the full image. At one decoder layer, \texttt{[REL]} applies rank-256 projections to the mean register state and the image-token states, costing less than 0.001 TFLOPs on LLaVA-1.5.

For \method{}, measured counts are 2.2--12.6 times the theoretical counts because their scope and counting convention differ. From $K=224$ to $K=32$ on LLaVA-1.5-7B, theoretical cost falls by 6.9$\times$, whereas measured cost falls by 2.4$\times$. Relative to HiDrop at individual budgets, measured costs are 3.2\% lower to 1.3\% higher on LLaVA-1.5 and 3.2--8.3\% lower on NeXT. Averaging costs over budgets gives reductions of 1.2\%, 1.0\%, and 4.7\% on 7B, 13B, and NeXT, respectively. Adaptive halting further reduces theoretical cost by 4.2--9.6\% and measured cost by 1.0--6.7\%.

For memory, a background thread samples PyTorch's allocated CUDA memory every millisecond during a separate pass. We report the time-weighted mean, including model weights and runtime tensors. \method{}'s time-weighted mean memory is 2.4--4.4\% above the full-token model on 7B, 1.4--3.2\% on 13B, and 2.1--7.0\% on NeXT. This is expected because all visual-token compression methods do not affect decoding, the most time-consuming stage; thus, the additional localizer weights primarily increase the time-weighted mean, outweighing the savings during the prefill stage. Methods that do not introduce additional weights that constantly hold GPU memory do not have this caveat.

\subsection{Latency}
\label{sec:supp_efficiency_latency}

\begin{table}[t]
    \centering
    \small
    \caption{\textbf{Text-localizer cost.} Measured on a $672\times504$ grayscale input ($24\times18$ cells).}
    \begin{tabular}{lrr}
        \toprule
        Module & Params (M) & TFLOPs \\
        \midrule
        Text localizer & 0.92 & 0.0016 \\
        CLIP-L/14-336 & $\approx304$ & 0.3650 \\
        \bottomrule
    \end{tabular}
    \label{tab:text_localizer_cost}
\end{table}

Without adaptive halting, \method{} averages 80.8, 124.2, and 94.5 ms on LLaVA-1.5-7B, LLaVA-1.5-13B, and LLaVA-NeXT-8B, respectively. The corresponding full-token latencies are 104.1, 141.7, and 159.6 ms, giving speedups of 1.29$\times$, 1.14$\times$, and 1.69$\times$. Relative to HiDrop, the speedups are 1.66$\times$, 1.27$\times$, and 1.51$\times$. \method{} is the fastest among the baselines at three of four budgets on both 7B and NeXT.

On Qwen3.8-27B, \method{} averages 472.9 ms against 645.5 ms for the full-token model, a 1.36$\times$ speedup. It is the fastest at the two largest ratios. Averaged over ratios, it is
1.03$\times$ faster than VisionZip++ and within 2\% of LearnPruner, while achieving substantially higher results (Table~\ref{tab:efficiency_qwen38_27b}).

On 7B, reducing $K$ from 224 to 32 lowers \method{}'s measured computation by 58.0\% and latency by 3.6\%, from 82.6 to 79.6 ms. Adaptive halting further reduces measured FLOPs by 3.4\%, 5.1\%, and 3.0\% on 7B, 13B, and NeXT, respectively. Mean latency with halting is 89.4, 134.2, and 113.6 ms in the batch-size-1 evaluation. These compute savings may translate into latency gains at larger batch sizes.

\section{Further Ablation Studies}
\label{sec:supp_ablation_studies}

\begin{table*}[t]
    \centering
    \caption{\textbf{Per-benchmark effects of the selection policy.} Entries are score differences from the matched \method{} reference in each benchmark's native 0--100 scale, averaged over four token budgets. POPE-R, POPE-P, and POPE-A denote the random, popular, and adversarial splits, respectively. $^\dagger$ denotes an inference-time ablation of \method{}.}
    \resizebox{\linewidth}{!}{%
        \begin{tabular}{l*{15}{r}}
            \toprule
            Variant
            & TextVQA & DocVQA & ChartQA
            & RefCOCO & RefCOCO+ & RefCOCOg
            & CountBench
            & POPE-R & POPE-P & POPE-A
            & GQA & V* & AI2D & MMStar & MMBench \\
            \midrule
            \rowcolor{gray!12}
            \multicolumn{16}{c}{\textit{LLaVA-1.5-7B}} \\
            \midrule
            Random Selection$^\dagger$
            & \cellcolor{retPink!20.11}$-1.87$
            & \cellcolor{retPink!36.00}$-4.81$
            & \cellcolor{retPink!18.81}$-1.63$
            & \cellcolor{retPink!12.05}$-0.38$
            & \cellcolor{retPink!10.92}$-0.17$
            & \cellcolor{retPink!12.32}$-0.43$
            & \cellcolor{retPink!21.57}$-2.14$
            & \cellcolor{retPink!10.27}$-0.05$
            & \cellcolor{retPink!10.38}$-0.07$
            & \cellcolor{retPink!10.27}$-0.05$
            & \cellcolor{retPink!13.30}$-0.61$
            & $+0.00$
            & \cellcolor{retPink!10.27}$-0.05$
            & \cellcolor{retPink!10.11}$-0.02$
            & \cellcolor{retGreen!10.16}$+0.03$ \\
            Oracle Selection$^\dagger$
            & \cellcolor{retGreen!11.14}$+0.21$
            & \cellcolor{retGreen!14.05}$+0.75$
            & \cellcolor{retGreen!11.62}$+0.30$
            & \cellcolor{retGreen!11.03}$+0.19$
            & \cellcolor{retGreen!11.24}$+0.23$
            & \cellcolor{retPink!10.16}$-0.03$
            & $+0.00$
            & \cellcolor{retPink!10.22}$-0.04$
            & \cellcolor{retPink!10.11}$-0.02$
            & \cellcolor{retGreen!10.05}$+0.01$
            & \cellcolor{retGreen!11.08}$+0.20$
            & \cellcolor{retGreen!11.41}$+0.26$
            & \cellcolor{retPink!10.22}$-0.04$
            & \cellcolor{retGreen!10.16}$+0.03$
            & \cellcolor{retGreen!10.27}$+0.05$ \\
            \midrule
            \rowcolor{gray!12}
            \multicolumn{16}{c}{\textit{LLaVA-1.5-13B}} \\
            \midrule
            Random Selection$^\dagger$
            & \cellcolor{retPink!13.35}$-0.62$
            & \cellcolor{retPink!21.62}$-2.15$
            & \cellcolor{retPink!16.81}$-1.26$
            & \cellcolor{retPink!10.65}$-0.12$
            & \cellcolor{retPink!10.59}$-0.11$
            & \cellcolor{retGreen!10.49}$+0.09$
            & \cellcolor{retPink!12.76}$-0.51$
            & \cellcolor{retGreen!10.16}$+0.03$
            & \cellcolor{retPink!10.11}$-0.02$
            & \cellcolor{retGreen!10.11}$+0.02$
            & \cellcolor{retPink!11.73}$-0.32$
            & \cellcolor{retPink!11.41}$-0.26$
            & \cellcolor{retPink!10.11}$-0.02$
            & \cellcolor{retPink!10.27}$-0.05$
            & $+0.00$ \\
            Oracle Selection$^\dagger$
            & \cellcolor{retGreen!11.08}$+0.20$
            & \cellcolor{retGreen!12.70}$+0.50$
            & \cellcolor{retGreen!10.86}$+0.16$
            & \cellcolor{retPink!10.32}$-0.06$
            & \cellcolor{retPink!10.22}$-0.04$
            & $+0.00$
            & \cellcolor{retGreen!11.14}$+0.21$
            & \cellcolor{retGreen!10.11}$+0.02$
            & \cellcolor{retGreen!10.11}$+0.02$
            & $+0.00$
            & \cellcolor{retGreen!10.86}$+0.16$
            & \cellcolor{retPink!10.70}$-0.13$
            & \cellcolor{retGreen!10.11}$+0.02$
            & \cellcolor{retPink!10.22}$-0.04$
            & \cellcolor{retPink!10.27}$-0.05$ \\
            \midrule
            \rowcolor{gray!12}
            \multicolumn{16}{c}{\textit{LLaVA-NeXT-8B}} \\
            \midrule
            Random Selection$^\dagger$
            & \cellcolor{retPink!37.95}$-5.17$
            & \cellcolor{retPink!99.68}$-16.59$
            & \cellcolor{retPink!100.00}$-16.65$
            & \cellcolor{retPink!10.76}$-0.14$
            & \cellcolor{retPink!11.24}$-0.23$
            & \cellcolor{retGreen!10.70}$+0.13$
            & \cellcolor{retPink!15.51}$-1.02$
            & \cellcolor{retPink!10.16}$-0.03$
            & \cellcolor{retPink!10.11}$-0.02$
            & \cellcolor{retPink!10.05}$-0.01$
            & \cellcolor{retPink!13.41}$-0.63$
            & \cellcolor{retGreen!10.70}$+0.13$
            & \cellcolor{retGreen!10.05}$+0.01$
            & \cellcolor{retGreen!10.43}$+0.08$
            & \cellcolor{retGreen!10.43}$+0.08$ \\
            Oracle Selection$^\dagger$
            & \cellcolor{retGreen!12.54}$+0.47$
            & \cellcolor{retGreen!22.97}$+2.40$
            & \cellcolor{retGreen!16.00}$+1.11$
            & \cellcolor{retGreen!12.22}$+0.41$
            & \cellcolor{retGreen!11.62}$+0.30$
            & \cellcolor{retGreen!13.14}$+0.58$
            & \cellcolor{retPink!12.76}$-0.51$
            & \cellcolor{retGreen!10.05}$+0.01$
            & \cellcolor{retGreen!10.11}$+0.02$
            & \cellcolor{retGreen!10.05}$+0.01$
            & \cellcolor{retGreen!12.16}$+0.40$
            & \cellcolor{retGreen!10.70}$+0.13$
            & \cellcolor{retPink!10.22}$-0.04$
            & $+0.00$
            & \cellcolor{retGreen!10.11}$+0.02$ \\
            \bottomrule
        \end{tabular}%
    }
    \label{tab:supp_selection_tasks}
\end{table*}

We extend the main-text ablations to examine how localization,
budget allocation, and decoder-stage selection contribute to
\method{}. We first test the importance of image-specific spatial
guidance (Section~\ref{sec:supp_spatial_information}) and the
budget-dependent textness threshold
(Section~\ref{sec:supp_budget_dependent_text_threshold}).
We then examine selection policy and pruning depth
(Section~\ref{sec:supp_selection_analysis}).
We also test whether greater prompt sensitivity improves
answering and test the remaining dependence on specific image
regions across decoder depth
(Section~\ref{sec:supp_prompt_sensitivity}).
Finally, we evaluate position tags
(Section~\ref{sec:supp_ablation_studies_position_tags}) and joint adaptation
(Section~\ref{sec:supp_ablation_studies_training}).

\subsection{Spatial Information in the Localizer Fields}
\label{sec:supp_spatial_information}

\begin{table}[t]
    \centering
    \caption{\textbf{Sensitivity to the spatial organization of localizer fields.} Entries are intervention-minus-parent changes on native 0--100 score scales, averaged equally over 15 metrics and four budgets. Pink and green indicate decreases and increases, respectively, with intensity reflecting magnitude on a shared scale.}
    \resizebox{\linewidth}{!}{%
        \begin{tabular}{lrrrrrr}
            \toprule
            & \multicolumn{2}{c}{LLaVA-1.5-7B}
            & \multicolumn{2}{c}{LLaVA-1.5-13B}
            & \multicolumn{2}{c}{LLaVA-NeXT-8B} \\
            \cmidrule(lr){2-3}
            \cmidrule(lr){4-5}
            \cmidrule(lr){6-7}
            Intervention
            & $-$\texttt{[REL]} & \method
            & $-$\texttt{[REL]} & \method
            & $-$\texttt{[REL]} & \method \\
            \midrule
            $T=0$
            & \cellcolor{retPink!11.83}$-0.13$
            & \cellcolor{retPink!11.27}$-0.09$
            & \cellcolor{retPink!10.42}$-0.03$
            & \cellcolor{retPink!10.28}$-0.02$
            & \cellcolor{retGreen!10.84}$+0.06$
            & \cellcolor{retPink!10.28}$-0.02$ \\
            Shuffled $T$
            & \cellcolor{retPink!14.92}$-0.35$
            & \cellcolor{retPink!13.38}$-0.24$
            & \cellcolor{retPink!13.80}$-0.27$
            & \cellcolor{retPink!12.11}$-0.15$
            & \cellcolor{retPink!24.63}$-1.04$
            & \cellcolor{retPink!22.23}$-0.87$ \\
            CLIP-$B$
            & \cellcolor{retPink!48.11}$-2.71$
            & \cellcolor{retPink!37.56}$-1.96$
            & \cellcolor{retPink!50.50}$-2.88$
            & \cellcolor{retPink!33.20}$-1.65$
            & \cellcolor{retPink!83.55}$-5.23$
            & \cellcolor{retPink!50.78}$-2.90$ \\
            Shuffled $B$
            & \cellcolor{retPink!44.73}$-2.47$
            & \cellcolor{retPink!34.75}$-1.76$
            & \cellcolor{retPink!45.02}$-2.49$
            & \cellcolor{retPink!38.83}$-2.05$
            & \cellcolor{retPink!71.88}$-4.40$
            & \cellcolor{retPink!60.20}$-3.57$ \\
            Both
            & \cellcolor{retPink!49.80}$-2.83$
            & \cellcolor{retPink!38.69}$-2.04$
            & \cellcolor{retPink!50.92}$-2.91$
            & \cellcolor{retPink!34.75}$-1.76$
            & \cellcolor{retPink!100.00}$-6.40$
            & \cellcolor{retPink!64.00}$-3.84$ \\
            \bottomrule
        \end{tabular}%
    }
    \label{tab:supp_spatial_fields}
\end{table}

The main ablations establish the contribution of localization. Here, we test whether its semantic correspondence to the input image matters. We replace localization fields by retraining for each intervention and evaluate both the separately trained tokenizer-only model and joint \method{} on the three LLaVA backbones and four budgets.

\paragraph{Controlled replacements.}
CLIP-$B$, Zero $T$, and Both correspond to $-$Object Localizer, $-$Text Localizer, and No Localizer in Table~\ref{tab:ablation}. We use the following interventions:
\begin{itemize}
    \item \textbf{CLIP-$B$:} reassign the original image's boundary values to edges by the rank of adjacent-cell CLIP cosine distances, keeping the values but changing their spatial arrangement.
    \item \textbf{Zero $T$:} set all textness values to zero, removing both text-specific allocation and text-border reinforcement.
    \item \textbf{Shuffled $T$:} replace textness with another image's field.
    \item \textbf{Shuffled $B$:} replace boundaries with another image's.
    \item \textbf{Both:} combine CLIP-$B$ and $T=0$.
\end{itemize}

\paragraph{Shuffle construction.}
Donor images are assigned within each benchmark without self-assignment, using seeds 1 and 2 for textness and boundaries, respectively. We transfer fields by cell-center nearest-neighbour resampling, handling horizontal and vertical edges separately and preserving recipient padding boundaries. On NeXT, both the base and high-resolution views are modified. Unlike CLIP-$B$, these controls need not preserve the recipient's value histogram; they test the effect of guidance from an unrelated image.

\paragraph{Image-specific guidance.}
Table~\ref{tab:supp_spatial_fields} evaluates parents with the same protocol as ablations; thus, these references can differ slightly from the main-comparison scores. Shuffled textness causes larger aggregate losses than zero textness for both model variants on every backbone. On NeXT, for example, shuffled textness reduces the joint model's mean score by 0.87 points, compared with 0.02 points for zero textness. Thus, assigning text-specific resolution to incorrect locations can be more harmful than withholding that guidance. Both CLIP-$B$ and shuffled boundaries also reduce performance across all backbones. The CLIP-$B$ result shows that the boundary-value distribution alone is insufficient.

\paragraph{Budget dependence.}
Boundary guidance becomes particularly important under aggressive compression. On RefCOCO, CLIP-$B$ reduces the tokenizer-only model's score by 16.56 points at $K=32$ on LLaVA-1.5-13B, compared with 2.54 points at $K=224$. On NeXT, the corresponding losses are 32.20 points at $K=160$ and 0.16 points at $K=1120$. Incorrect spatial guidance can nevertheless remain harmful at larger budgets: shuffled boundaries still cost 5.32 points on NeXT RefCOCO at $K=1120$.

\paragraph{Comparison with $-$\texttt{[REL]}.}
\method{} incurs smaller aggregate losses than $-$\texttt{[REL]} under both shuffled controls and CLIP-based boundary replacement. This comparison evaluates re-trained designs with different token allocation schedules. Together, the interventions show that the trained models benefit from image-specific spatial guidance, beyond the presence of localization fields or their value statistics.

\subsection{Budget-Dependent Text Threshold}
\label{sec:supp_budget_dependent_text_threshold}

\begin{table}[t]
    \centering
    \small
    \caption{\textbf{Text-threshold comparison on LLaVA-1.5-7B.} Fixed uses $\tau=0.5$ at every budget; Budget-dependent uses the displayed threshold. Retention averages 15 full-token-normalized metrics. The mean weights all four budgets equally.}
    \begin{tabular}{rrrrr}
        \toprule
        & & \multicolumn{2}{c}{Retention (\%)} & \\
        \cmidrule(lr){3-4}
        $K$ & $\tau(K)$ & Fixed & Budget-dependent
        & $\Delta$ (pp) \\
        \midrule
        224 & 0.8056 & 100.34 & 101.75 & $+1.41$ \\
        128 & 0.8889 &  97.92 &  99.05 & $+1.13$ \\
        64 & 0.9444 &  92.83 &  95.26 & $+2.43$ \\
        32 & 0.9722 &  86.73 &  88.91 & $+2.18$ \\
        \midrule
        Mean & --- & 94.46 & 96.24 & $+1.79$ \\
        \bottomrule
    \end{tabular}%
    \label{tab:supp_text_threshold_retention}
\end{table}

We compare two separately trained text threshold configurations on LLaVA-1.5-7B: fixed $\tau=0.5$ and the budget-dependent schedule in \method. Both are tokenizer-only without \texttt{[REL]} and retain $K$ image tokens, so $K_m=K$ and
\begin{equation}
    \tau(K)=1-\frac{K}{1152}.
\end{equation}
At smaller budgets, the higher threshold requires greater confidence before assigning cells to the text group. Each configuration is trained with its corresponding threshold rule.

Table~\ref{tab:supp_text_threshold_retention} uses the supplementary
evaluation references described in Section~\ref{sec:supp_spatial_information}. The budget-dependent configuration improves mean retention at every budget by 1.13--2.43 points, increasing the four-budget average from 94.46\% to 96.24\%. The larger gains occur at $K=32$ and $K=64$, where allocating tokens to low-confidence text competes most strongly with preserving other image content.

Task-wise changes reveal the allocation tradeoff. Averaged across
budgets, RefCOCO, RefCOCO+, and RefCOCOg improve by 2.42, 2.49, and
2.21 points, respectively. DocVQA improves by 0.79 points, while TextVQA declines by 0.50 points. Together with the precision-recall tradeoff in Table~\ref{tab:text_localizer_thresholds}, these results support balancing text-specific resolution against the capacity available to other regions.

\subsection{\texttt{[REL]} Necessity Pruning}
\label{sec:supp_selection_analysis}

We examine the effects of the selection policy and pruning depth separately. The first comparison changes only token ranking within a fixed checkpoint; the second evaluates a separately trained variant with earlier pruning.

\subsubsection{Selection Policy at a Fixed Depth}

We compare \method{} with Random Selection and Oracle Selection at the same pruning layer and token counts. Random Selection samples $K_p$ image tokens uniformly. Oracle Selection gives the full-token model the correct answer and keeps the $K_p$ image tokens that the answer attends to most, averaged over all layers after the pruning layer. Because it needs the correct answer, it is not used at inference; we only use it as a reference for how much token selection can help.

Table~\ref{tab:supp_selection_tasks} reports task-level changes from matched supplementary evaluations. Random selection is particularly damaging on reading tasks. On LLaVA-NeXT-8B, its budget-averaged losses reach 5.17 points on TextVQA, 16.59 on DocVQA, and 16.65 on ChartQA. In contrast, POPE, MMStar, and MMBench change little on average. Thus, learned selection's contribution varies substantially across tasks.

Oracle gains are generally smaller, but the overall comparison hides task-specific differences. For example, on NeXT Oracle Selection improves DocVQA by 2.40 points on average, reaching 3.87 points at $K=160$. Thus, the limited headroom across metrics coexists with larger gains on some reading tasks. However, occasional negative differences also demonstrate that gold-answer attention does not guarantee a better retained set.

\begin{figure*}
    \centering
    \includegraphics[width=\linewidth]{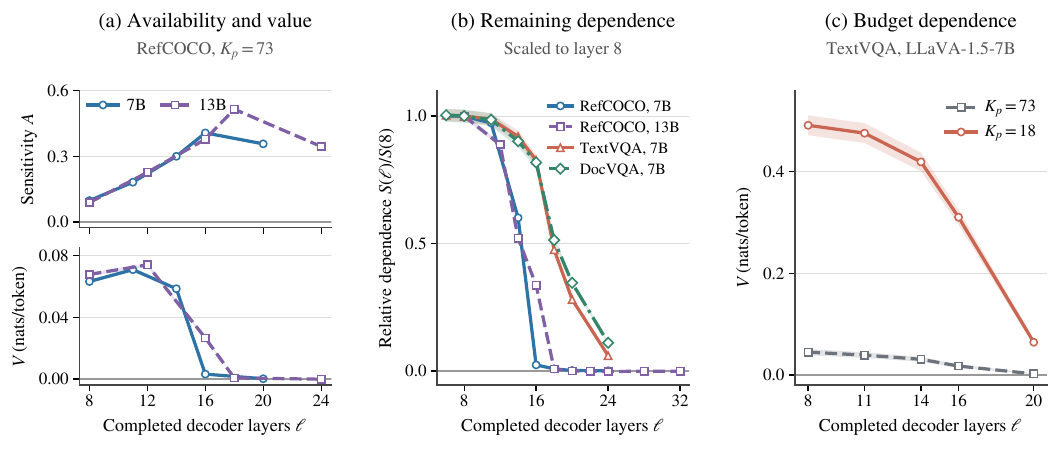}
    \caption{\textbf{Prompt-selection availability and remaining benefit vary with depth, task, and budget.} (a) On RefCOCO, selection sensitivity $A$ increases while the answer-informed advantage $V$ falls toward zero near layers 16 and 18 on 7B and 13B, respectively. $A$ measures changes in prompt-attention selections through one minus their mean pairwise Jaccard overlap. $V$ is reference-minus-oracle answer NLL, with the reference averaging oracle scores from other question groups on the same image and excluding the current referent on RefCOCO. (b) Deletion contrast $S$ compares masking target-region tokens with masking equally many control tokens. Scaling each curve by its measured $S(8)$ compares timing rather than effect size and
    reveals later dependence on text regions. (c) Tighter TextVQA retention increases $V$. All panels use $-$\texttt{[REL]} with $K_m=293$; panel (a) retains $K_p=73$. Markers denote measured means; bands show $\pm1$ standard error for $V$ and $S$, with the latter rescaled by the same fixed $S(8)$. NLL differences use teacher-forced gold answers and are in nats per answer token. Stage counts are fixed across depths, so layer-averaged computation varies.}
    \label{fig:supp_visual_dependence}
\end{figure*}

\subsubsection{Selection Depth at a Matched Budget}

\begin{table}[t]
    \centering
    \caption{\textbf{Pruning depth on LLaVA-1.5-7B at matched budgets.} Entries are native-score changes relative to midpoint \method{}, averaged over four budgets. $\Delta_{\mathrm{all}}$ averages 15 metrics from 13 benchmarks; $\Delta_{\mathrm{REC}}$ averages RefCOCO/+/g. The early oracle is evaluated on the same three grounding benchmarks. Stronger pink indicates a larger decrease on a shared scale.}
    \resizebox{\linewidth}{!}{%
        \begin{tabular}{lrrrr}
            \toprule
            Configuration & $\ell$ & $\rho$
            & $\Delta_{\mathrm{all}}$ & $\Delta_{\mathrm{REC}}$ \\
            \midrule
            \method{} & 16 & $1/4$
            & $0.00$ & $0.00$ \\
            Earlier pruning & 8 & $1/4$
            & \cellcolor{retPink!29.65}$-1.12$
            & \cellcolor{retPink!100.00}$-5.13$ \\
            Earlier pruning & 8 & $1/2$
            & \cellcolor{retPink!22.98}$-0.74$
            & \cellcolor{retPink!59.30}$-2.81$ \\
            Earlier pruning + Oracle & 8 & $1/4$
            & ---
            & \cellcolor{retPink!31.75}$-1.24$ \\
            \bottomrule
        \end{tabular}%
    }
    \label{tab:supp_selection_depth}
\end{table}

We evaluate an earlier-pruning variant trained to select after eight decoder layers on LLaVA-1.5-7B, compared with \method{} after sixteen layers. Let $\ell$ denote the number of completed layers before pruning. We choose $K_m$ and $K_p$ to match the layer-averaged budget, up to integer rounding:
\begin{equation}
    K=\frac{\ell K_m+(L-\ell)K_p}{L}.
\end{equation}
The earlier-pruning checkpoint is evaluated with target ratios $\rho=K_p/K_m\in\{1/4,1/2\}$. Matching $K$ controls the average number of image tokens processed per layer, while allowing their allocation across depth to change.

Earlier pruning reduces mean performance under both retention ratios (Table~\ref{tab:supp_selection_depth}), with larger losses on grounding. For $\rho=1/4$, Oracle Selection helps to close the gap, leaving a 1.24-point deficit relative to \method{} at the midpoint. Thus, selection quality matters more at this earlier boundary. These results support preserving broader visual access through the early decoder layers. Section~\ref{sec:supp_prompt_sensitivity} examines how this relates to prompt sensitivity and remaining visual reading.

\subsection{Prompt Sensitivity and Visual Dependence}
\label{sec:supp_prompt_sensitivity}

\begin{table}[t]
    \centering
    \caption{\textbf{Remaining dependence on the target region.} Entries report the paired deletion contrast $S(\ell)$ in nats per answer token, with standard errors over examples. Both interventions mask the same number of visual tokens. These inference-time diagnostics use $-$\texttt{[REL]} and the eligible cohorts described in the text.}
    \resizebox{\linewidth}{!}{%
        \begin{tabular}{lrrrr}
            \toprule
            & \multicolumn{3}{c}{LLaVA-1.5-7B}
            & LLaVA-1.5-13B \\
            \cmidrule(lr){2-4}
            \cmidrule(lr){5-5}
            $\ell$ & RefCOCO & TextVQA & DocVQA & RefCOCO \\
            \midrule
            8
            & $0.1045\pm0.0011$
            & $1.5175\pm0.0351$
            & $0.4964\pm0.0128$
            & $0.1137\pm0.0011$ \\
            14
            & $0.0628\pm0.0007$
            & $1.3961\pm0.0338$
            & $0.4475\pm0.0116$
            & $0.0592\pm0.0007$ \\
            16
            & $0.0025\pm0.0001$
            & $1.2547\pm0.0319$
            & $0.4062\pm0.0108$
            & $0.0383\pm0.0005$ \\
            18
            & $0.0009\pm0.0001$
            & $0.7224\pm0.0231$
            & $0.2549\pm0.0078$
            & $0.0009\pm0.0001$ \\
            20
            & $0.0002\pm0.0001$
            & $0.4260\pm0.0166$
            & $0.1717\pm0.0058$
            & $0.0000\pm0.0001$ \\
            24
            & $0.0001\pm0.0000$
            & $0.0916\pm0.0054$
            & $0.0551\pm0.0022$
            & $-0.0002\pm0.0000$ \\
            \bottomrule
        \end{tabular}%
    }
    \label{tab:supp_regional_dependence}
\end{table}

We examine whether changing the prompt changes visual selection, whether explicit supervision can increase this sensitivity, and whether doing so improves answering. We then distinguish when prompt-conditioned selection becomes available from when it remains useful, using attention-based probes, selection sweeps, and regional deletion across decoder depth. These experiments cover LLaVA-1.5-7B and LLaVA-1.5-13B, with adaptive halting disabled. As in Section~\ref{sec:supp_selection_analysis}, $\ell$ counts completed decoder layers.

\subsubsection{Sensitivity to Prompt Changes}

\begin{figure*}[t]
    \centering
    \includegraphics[width=\linewidth]{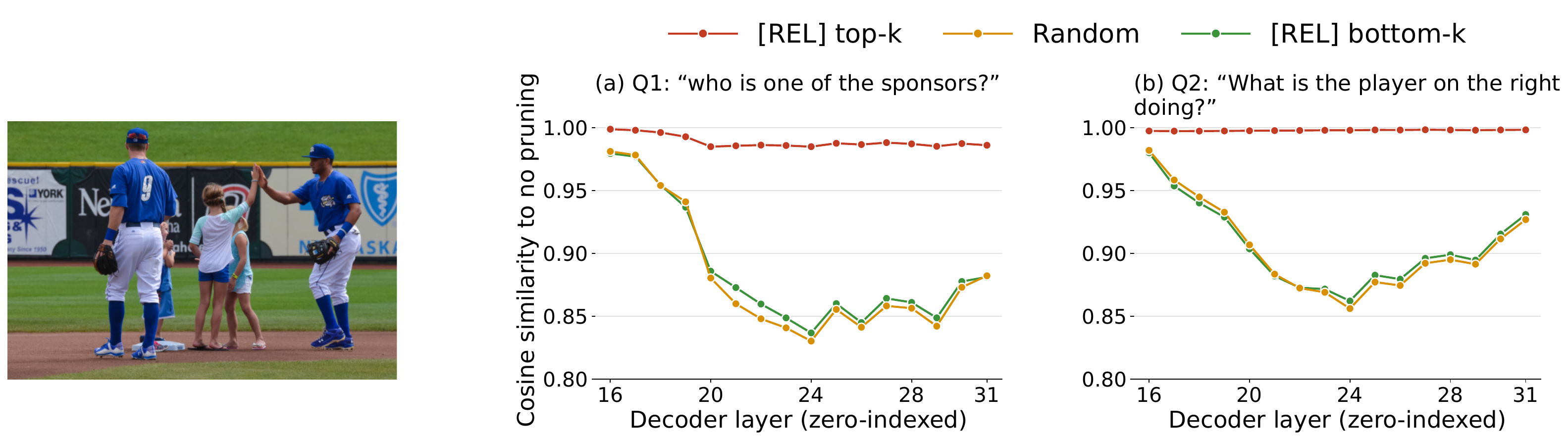}
    \caption{\textbf{Hidden state similarity comparison.} Cosine similarity of the final prompt token's hidden state between the unpruned model and different pruning strategies (\texttt{[REL]} top-k, random pruning, and \texttt{[REL]} bottom-k). Keeping the top $K_p$ tokens ranked by \texttt{[REL]} necessity score preserves the hidden state, while keeping $K_p$ random tokens or the bottom $K_p$ tokens does not ($K_p=26$).}
    \label{fig:supp_hidden_state_comparison}
\end{figure*}

We evaluate \method{} on LLaVA-1.5-7B at $K=64$, with $(K_m,K_p)=(102,26)$ and $\ell=16$. For two expressions naming
different objects $A$ and $B$ in the same image, we measure the fraction of cells inside $A$'s box whose regions survive pruning. The \emph{swap gain} is this fraction when the prompt names $A$ minus the fraction when it names $B$. We exclude box pairs with IoU at least 0.5, average ordered pairs within each image, and include 95\% confidence intervals using 2,000 image-level bootstrap resamples. Table~\ref{tab:supp_prompt_swaps} shows swap gains near zero and nearly identical \texttt{[REL]} rankings for different referents. Despite these nearly unchanged rankings, grounding accuracy reaches 67.54\% on RefCOCO and 62.66\% on RefCOCOg. The decoder therefore answers different prompts effectively even when the scores change little.

\begin{table}[t]
    \centering
    \caption{\textbf{Prompt sensitivity of \method{} on LLaVA-1.5-7B.} Score-rank correlations compare expressions naming the same or different referents. Accuracy is grounding accuracy at IoU $\geq0.5$ over each validation split.}
    \resizebox{\linewidth}{!}{%
        \begin{tabular}{llrrr}
            \toprule
            Split & Swap gain [95\% CI]
            & $r_{\mathrm{same}}$ & $r_{\mathrm{different}}$
            & Accuracy (\%) \\
            \midrule
            RefCOCO val
            & $-0.0002\;[-0.0005,+0.0002]$
            & 0.999 & 0.998 & 67.54 \\
            RefCOCOg val
            & $+0.0001\;[-0.0007,+0.0010]$
            & 0.997 & 0.997 & 62.66 \\
            \bottomrule
        \end{tabular}%
    }
    \label{tab:supp_prompt_swaps}
\end{table}

\subsubsection{Capacity for Prompt-Conditioned Selection}

To test whether this behaviour is imposed by the readout architecture, we freeze the 7B model and train only its scoring readout using RefCOCO training boxes. At $\ell=16$ and $(K_m,K_p)=(205,51)$, supervision encourages score probability mass on regions covering the named object. Training and held-out examples are separated by image.

Without changing the original readout architecture, two training seeds achieve swap gains of $+0.395$ and $+0.394$. Their highest-scoring token belongs to the target region in 88.5\% and 87.2\% of held-out examples, respectively. The architecture can indeed express prompt-conditioned rankings.

\subsubsection{Effect of Explicit Prompt Supervision}

We then refine the selector with region supervision from Visual Genome~\cite{krishna2016visualgenomeconnectinglanguage} referring expressions, GQA object links, and TextCaps~\cite{sidorov2020textcapsdatasetimagecaptioning} OCR boxes. The objective includes a 1.5-cell margin around target boxes and KL regularization toward the original selector on every prompt. This intervention produces substantial prompt sensitivity: RefCOCO swap gains range from $+0.268$ to $+0.284$ on 7B and from $+0.230$ to $+0.262$ on 13B across two seeds.

The downstream comparisons nevertheless show small aggregate changes (Table~\ref{tab:supp_prompt_refinement}). Averaged over four budgets and 15 metrics from 13 benchmarks, native scores change by $+0.00$ and $+0.02$ points on 7B and 13B, respectively. Therefore, increasing prompt sensitivity yields a negligible performance improvement in these refinement runs.

\begin{table}[t]
    \centering
    \caption{\textbf{Effect of prompt-supervised selector refinement.} Entries are native 0--100 score changes from the matched unmodified \method{} checkpoint, averaged equally over 15 metrics from 13 benchmarks. The final column averages all four budgets. The selector receives additional training with annotated object and text regions. Pink and green indicate decreases and increases, respectively.}
    \resizebox{\linewidth}{!}{%
        \begin{tabular}{lrrrrr}
            \toprule
            Backbone & $K=32$ & $K=64$ & $K=128$ & $K=224$ & Mean \\
            \midrule
            LLaVA-1.5-7B
            & \cellcolor{retGreen!82.00}$+0.08$
            & \cellcolor{retPink!46.00}$-0.04$
            & \cellcolor{retGreen!19.00}$+0.01$
            & \cellcolor{retPink!46.00}$-0.04$
            & $0.00$ \\
            LLaVA-1.5-13B
            & \cellcolor{retGreen!73.00}$+0.07$
            & $0.00$
            & \cellcolor{retGreen!55.00}$+0.05$
            & \cellcolor{retPink!37.00}$-0.03$
            & \cellcolor{retGreen!28.00}$+0.02$ \\
            \bottomrule
        \end{tabular}%
    }
    \label{tab:supp_prompt_refinement}
\end{table}

\begin{table*}[t]
    \centering
    \setlength{\tabcolsep}{3pt}
    \caption{\textbf{Per-benchmark position-tag ablation on LLaVA-1.5-7B.} The no-tag variant is retrained with the additive position tag disabled during both training and evaluation. Entries are scores on native 0--100 scales. Pink and green in the no-tag rows indicate decreases and increases relative to \method{} at the same budget, with intensity reflecting score differences on a shared scale.}
    \resizebox{\linewidth}{!}{%
        \begin{tabular}{l*{15}{r}}
            \toprule
            Variant & GQA & MMBench & MMStar & AI2D & POPE-R & POPE-P & POPE-A & V$^*$ & CountBench & RefCOCO & RefCOCO+ & RefCOCOg & TextVQA & DocVQA & ChartQA \\
            \midrule
            \rowcolor{gray!12}
            \multicolumn{16}{c}{$K=224$} \\
            \midrule
            \method{} & 61.97 & 66.46 & 34.33 & 55.44 & 86.57 & 85.43 & 83.69 & 42.93 & 51.73 & 75.75 & 66.52 & 71.90 & 58.99 & 22.49 & 19.68 \\
            No tag & \cellcolor{retGreen!11.72}62.07 & \cellcolor{retPink!17.39}66.03 & \cellcolor{retGreen!16.87}34.73 & \cellcolor{retPink!18.93}54.92 & \cellcolor{retPink!10.52}86.54 & \cellcolor{retGreen!16.48}85.81 & \cellcolor{retGreen!19.77}84.26 & \cellcolor{retGreen!63.93}46.07 & \cellcolor{retPink!17.04}51.32 & \cellcolor{retPink!21.51}75.08 & \cellcolor{retPink!14.12}66.28 & \cellcolor{retGreen!16.53}72.28 & \cellcolor{retPink!12.40}58.85 & \cellcolor{retPink!21.85}21.80 & \cellcolor{retPink!16.87}19.28 \\
            \midrule
            \rowcolor{gray!12}
            \multicolumn{16}{c}{$K=128$} \\
            \midrule
            \method{} & 61.75 & 66.10 & 33.93 & 54.99 & 87.11 & 86.22 & 84.32 & 42.93 & 51.73 & 72.09 & 63.41 & 68.24 & 58.44 & 21.36 & 19.60 \\
            No tag & \cellcolor{retPink!13.61}61.54 & \cellcolor{retPink!18.76}65.59 & \cellcolor{retGreen!34.05}35.33 & \cellcolor{retPink!16.70}54.60 & \cellcolor{retPink!20.19}86.52 & \cellcolor{retPink!19.18}85.69 & \cellcolor{retPink!13.42}84.12 & \cellcolor{retGreen!73.03}46.60 & \cellcolor{retPink!55.52}49.08 & \cellcolor{retPink!41.43}70.26 & \cellcolor{retPink!29.75}62.26 & \cellcolor{retPink!34.56}66.81 & \cellcolor{retPink!18.42}57.95 & \cellcolor{retPink!19.79}20.79 & \cellcolor{retPink!25.80}18.68 \\
            \midrule
            \rowcolor{gray!12}
            \multicolumn{16}{c}{$K=64$} \\
            \midrule
            \method{} & 60.98 & 64.24 & 33.60 & 55.28 & 86.94 & 85.99 & 83.77 & 45.03 & 49.90 & 67.54 & 58.97 & 62.66 & 57.37 & 18.64 & 18.08 \\
            No tag & \cellcolor{retPink!10.52}60.95 & \cellcolor{retPink!13.09}64.06 & \cellcolor{retGreen!40.92}35.40 & \cellcolor{retPink!21.68}54.60 & \cellcolor{retPink!29.30}85.82 & \cellcolor{retPink!25.54}85.09 & \cellcolor{retPink!18.59}83.27 & \cellcolor{retPink!28.03}43.98 & \cellcolor{retPink!48.47}47.66 & \cellcolor{retPink!56.89}64.81 & \cellcolor{retPink!56.72}56.25 & \cellcolor{retPink!47.10}60.50 & \cellcolor{retPink!20.13}56.78 & \cellcolor{retPink!11.89}18.53 & \cellcolor{retPink!18.24}17.60 \\
            \midrule
            \rowcolor{gray!12}
            \multicolumn{16}{c}{$K=32$} \\
            \midrule
            \method{} & 59.56 & 62.76 & 33.67 & 54.40 & 85.90 & 84.78 & 82.43 & 46.60 & 45.62 & 61.97 & 52.48 & 55.84 & 55.17 & 15.50 & 16.48 \\
            No tag & \cellcolor{retGreen!11.89}59.67 & \cellcolor{retGreen!18.24}63.24 & \cellcolor{retGreen!41.95}35.53 & \cellcolor{retPink!13.26}54.21 & \cellcolor{retPink!34.47}84.48 & \cellcolor{retPink!24.87}83.91 & \cellcolor{retPink!15.29}82.12 & \cellcolor{retPink!100.00}41.36 & \cellcolor{retGreen!41.43}47.45 & \cellcolor{retPink!71.49}58.39 & \cellcolor{retPink!59.12}49.62 & \cellcolor{retPink!66.16}52.57 & \cellcolor{retPink!14.81}54.89 & \cellcolor{retPink!13.78}15.28 & \cellcolor{retGreen!12.75}16.64 \\
            \bottomrule
        \end{tabular}%
    }
    \label{tab:supp_position_tag_ablation}
\end{table*}

\subsubsection{Availability and Value Across Decoder Depth}

We sweep the separately trained $-$\texttt{[REL]} models with $K_m=293$ and $K_p=73$. At boundary $\ell$, a prompt-conditioned selector $P$ ranks visual tokens by attention from the prompt, averaged over heads, prompt positions, and the three preceding layers. This probes the information in decoder attention, separately from the learned \texttt{[REL]} readout tested above. For an image $x$, let $\mathcal{T}_P^q(\ell)$ be the retained set for prompt $q$. We measure selection sensitivity as
\begin{equation}
    A(\ell)=1-\mathbb{E}_{x}\mathbb{E}_{q\ne q'}
    \frac{|\mathcal{T}_P^q(\ell)\cap\mathcal{T}_P^{q'}(\ell)|}
         {|\mathcal{T}_P^q(\ell)\cup\mathcal{T}_P^{q'}(\ell)|},
    \label{eq:supp_selection_availability}
\end{equation}
where $A$ measures how much the retained token set changes with the prompt. On RefCOCO, pairs name different objects in the same image. We take the average across the pairs within each image before averaging across images.

To measure the remaining benefit of question-specific selection, we compare an answer-attention oracle $O$ with a reference $B$ that averages oracle scores from other question groups on the same image. The oracle averages attention from gold-answer prediction positions over heads and layers from $\ell$ onward in an unpruned teacher-forced pass. RefCOCO groups are referred objects, so $B$ excludes all paraphrases of the current referent; elsewhere, each question is a group. The paired advantage is calculated by
\begin{equation}
    V(\ell)=\mathbb{E}\!\left[
        \operatorname{NLL}_{B}(\ell)-\operatorname{NLL}_{O}(\ell)
    \right],
    \label{eq:supp_selection_value}
\end{equation}
using the same gold-answer sequence for both interventions and only examples with another question group available. NLL is measured in nats per answer token. Both $B$ and $O$ are exposed to gold-answer information.

Figure~\ref{fig:supp_visual_dependence}(a) shows availability increasing as object-specific benefit declines. On 7B RefCOCO, $A$ rises from 0.098 at $\ell=8$ to 0.407 at $\ell=16$, while $V$ falls from 0.0633 to 0.0031. On 13B, $A$ reaches 0.515 at $\ell=18$, when $V$ is only 0.0006. A separate zero-shot probe on 300 RefCOCO images, each with two referred objects, tests grounding directly, selecting tokens using prompt attention at individual layers. Retained-cell coverage of the named box minus the comparison box rises from 0.023 at $\ell=8$ to 0.110 at $\ell=11$ on 7B, and from 0.026 to 0.098 between $\ell=8$ and $\ell=12$ on 13B. These probes indicate a relatively narrow overlap between appreciable prompt-to-region signals and substantial selection benefit, mainly at layers 11--14 on 7B and 12--16 on 13B. Weaker signals already occur at earlier layers.

Actual prompt-conditioned selection shows a similar pattern. Compared with attention from the fixed prompt ``Describe this image.'' averaged over the same three preceding layers, $P$ reduces RefCOCO answer NLL by 0.0161 and 0.0253 at $\ell=11$ and 14 on 7B, but only 0.0003 at $\ell=16$.

\subsubsection{Remaining Dependence on Specific Image Regions}

To test remaining regional access directly, we use the same
tokenizer-only ($-$\texttt{[REL]} in Table~\ref{tab:ablation}) models with $K_m=293$. Starting from an unpruned teacher-forced pass, we mask selected visual keys from boundary $\ell$ onward. Each example compares withdrawal of tokens covering the named object or answer text with withdrawal of an equal number of tokens covering another object or other text:
\begin{equation}
    S(\ell)=
    \mathbb{E}\!\left[
        \operatorname{NLL}_{\mathrm{target}}(\ell)
        -\operatorname{NLL}_{\mathrm{other}}(\ell)
    \right].
    \label{eq:supp_regional_dependence}
\end{equation}
NLL is measured in nats per answer token, using the same target sequence for both interventions. Positive $S$ indicates greater sensitivity to withdrawing the target region.

For RefCOCO, candidate tokens have at least half their cells inside the relevant box and are excluded if they also qualify for the comparison box. For reading tasks, normalized answer words are matched to OCR boxes; control tokens touch other text without touching answer text. Each side contributes the same number of tokens, ranked by regional coverage. The eligible cohorts contain 10,170 RefCOCO expressions, 2,518 TextVQA questions, and 4,817 DocVQA questions. The reading cohorts require a matching OCR word and at least one eligible token on each side.

Figure~\ref{fig:supp_visual_dependence}(b) shows RefCOCO dependence falling near layers 16 (7B) and 18 (13B), while reading dependence persists later (Table~\ref{tab:supp_regional_dependence}). Near-zero $S$ indicates similar average effects of target and control withdrawal on answer NLL.

\paragraph{Dependence on depth and budget.}
At $\ell=16$ and $K_p=73$ on 7B, $V$ is 0.0491 for DocVQA versus 0.0031 for RefCOCO. Reducing TextVQA's $K_p$ from 73 to 18 raises $V(16)$ from 0.0174 to 0.3107 as shown in Figure~\ref{fig:supp_visual_dependence}(c).

The observed narrow overlap of selection availability and benefit on grounding supports broad early visual access. Reading dependence persists later, and tighter budgets increase the remaining selection benefit.

\begin{table*}[t]
    \centering
    \small
    \setlength{\tabcolsep}{2pt}
    \caption{\textbf{Fine-tuning effects on the uncompressed model.} Scores of the original LLaVA-1.5-7B and Qwen3.8-27B backbones compared with after our joint adaptation recipe. The adaptation only slightly changes LLaVA, which means the major gains are attributed to the method. We also compare Qwen3.8-27B, which experiences larger gains attributed to fine-tuning on unseen data.}
    \resizebox{\linewidth}{!}{%
        \begin{tabular}{l*{15}{r}}
            \toprule
            Variant & GQA & MMBench & MMStar & AI2D & POPE-R & POPE-P & POPE-A & V$^*$ & CountBench & RefCOCO & RefCOCO+ & RefCOCOg & TextVQA & DocVQA & ChartQA \\
            \midrule
            \rowcolor{gray!12}
            \multicolumn{16}{c}{\textit{LLaVA-1.5-7B}} \\
            \midrule
            Vanilla & 61.89 & 64.11 & 32.87 & 54.63 & 87.47 & 86.29 & 84.37 & 50.26 & 45.62 & 72.27 & 61.32 & 68.08 & 58.83 & 21.50 & 19.92 \\
            Fine-tuned & 61.69 & 64.02 & 32.93 & 54.27 & 87.40 & 86.26 & 84.38 & 47.12 & 45.82 & 72.44 & 61.60 & 68.30 & 58.77 & 21.36 & 20.00 \\
            $\Delta$ & \cellcolor{retPink!15.73}$-0.20$ & \cellcolor{retPink!12.58}$-0.09$ & \cellcolor{retGreen!11.72}$+0.06$ & \cellcolor{retPink!20.32}$-0.36$ & \cellcolor{retPink!12.01}$-0.07$ & \cellcolor{retPink!10.86}$-0.03$ & \cellcolor{retGreen!10.29}$+0.01$ & \cellcolor{retPink!100.00}$-3.14$ & \cellcolor{retGreen!15.73}$+0.20$ & \cellcolor{retGreen!14.87}$+0.17$ & \cellcolor{retGreen!18.03}$+0.28$ & \cellcolor{retGreen!16.31}$+0.22$ & \cellcolor{retPink!11.72}$-0.06$ & \cellcolor{retPink!14.01}$-0.14$ & \cellcolor{retGreen!12.29}$+0.08$ \\
            \midrule
            \rowcolor{gray!12}
            \multicolumn{16}{c}{\textit{Qwen3.8-27B}} \\
            \midrule
            Vanilla & 61.85 & 83.21 & 70.73 & 81.90 & 90.53 & 88.74 & 87.23 & 81.68 & 93.48 & 91.31 & 85.44 & 89.46 & 82.61 & 96.42 & 89.28 \\
            Fine-tuned & 66.08 & 87.91 & 70.80 & 90.25 & 91.35 & 89.50 & 88.02 & 88.48 & 93.48 & 92.34 & 88.12 & 91.03 & 84.54 & 96.42 & 90.80 \\
            $\Delta$ & \cellcolor{retGreen!55.70}$+4.24$ & \cellcolor{retGreen!60.77}$+4.71$ & \cellcolor{retGreen!10.75}$+0.07$ & \cellcolor{retGreen!100.00}$+8.35$ & \cellcolor{retGreen!18.84}$+0.82$ & \cellcolor{retGreen!18.19}$+0.76$ & \cellcolor{retGreen!18.51}$+0.79$ & \cellcolor{retGreen!83.40}$+6.81$ & $0.00$ & \cellcolor{retGreen!21.10}$+1.03$ & \cellcolor{retGreen!38.89}$+2.68$ & \cellcolor{retGreen!26.92}$+1.57$ & \cellcolor{retGreen!30.80}$+1.93$ & $0.00$ & \cellcolor{retGreen!26.38}$+1.52$ \\
            \bottomrule
        \end{tabular}%
    }
    \label{tab:adaptation_ablation}
\end{table*}

\subsection{\texttt{[REL]} Hidden State Preservation}
\label{sec:supp_rel_hidden_state_preservation}

\begin{figure}[t]
    \centering
    \includegraphics[width=\linewidth]{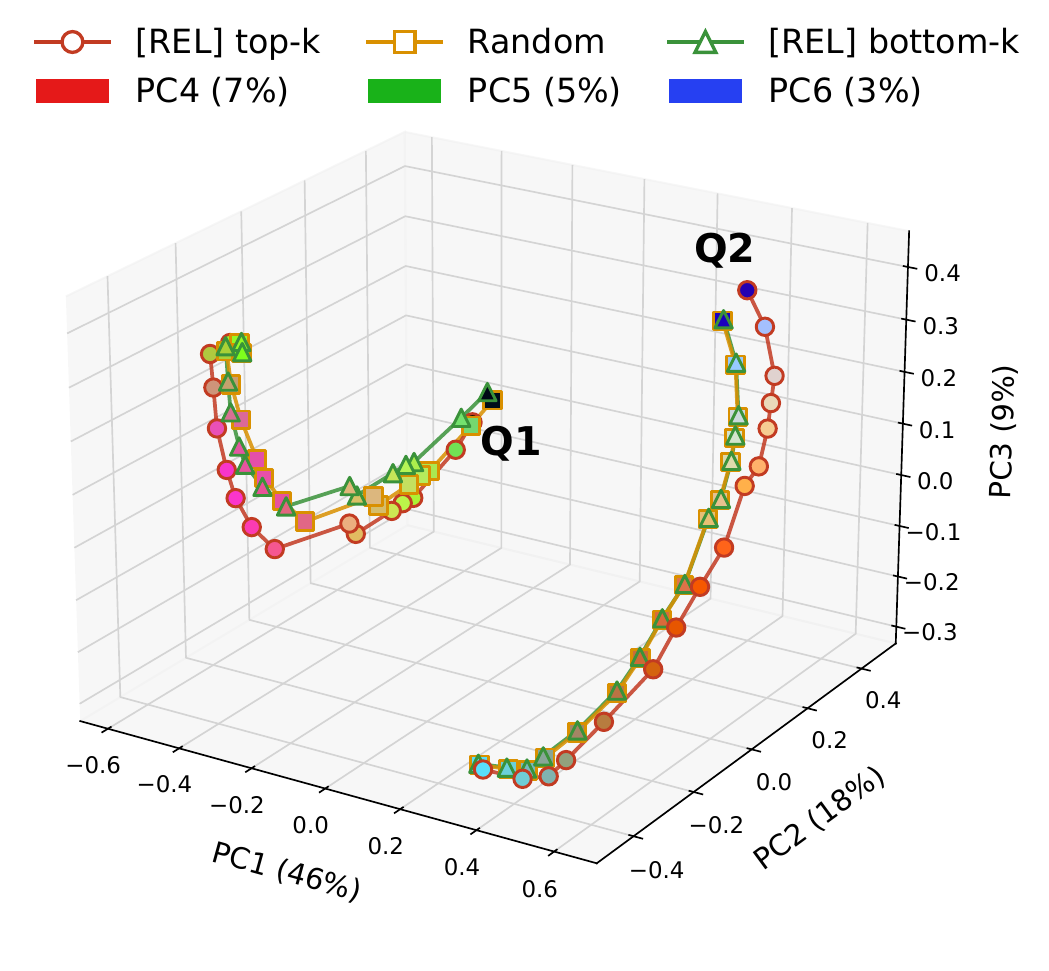}
    \caption{\textbf{PCA of final-token hidden states.} Hidden states of the final prompt token at layers 16--31 for two questions on the same image are projected onto the first six principal components of the unpruned model's states, explained variance in parentheses. As expected, the two questions show very different paths when forming the answers.}
    \label{fig:supp_question_pca}
\end{figure}

To understand why greater prompt sensitivity provides very limited headroom, we analyze the hidden-state representation across layers with and without \texttt{[REL]} necessity pruning. We run the experiment on LLaVA-1.5-7B at $K=64$, with $K_p=26$. As shown in Figure~\ref{fig:supp_hidden_state_comparison}, retaining the top $K_p$ tokens selected by \texttt{[REL]} leaves the hidden state nearly unchanged at every layer, while retaining randomly chosen tokens or the bottom $K_p$ tokens causes a substantial deviation. Different questions on the same image produce distinct representations, as shown by the PCA in Figure~\ref{fig:supp_question_pca}. Thus, we theorize that the selected tokens contain enough information to form the same hidden states per question. On the full 5000-sample TextVQA validation set, the average cosine similarity to the unpruned model is 0.995 under \texttt{[REL]} pruning, 0.967 under random pruning, and 0.930 when selecting the $K_p$ image tokens with the lowest \texttt{[REL]} necessity scores. Thus, we conclude that \texttt{[REL]} registers identify the general visual information required by later layers, allowing question-specific representations to be preserved.

\subsection{Position Tags}
\label{sec:supp_ablation_studies_position_tags}

To isolate the contribution of position tags, we retrain \method{} on LLaVA-1.5-7B with the additive position tag disabled during both training and evaluation. Table~\ref{tab:supp_position_tag_ablation} shows that removing the tags most consistently hurts grounding at tighter budgets. Averaged over RefCOCO, RefCOCO+, and RefCOCOg, scores fall by 3.24 points at $K=32$ and 2.54 points at $K=64$. TextVQA and DocVQA decline at every budget, whereas MMStar improves. Mean retention over all 15 metrics falls by about 1.2, 1.4, and 0.8 points at $K=32,64,128$, respectively, and is essentially unchanged at $K=224$. Together with what Figure~\ref{fig:supp_position_tag} shows qualitatively, the position tags are therefore necessary for preserving spatial information under aggressive compression.

\subsection{Joint Adaptation}
\label{sec:supp_ablation_studies_training}

\begin{table}[t]
    \centering
    \caption{\textbf{Calibrated benchmark-averaged retention rates on Qwen3.8-27B.} \method{}'s retention rates are with respect to the fine-tuned full-token in Table~\ref{tab:adaptation_ablation}. Both baselines' retention rates are still with respect to the vanilla full-token. Even under this slight disadvantage, \method{} still outperforms both baselines at every budget.}
    \resizebox{\linewidth}{!}{%
        \begin{tabular}{lrrrr}
            \toprule
            Method & $r=38.89\%$ & 22.22\% & 11.11\% & 5.56\%\\
            \midrule
            VisionZip++~\cite{yang2025visionzip} & \colorbox{lightgreen}{95.76\%} & \colorbox{lightgreen}{88.50\%} & \colorbox{lightgreen}{77.34\%} & \colorbox{lightgreen}{67.61\%}\\
            LearnPruner~\cite{takezoe2026learnpruner} & 79.43\% & 71.74\% & 64.16\% & 53.57\%\\
            \textbf{Varest (Ours)} & \colorbox{green}{\textbf{96.56\%}} & \colorbox{green}{\textbf{93.87\%}} & \colorbox{green}{\textbf{88.49\%}} & \colorbox{green}{\textbf{81.48\%}}\\
            \bottomrule
        \end{tabular}%
    }
    \label{tab:supp_average_retention_rates_qwen}
\end{table}

We investigate the role of fine-tuning in the performance gains of \method. We fine-tune the vanilla uncompressed LLaVA-1.5-7B backbone using our training procedure specified in Section \ref{sec:supp_training} for joint adaptation. Evaluation across 15 metrics from 13 benchmarks, counting the three POPE splits separately, shows a mean performance change of $-0.20$ points for the fine-tuned full-token model (Table~\ref{tab:adaptation_ablation}).

On the other hand, our adaptation protocol improves the uncompressed Qwen3.8-27B by 2.35 points on average. These strictly increasing gains are likely because the external mix665k dataset is better aligned with the benchmarks. Therefore, both the amount of data and the capacity of fine-tuning affect the magnitudes of these gains. We already force both baselines to train on the full instruction dataset to avoid an uneven amount of unseen data, so we only need to isolate the potential unfair advantage of more trainable parameters during adaptation. In Table~\ref{tab:supp_average_retention_rates_qwen}, we report the same retention rates from Table~\ref{tab:average_retention_rates_qwen}, but with \method{} calibrated against the fine-tuned full-token in Table~\ref{tab:adaptation_ablation}. This setup puts our method at a slight disadvantage because the baseline training protocols would also improve the full-token due to exposure to the same amount of unseen data. However, even under this comparison setup, \method{} still consistently outperforms the baselines. Overall, this shows that our leading performance on Qwen is not due to the training protocol, as with LLaVA.

\begin{table*}[p]
    \centering
    \setlength{\tabcolsep}{4pt}
    \caption{\textbf{Comparison with the selected baselines on LLaVA-1.5-7B.}}
    \resizebox{\linewidth}{!}{%
\begin{tabular}{lccccccccccccccc}
        \toprule
        \multirow{2}{*}{Method} & \multicolumn{4}{c}{General VQA} & \multicolumn{3}{c}{POPE} & \multicolumn{2}{c}{Fine-grained} & \multicolumn{3}{c}{Grounding} & \multicolumn{3}{c}{Text-rich} \\
        \cmidrule(lr){2-5}\cmidrule(lr){6-8}\cmidrule(lr){9-10}\cmidrule(lr){11-13}\cmidrule(lr){14-16}
        & GQA & MMB & MMStar & AI2D & Rand. & Pop. & Adv. & V$^*$ & Count & RefCOCO & RefCOCO+ & RefCOCOg & VQA$^\text{T}$ & DocVQA & ChartQA \\
        \midrule
        \rowcolor{gray!12}\multicolumn{16}{c}{\textit{Uncompressed, 576 Tokens (100\%)}} \\
        Full-Token~\cite{liu2024llava15} & 61.9 & 64.1 & 32.9 & 54.6 & 87.5 & 86.3 & 84.4 & 50.3 & 45.6 & 72.3 & 61.3 & 68.1 & 58.8 & 21.5 & 19.9 \\
        \midrule
        \rowcolor{gray!12}\multicolumn{16}{c}{\textit{Retain 224 Tokens ($\downarrow$61.1\%)}} \\
        VisionZip++~\cite{yang2025visionzip} {\scriptsize[CVPR'25]} & 60.0 & 63.4 & 32.6 & 55.1 & \colorbox{lightgreen}{86.8} & \colorbox{lightgreen}{85.8} & 83.5 & \colorbox{lightgreen}{48.2} & \colorbox{lightgreen}{47.9} & 39.6 & 32.5 & 35.5 & \colorbox{lightgreen}{58.4} & 20.6 & 19.6 \\
        N\"uwa~\cite{huang2026nuwa} {\scriptsize[ICLR'26]} & 61.3 & 62.9 & 32.1 & 54.0 & 86.3 & 85.3 & 83.2 & 46.1 & 46.8 & 64.4 & 55.1 & 62.1 & \colorbox{lightgreen}{58.4} & 20.2 & \colorbox{green}{\textbf{20.1}} \\
        DUET-VLM~\cite{singh2026duet} {\scriptsize[CVPR'26]} & \colorbox{green}{\textbf{62.6}} & \colorbox{green}{\textbf{66.8}} & \colorbox{lightgreen}{34.1} & \colorbox{lightgreen}{55.2} & \colorbox{green}{\textbf{87.3}} & \colorbox{green}{\textbf{86.5}} & \colorbox{green}{\textbf{84.5}} & \colorbox{green}{\textbf{51.3}} & 46.8 & 71.9 & \colorbox{lightgreen}{62.3} & \colorbox{lightgreen}{66.8} & 57.7 & \colorbox{lightgreen}{21.3} & \colorbox{lightgreen}{19.8} \\
\bottomrule
    \end{tabular}%
    }
    \label{tab:baselines_llava7b}
\end{table*}

\begin{table*}[p]
    \centering
    \setlength{\tabcolsep}{4pt}
    \caption{\textbf{Comparison with the selected baselines on LLaVA-1.5-13B.}}
    \resizebox{\linewidth}{!}{%
    \begin{tabular}{lccccccccccccccc}
        \toprule
        \multirow{2}{*}{Method} & \multicolumn{4}{c}{General VQA} & \multicolumn{3}{c}{POPE} & \multicolumn{2}{c}{Fine-grained} & \multicolumn{3}{c}{Grounding} & \multicolumn{3}{c}{Text-rich} \\
        \cmidrule(lr){2-5}\cmidrule(lr){6-8}\cmidrule(lr){9-10}\cmidrule(lr){11-13}\cmidrule(lr){14-16}
        & GQA & MMB & MMStar & AI2D & Rand. & Pop. & Adv. & V$^*$ & Count & RefCOCO & RefCOCO+ & RefCOCOg & VQA$^\text{T}$ & DocVQA & ChartQA \\
        \midrule
        \rowcolor{gray!12}\multicolumn{16}{c}{\textit{Uncompressed, 576 Tokens (100\%)}} \\
        Full-Token~\cite{liu2024llava15} & 63.3 & 67.7 & 34.1 & 60.5 & 87.3 & 86.4 & 84.6 & 51.3 & 50.1 & 81.0 & 72.2 & 76.1 & 61.6 & 23.5 & 21.6 \\
        \midrule
        \rowcolor{gray!12}\multicolumn{16}{c}{\textit{Retain 224 Tokens ($\downarrow$61.1\%)}} \\
        VisionZip++~\cite{yang2025visionzip} {\scriptsize[CVPR'25]} & 61.8 & 66.9 & 32.7 & 59.0 & 87.1 & \colorbox{lightgreen}{86.4} & 84.3 & 46.1 & 47.9 & 44.2 & 39.5 & 39.4 & 60.1 & \colorbox{lightgreen}{23.3} & 21.6 \\
        N\"uwa~\cite{huang2026nuwa} {\scriptsize[ICLR'26]} & 62.6 & \colorbox{green}{\textbf{67.9}} & 34.3 & \colorbox{lightgreen}{60.1} & 86.6 & 85.8 & 84.1 & \colorbox{lightgreen}{48.7} & 50.5 & 77.4 & \colorbox{lightgreen}{68.6} & 73.2 & \colorbox{lightgreen}{61.1} & 22.9 & \colorbox{lightgreen}{22.5} \\
        DUET-VLM~\cite{singh2026duet} {\scriptsize[CVPR'26]} & \colorbox{green}{\textbf{63.7}} & \colorbox{lightgreen}{67.8} & 35.6 & 58.7 & 87.1 & 86.1 & \colorbox{lightgreen}{84.5} & \colorbox{lightgreen}{48.7} & \colorbox{lightgreen}{53.0} & 74.2 & 66.1 & 70.3 & 60.3 & 22.7 & \colorbox{green}{\textbf{22.9}} \\
\bottomrule
    \end{tabular}%
    }
    \label{tab:baselines_llava13b}
\end{table*}

\begin{table*}[p]
    \centering
    \setlength{\tabcolsep}{4pt}
    \caption{\textbf{Comparison with the selected baselines on LLaVA-NeXT-8B.}}
    \resizebox{\linewidth}{!}{%
    \begin{tabular}{lccccccccccccccc}
        \toprule
        \multirow{2}{*}{Method} & \multicolumn{4}{c}{General VQA} & \multicolumn{3}{c}{POPE} & \multicolumn{2}{c}{Fine-grained} & \multicolumn{3}{c}{Grounding} & \multicolumn{3}{c}{Text-rich} \\
        \cmidrule(lr){2-5}\cmidrule(lr){6-8}\cmidrule(lr){9-10}\cmidrule(lr){11-13}\cmidrule(lr){14-16}
        & GQA & MMB & MMStar & AI2D & Rand. & Pop. & Adv. & V$^*$ & Count & RefCOCO & RefCOCO+ & RefCOCOg & VQA$^\text{T}$ & DocVQA & ChartQA \\
        \midrule
        \rowcolor{gray!12}\multicolumn{16}{c}{\textit{Uncompressed, 2880 Tokens (100\%)}} \\
        Full-Token~\cite{li2024llavanext-strong} & 65.2 & 72.0 & 40.4 & 72.5 & 88.2 & 87.4 & 86.1 & 60.2 & 61.7 & 70.0 & 65.9 & 62.7 & 65.7 & 72.6 & 70.5 \\
        \midrule
        \rowcolor{gray!12}\multicolumn{16}{c}{\textit{Retain 1120 Tokens ($\downarrow$61.1\%)}} \\
        VisionZip++~\cite{yang2025visionzip} {\scriptsize[CVPR'25]} & \colorbox{green}{\textbf{64.7}} & \colorbox{lightgreen}{71.4} & 39.4 & \colorbox{lightgreen}{70.5} & 88.3 & 87.3 & 85.9 & \colorbox{lightgreen}{59.2} & 54.0 & 25.9 & 22.9 & 21.7 & \colorbox{lightgreen}{64.5} & 65.2 & 63.0 \\
        N\"uwa~\cite{huang2026nuwa} {\scriptsize[ICLR'26]} & 60.4 & 59.6 & 25.1 & 30.3 & 88.3 & 87.4 & 85.4 & 40.3 & 24.2 & 43.6 & 39.5 & 40.1 & 44.0 & 11.6 & 38.7 \\
        DUET-VLM~\cite{singh2026duet} {\scriptsize[CVPR'26]} & 63.9 & 69.8 & \colorbox{lightgreen}{42.6} & 69.8 & \colorbox{lightgreen}{88.6} & \colorbox{green}{\textbf{88.0}} & \colorbox{lightgreen}{86.1} & \colorbox{green}{\textbf{62.8}} & 48.5 & 52.2 & 47.2 & 47.5 & 57.6 & 65.8 & 66.4 \\
\bottomrule
    \end{tabular}%
    }
    \label{tab:baselines_llavanext}
\end{table*}

\begin{table*}[p]
    \centering
    \setlength{\tabcolsep}{4pt}
    \caption{\textbf{Comparison with selected baselines on Qwen3.8-27B.}}
    \resizebox{\linewidth}{!}{%
    \begin{tabular}{lccccccccccccccc}
        \toprule
        \multirow{2}{*}{Method} & \multicolumn{4}{c}{General VQA} & \multicolumn{3}{c}{POPE} & \multicolumn{2}{c}{Fine-grained} & \multicolumn{3}{c}{Grounding} & \multicolumn{3}{c}{Text-rich} \\
        \cmidrule(lr){2-5}\cmidrule(lr){6-8}\cmidrule(lr){9-10}\cmidrule(lr){11-13}\cmidrule(lr){14-16}
        & GQA & MMB & MMStar & AI2D & Rand. & Pop. & Adv. & V$^*$ & Count & RefCOCO & RefCOCO+ & RefCOCOg & VQA$^\text{T}$ & DocVQA & ChartQA \\
        \midrule
        \rowcolor{gray!12}\multicolumn{16}{c}{\textit{Uncompressed, All Native Visual Tokens (100\%)}} \\
        Full-Token~\cite{qwen38} & 61.8 & 83.2 & 70.7 & 81.9 & 90.5 & 88.7 & 87.2 & 81.7 & 93.5 & 91.3 & 85.4 & 89.5 & 82.6 & 96.4 & 89.3 \\
        \midrule
        \rowcolor{gray!12}\multicolumn{16}{c}{\textit{Retain 38.89\% of Visual Tokens ($\downarrow$61.1\%)}} \\
        VisionZip++~\cite{yang2025visionzip} {\scriptsize[CVPR'25]} & \colorbox{lightgreen}{64.5} & \colorbox{lightgreen}{84.9} & \colorbox{lightgreen}{63.9} & \colorbox{lightgreen}{84.5} & \colorbox{lightgreen}{91.1} & \colorbox{lightgreen}{89.3} & \colorbox{lightgreen}{87.1} & \colorbox{lightgreen}{82.2} & \colorbox{lightgreen}{75.6} & \colorbox{lightgreen}{80.6} & \colorbox{lightgreen}{74.5} & \colorbox{lightgreen}{79.7} & 80.6 & \colorbox{lightgreen}{94.0} & \colorbox{lightgreen}{84.4} \\
        LearnPruner~\cite{takezoe2026learnpruner} {\scriptsize[ICLR'26]} & 60.3 & 79.4 & 60.3 & 77.3 & \colorbox{lightgreen}{91.1} & 88.3 & 86.6 & 81.7 & 39.3 & 33.4 & 29.8 & 28.6 & \colorbox{lightgreen}{81.3} & 91.3 & 72.3 \\
\bottomrule
    \end{tabular}%
    }
    \label{tab:baselines_qwen38_27b}
\end{table*}
\begin{table*}[!tp]
    \centering
    \caption{\textbf{Computational and runtime efficiency measurements on LLaVA-1.5-7B.} Green and pink indicate lower and higher costs, respectively, relative to the full-token baseline, with intensity reflecting the relative change (capped at 100\%).}
    \resizebox{\linewidth}{!}{%
        \begin{tabular}{l|rrrrrrr|rrrrrrr|rrr}
            \toprule
            $K$ & ToMe & VisionZip & VisionZip++ & DivPrune & VFlowOpt & OTPrune & ApET & FastV & SparseVLM & PyramidDrop & N\"uwa & DUET-VLM & LearnPruner & HiDrop & CORE & \textbf{\method} & \textbf{\method~+ AH}\\
            \midrule
            \rowcolor{gray!12}
            \multicolumn{18}{c}{\textit{Theoretical Complexity (TFLOPs)}}\\
            \midrule
            576 & \multicolumn{17}{c}{3.8172 (uncompressed)}\\
            \midrule
            224 & \effG{61.65}{1.4638} & \effG{61.65}{1.4638} & \effG{61.65}{1.4638} & \effG{61.65}{1.4638} & \effG{61.56}{1.4673} & \effG{61.65}{1.4638} & \effG{61.57}{1.4671} & \effG{61.66}{1.4636} & \effG{59.22}{1.5566} & \effG{61.34}{1.4758} & \effG{61.46}{1.4712} & \effG{61.55}{1.4679} & \effG{61.55}{1.4678} & \effG{61.30}{1.4773} & -- & \effG{62.64}{1.4261} & \effG{65.29}{1.3248}\\
            128 & \effG{78.17}{0.8332} & \effG{78.17}{0.8332} & \effG{78.17}{0.8332} & \effG{78.17}{0.8332} & \effG{78.16}{0.8335} & \effG{78.17}{0.8332} & \effG{78.14}{0.8343} & \effG{78.13}{0.8348} & \effG{75.00}{0.9542} & \effG{77.95}{0.8417} & \effG{78.11}{0.8356} & \effG{78.12}{0.8351} & \effG{78.18}{0.8329} & \effG{77.99}{0.8400} & -- & \effG{78.16}{0.8335} & \effG{79.69}{0.7753}\\
            64 & \effG{89.12}{0.4155} & \effG{89.12}{0.4155} & \effG{89.12}{0.4155} & \effG{89.12}{0.4155} & \effG{89.15}{0.4141} & \effG{89.12}{0.4155} & \effG{89.11}{0.4158} & \effG{88.93}{0.4224} & \effG{85.82}{0.5411} & \effG{88.99}{0.4203} & \effG{89.10}{0.4161} & \effG{89.14}{0.4144} & \effG{88.98}{0.4208} & \effG{88.96}{0.4214} & -- & \effG{89.10}{0.4159} & \effG{89.85}{0.3873}\\
\bottomrule
        \end{tabular}%
    }
    \label{tab:efficiency_7b}
\end{table*}

\begin{table*}[!tp]
    \centering
    \caption{\textbf{Computational and runtime efficiency measurements on LLaVA-1.5-13B.} Green and pink indicate lower and higher costs, respectively, relative to the full-token baseline, with intensity reflecting the relative change (capped at 100\%).}
    \resizebox{\linewidth}{!}{%
        \begin{tabular}{l|rrrrrrr|rrrrrrr|rrr}
            \toprule
            $K$ & ToMe & VisionZip & VisionZip++ & DivPrune & VFlowOpt & OTPrune & ApET & FastV & SparseVLM & PyramidDrop & N\"uwa & DUET-VLM & LearnPruner & HiDrop & CORE & \textbf{\method} & \textbf{\method~+ AH}\\
            \midrule
            \rowcolor{gray!12}
            \multicolumn{18}{c}{\textit{Theoretical Complexity (TFLOPs)}}\\
            \midrule
            576 & \multicolumn{17}{c}{7.4441 (uncompressed)}\\
            224 & \effG{61.55}{2.8626} & \effG{61.55}{2.8626} & \effG{61.55}{2.8626} & \effG{61.55}{2.8626} & \effG{61.56}{2.8616} & \effG{61.55}{2.8626} & \effG{61.48}{2.8678} & \effG{61.61}{2.8580} & \effG{59.57}{3.0100} & \effG{61.29}{2.8814} & \effG{61.39}{2.8742} & \effG{61.47}{2.8684} & \effG{61.46}{2.8689} & \effG{61.23}{2.8859} & -- & \effG{62.56}{2.7874} & \effG{66.16}{2.5194}\\
            128 & \effG{78.09}{1.6307} & \effG{78.09}{1.6307} & \effG{78.09}{1.6307} & \effG{78.09}{1.6307} & \effG{79.21}{1.5476} & \effG{78.09}{1.6307} & \effG{78.07}{1.6324} & \effG{78.01}{1.6373} & \effG{75.55}{1.8198} & \effG{77.92}{1.6440} & \effG{78.04}{1.6345} & \effG{78.05}{1.6337} & \effG{78.11}{1.6296} & \effG{77.91}{1.6443} & -- & \effG{78.10}{1.6306} & \effG{80.19}{1.4749}\\
            64 & \effG{89.07}{0.8137} & \effG{89.07}{0.8137} & \effG{89.07}{0.8137} & \effG{89.07}{0.8137} & \effG{89.11}{0.8105} & \effG{89.07}{0.8137} & \effG{89.06}{0.8141} & \effG{88.93}{0.8244} & \effG{86.37}{1.0144} & \effG{88.97}{0.8210} & \effG{89.06}{0.8146} & \effG{89.10}{0.8112} & \effG{88.93}{0.8238} & \effG{88.94}{0.8236} & -- & \effG{89.06}{0.8143} & \effG{90.10}{0.7366}\\
\bottomrule
        \end{tabular}%
    }
    \label{tab:efficiency_13b}
\end{table*}

\begin{table*}[!tp]
    \centering
    \caption{\textbf{Computational and runtime efficiency measurements on LLaVA-NeXT-8B.} Green and pink indicate lower and higher costs, respectively, relative to the full-token baseline, with intensity reflecting the relative change (capped at 100\%).}
    \resizebox{\linewidth}{!}{%
        \begin{tabular}{l|rrrrrrr|rrrrrrr|rrr}
            \toprule
            $K$ & ToMe & VisionZip & VisionZip++ & DivPrune & VFlowOpt & OTPrune & ApET & FastV & SparseVLM & PyramidDrop & N\"uwa & DUET-VLM & LearnPruner & HiDrop & CORE & \textbf{\method} & \textbf{\method~+ AH}\\
            \midrule
            \rowcolor{gray!12}
            \multicolumn{18}{c}{\textit{Theoretical Complexity (TFLOPs)}}\\
            \midrule
            2880 & \multicolumn{17}{c}{18.7535 (uncompressed)}\\
            \midrule
            1120 & \effG{62.84}{6.9690} & \effG{62.84}{6.9690} & \effG{61.71}{7.1802} & \effG{62.84}{6.9690} & \effG{62.70}{6.9957} & \effG{62.84}{6.9690} & \effG{62.56}{7.0212} & \effG{62.55}{7.0224} & \effG{62.44}{7.0430} & \effG{61.84}{7.1561} & \effG{62.10}{7.1077} & \effG{61.29}{7.2586} & \effG{62.49}{7.0353} & \effG{61.49}{7.2223} & -- & \effG{62.45}{7.0423} & \effG{64.25}{6.7042}\\
            640 & \effG{79.05}{3.9289} & \effG{79.05}{3.9289} & \effG{77.65}{4.1923} & \effG{79.05}{3.9289} & \effG{78.94}{3.9495} & \effG{79.05}{3.9289} & \effG{78.95}{3.9470} & \effG{78.59}{4.0142} & \effG{78.56}{4.0209} & \effG{78.61}{4.0113} & \effG{78.80}{3.9763} & \effG{77.47}{4.2245} & \effG{78.91}{3.9554} & \effG{77.96}{4.1328} & -- & \effG{78.92}{3.9528} & \effG{79.92}{3.7657}\\
            320 & \effG{89.62}{1.9461} & \effG{89.62}{1.9461} & \effG{88.05}{2.2408} & \effG{89.62}{1.9461} & \effG{89.62}{1.9461} & \effG{89.62}{1.9461} & \effG{89.60}{1.9509} & \effG{89.03}{2.0575} & \effG{89.03}{2.0581} & \effG{89.20}{2.0246} & \effG{89.55}{1.9589} & \effG{88.06}{2.2399} & \effG{89.58}{1.9544} & \effG{88.44}{2.1682} & -- & \effG{89.59}{1.9521} & \effG{90.06}{1.8639}\\
\bottomrule
        \end{tabular}%
    }
    \label{tab:efficiency_next}
\end{table*}

\begin{table*}
    \centering
    \footnotesize
    \setlength{\tabcolsep}{3pt}
    \caption{\textbf{Computational and runtime efficiency measurements on Qwen3.8-27B.} Green and pink indicate lower and higher costs, respectively, relative to the full-token baseline, with intensity reflecting the relative change (capped at 100\%).}
    \resizebox{\linewidth}{!}{%
        \begin{tabular}{l|rrr|rrr|rrr|rrr}
            \toprule
            \rowcolor{gray!12}
            Tokens retained & \multicolumn{3}{c}{\textit{Theoretical Complexity (TFLOPs)}} & \multicolumn{3}{c}{\textit{Measured Complexity (TFLOPs)}} & \multicolumn{3}{c}{\textit{Mean VRAM (MiB)}} & \multicolumn{3}{c}{\textit{Mean End-to-End Latency (ms)}} \\
            \cmidrule(lr){2-4}\cmidrule(lr){5-7}\cmidrule(lr){8-10}\cmidrule(lr){11-13}
            & VisionZip++ & LearnPruner & \textbf{\method} & VisionZip++ & LearnPruner & \textbf{\method} & VisionZip++ & LearnPruner & \textbf{\method} & VisionZip++ & LearnPruner & \textbf{\method} \\
            \midrule
            100\% (uncompressed) & \multicolumn{3}{c}{49.7716} & \multicolumn{3}{c}{118.9300} & \multicolumn{3}{c}{52448.2} & \multicolumn{3}{c}{645.5} \\
            \midrule
            38.89\% & \effG{63.82}{18.0097} & \effG{63.25}{18.2900} & \effG{63.28}{18.2762} & \effG{48.58}{61.1530} & \effG{48.54}{61.2016} & \effG{48.33}{61.4466} & \effP{0.12}{52513.6} & \effG{0.06}{52416.8} & \effP{2.24}{53622.2} & \effG{16.25}{540.6} & \effG{18.50}{526.1} & \effG{20.70}{511.9} \\
            22.22\% & \effG{79.76}{10.0730} & \effG{79.55}{10.1802} & \effG{79.64}{10.1326} & \effG{61.56}{45.7132} & \effG{61.67}{45.5830} & \effG{61.45}{45.8481} & \effP{0.08}{52490.8} & \effG{0.10}{52396.5} & \effP{1.79}{53389.3} & \effG{23.49}{493.9} & \effG{25.16}{483.1} & \effG{25.86}{478.6} \\
            11.11\% & \effG{90.04}{4.9566} & \effG{89.95}{5.0041} & \effG{90.07}{4.9420} & \effG{70.13}{35.5191} & \effG{70.32}{35.2967} & \effG{70.08}{35.5826} & \effP{0.06}{52478.8} & \effG{0.12}{52383.6} & \effP{1.47}{53221.7} & \effG{29.53}{454.9} & \effG{30.88}{446.2} & \effG{29.19}{457.1} \\
\bottomrule
        \end{tabular}%
    }
    \label{tab:efficiency_qwen38_27b}
\end{table*}

\begin{table*}[p]
    \centering
    \captionsetup{justification=raggedright,singlelinecheck=false,hypcap=true}
    \begingroup
    \small
    \setlength{\tabcolsep}{5pt}
    \setlength{\aboverulesep}{0pt}
    \setlength{\belowrulesep}{0pt}
    \renewcommand{\arraystretch}{1.00}
    \newcommand{\baselinecell}[1]{\raggedright #1}
    \newcommand{\trainingicon}[1]{\raisebox{-0.25\height}{\includegraphics[height=1.9em]{#1}}\hspace{0.2em}}
    \newcommand{\frozenicon}{\trainingicon{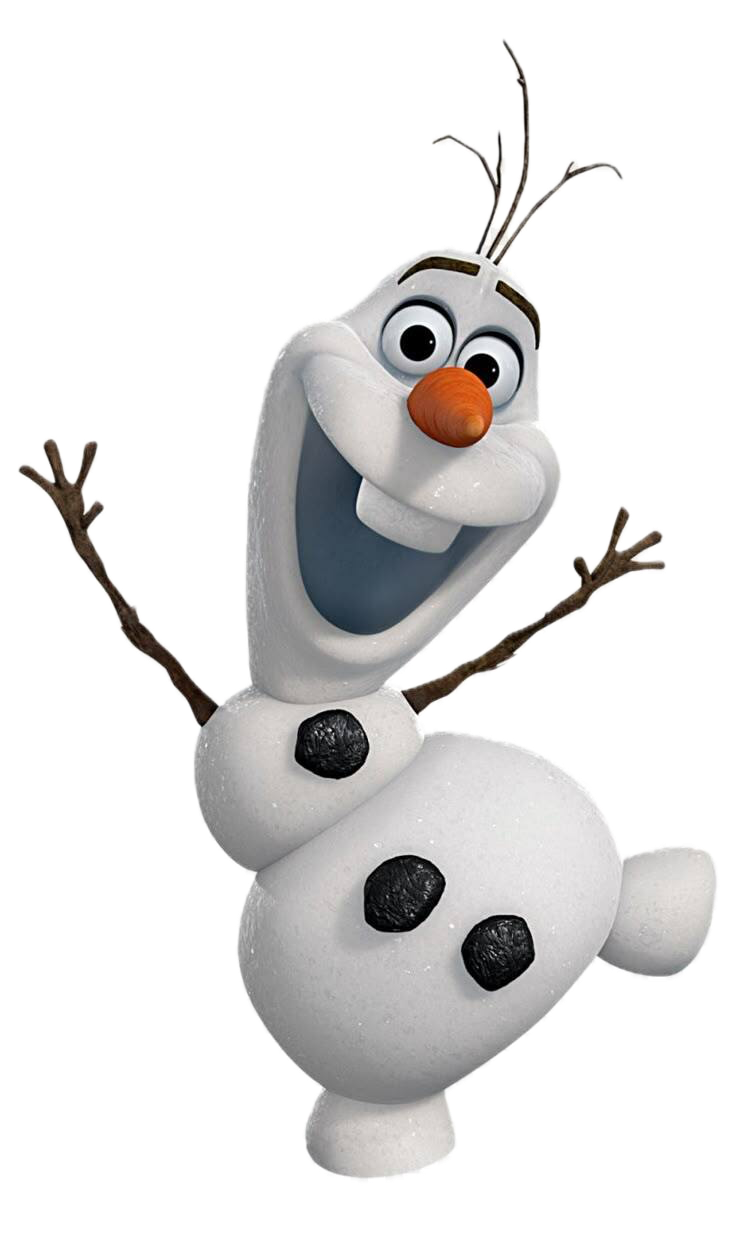}}
    \newcommand{\trainableicon}{\trainingicon{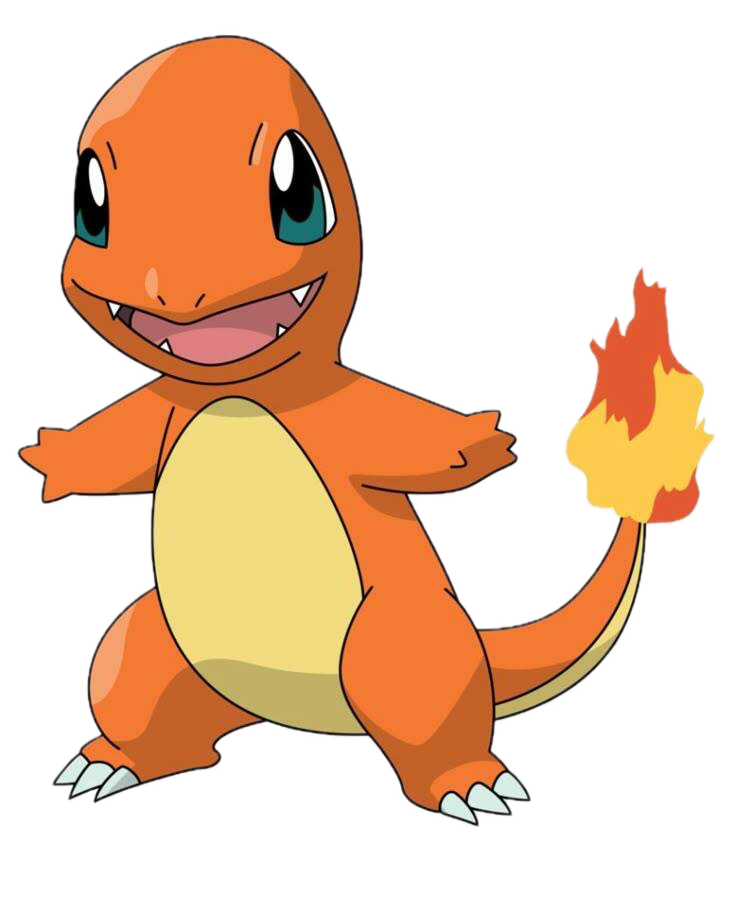}}
    \newcommand{\baselinerule}{\arrayrulecolor{black!20}\specialrule{0.3pt}{0pt}{0pt}\arrayrulecolor{black}}
    \newcommand{\headerstrut}{\rule[-0.5\baselineskip]{0pt}{1.5\baselineskip}}
    \caption{\textbf{Baseline training requirements and inference changes.}}
    \begin{tabular}{
        p{\dimexpr0.15\linewidth-2\tabcolsep\relax}
        p{\dimexpr0.34\linewidth-2\tabcolsep\relax}
        p{\dimexpr0.51\linewidth-2\tabcolsep\relax}
    }
        \toprule
        \rowcolor{orange}
        \baselinecell{\headerstrut\textbf{Method}} & \baselinecell{\textbf{Training Protocol}} & \baselinecell{\textbf{Changes in Our Implementation}} \tabularnewline
        \midrule
        % llmoe/compressor/tome/tome.py; tome_llava_next.py
        \baselinecell{ToMe~\cite{bolya2023tome}} & \baselinecell{\frozenicon None (training-free)} & \baselinecell{On NeXT, move merging after CLIP feature extraction so crop grids can be reconstructed and unpadded before compression. Merge within unpadded crop regions, order merged tokens by spatial centroid, and preserve row separators.} \tabularnewline
        \baselinerule
        % llmoe/compressor/vision_zip/vision_zip.py; vision_zip_llava_next_pp.py
        \rowcolor{lightorange!30}
        \baselinecell{VisionZip~\cite{yang2025visionzip}} & \baselinecell{\frozenicon None (training-free)} & \baselinecell{None.} \tabularnewline
        \baselinerule
        % docs/vision_zip_plus_plus_llava_next_8b.md; llmoe/compressor/vision_zip/vision_zip_ft.py
        \baselinecell{VisionZip++~\cite{yang2025visionzip}} & \baselinecell{\trainableicon Projector-only tuning on 10\% of instruction data, separately per budget; vision encoder and LLM frozen.} & \baselinecell{On NeXT, we exclude padding and row separators from selection and budget allocation to retain image-row structure. On Qwen3.8-27B, we train it on the entire mix-665k dataset since it is external, which is 4.18 percentage points (benchmark-averaged retention) higher on average than using a 10\% subset as the official recipe.} \tabularnewline
        \baselinerule
        % llmoe/compressor/divprune/divprune.py
        \rowcolor{lightorange!30}
        \baselinecell{DivPrune~\cite{alvar2025divprune}} & \baselinecell{\frozenicon None (training-free)} & \baselinecell{None.} \tabularnewline
        \baselinerule
        % llmoe/compressor/vflowopt/vflowopt.py; vflowopt_llava_next.py
        \baselinecell{VFlowOpt~\cite{yang2025vflowopt}} & \baselinecell{\frozenicon None (training-free); pruning-policy search on calibration data.} & \baselinecell{Restore high-ranked tokens from recycling when needed to meet target budget.} \tabularnewline
        \baselinerule
        % llmoe/compressor/otprune/otprune.py
        \rowcolor{lightorange!30}
        \baselinecell{OTPrune~\cite{chen2026otprune}} & \baselinecell{\frozenicon None (training-free)} & \baselinecell{None.} \tabularnewline
        \baselinerule
        % llmoe/compressor/apet/apet.py; apet_llava_next.py
        \baselinecell{ApET~\cite{ma2026apet}} & \baselinecell{\frozenicon None (training-free)} & \baselinecell{On NeXT, replace per-crop retention with an image-wide budget allocation across crop patches.} \tabularnewline
        \baselinerule
        % llmoe/compressor/fastv/fastv.py
        \rowcolor{lightorange!30}
        \baselinecell{FastV~\cite{chen2024fastv}} & \baselinecell{\frozenicon None (training-free)} & \baselinecell{Since the original FastV paper uses post-pruning token counts, recompute retention to match layer-averaged budgets. On LLaVA-1.5 and NeXT at 32/5.56\%, advance the cut to make it feasible to realize the desired budget. On NeXT, compute retention per image.} \tabularnewline
        \baselinerule
        % llmoe/compressor/sparse_vlm/sparse_vlm.py; sparse_vlm_llava_next.py
        \baselinecell{SparseVLM~\cite{zhang2025sparsevlm}} & \baselinecell{\frozenicon None (training-free)} & \baselinecell{On LLaVA-1.5 at 32, advance the first cut to reduce unpruned overhead so that it is possible to realize the desired token budget. On NeXT, scale retained counts per image, including recycled tokens in the budget.} \tabularnewline
        \baselinerule
        % llmoe/compressor/pyramiddrop/pyramiddrop.py
        \rowcolor{lightorange!30}
        \baselinecell{PyramidDrop~\cite{xing2025pyramiddrop}} & \baselinecell{\frozenicon Training-free variant; pretrained backbone remains frozen.} & \baselinecell{On LLaVA-1.5 at 128/64/32, advance cuts to reach the desired average token budgets. On NeXT at 640/320/160, advance the first cut to reach low budgets.} \tabularnewline
        \baselinerule
        % llmoe/compressor/nuwa/nuwa.py; nuwa_llava_next.py
        \baselinecell{N\"uwa~\cite{huang2026nuwa}} & \baselinecell{\frozenicon None (training-free)} & \baselinecell{None.} \tabularnewline
        \baselinerule
        % llmoe/compressor/duet/duet.py; docs/duet_llava_next_implementation_strategy.md
        \rowcolor{lightorange!30}
        \baselinecell{DUET-VLM~\cite{singh2026duet}} & \baselinecell{\trainableicon Two-stage training: projector alignment, then joint projector and full-LLM instruction tuning; vision encoder frozen.} & \baselinecell{On NeXT, correct crop packing to match NeXT's unpadded image layout and preserve budget-exempt row separators.} \tabularnewline
        \baselinerule
        % llmoe/compressor/learnpruner/learn_pruner.py; learn_pruner_llava_next.py; learn_pruner_tr.py
        \baselinecell{LearnPruner~\cite{takezoe2026learnpruner}} & \baselinecell{\trainableicon Budget-agnostic training of the scoring MLP on 10\% of data; VLM frozen.} & \baselinecell{On Qwen3.8-27B, we train it on the entire mix-665k dataset since it is external, which is 0.33 percentage points (benchmark-averaged retention) lower on average than using a 10\% subset as the official recipe.} \tabularnewline
        \baselinerule
        % llmoe/compressor/hidrop/hidrop.py; docs/hidrop_llava_next_implementation_strategy.md
        \rowcolor{lightorange!30}
        \baselinecell{HiDrop~\cite{wu2026hidrop}} & \baselinecell{\trainableicon Per-budget projector alignment, then joint projector and full-LLM instruction tuning; vision encoder frozen.} & \baselinecell{On LLaVA-1.5 at 32, add an early cut to reduce unpruned overhead to allow the desired token budget. On NeXT, scale retained counts by image patch count; preserve row structure with budget-exempt separators until visual exit.} \tabularnewline
        \baselinerule
        % llmoe/compressor/core/core.py; core_tr.py; docs/core_llava_next_8b_implementation_decisions.md
        \baselinecell{CORE~\cite{lei2026core}} & \baselinecell{\trainableicon Projector alignment, then joint projector and rank-512 LLM LoRA tuning; vision stack frozen. Budget-agnostic training with 1,024 visual tokens.} & \baselinecell{Use the fixed-rate version of CORE, enforcing an exact $K$-token budget. Replace InternLM2 with Vicuna (LLaVA-1.5) or Llama-3 (LLaVA-NeXT) to match the other methods' LLMs.} \tabularnewline
        \bottomrule
    \end{tabular}
    \label{tab:supp_baseline_adaptations}
    \endgroup
\end{table*}

\begin{figure*}[p]
    \centering
    \includegraphics[width=1\linewidth]{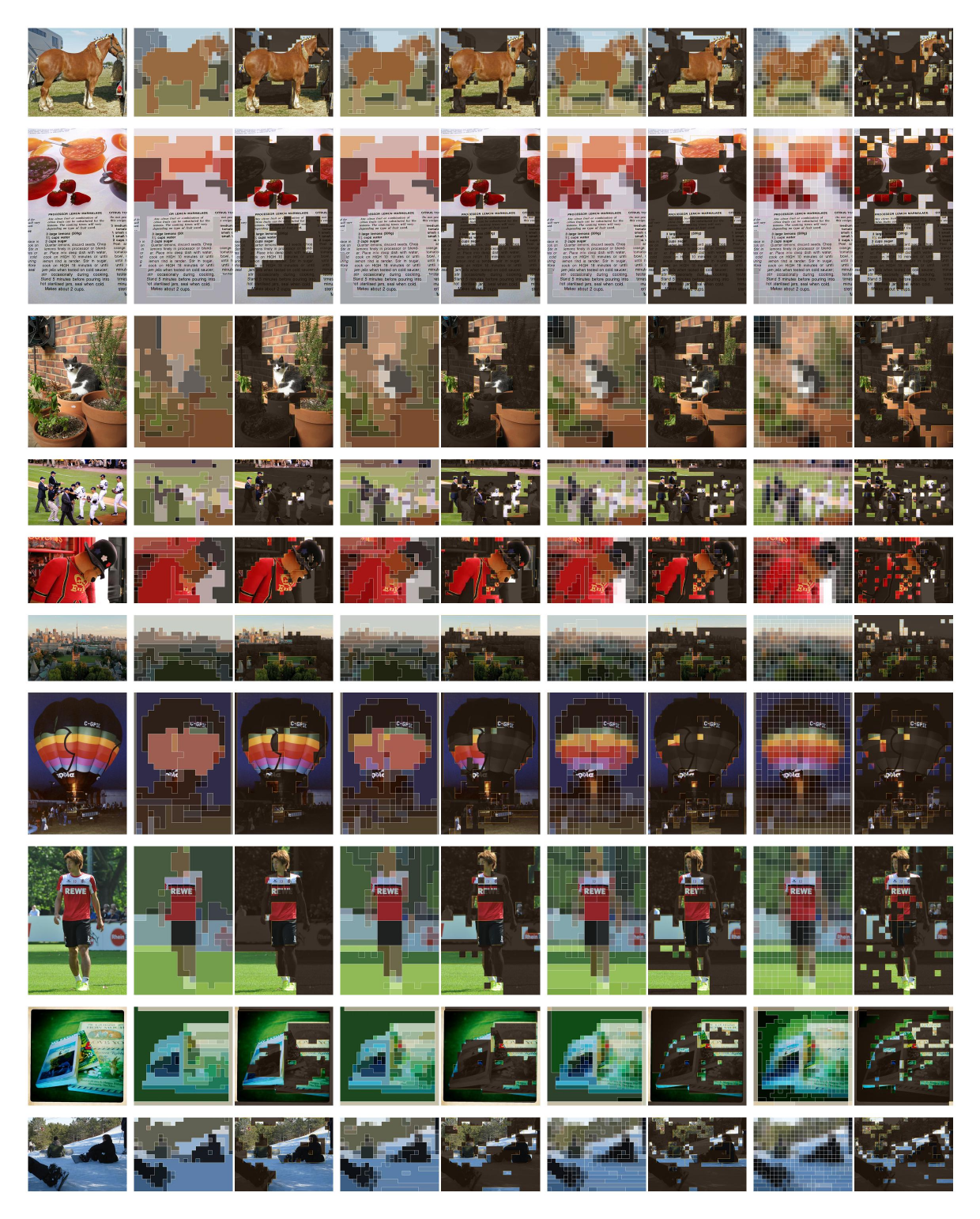}
    \caption{\textbf{Gallery of \method{} tokenization and \texttt{[REL]} necessity pruning on LLaVA-1.5-7B.}}
    \label{fig:gallery_llava15_7b}
\end{figure*}

\begin{figure*}[p]
    \centering
    \includegraphics[width=1\linewidth]{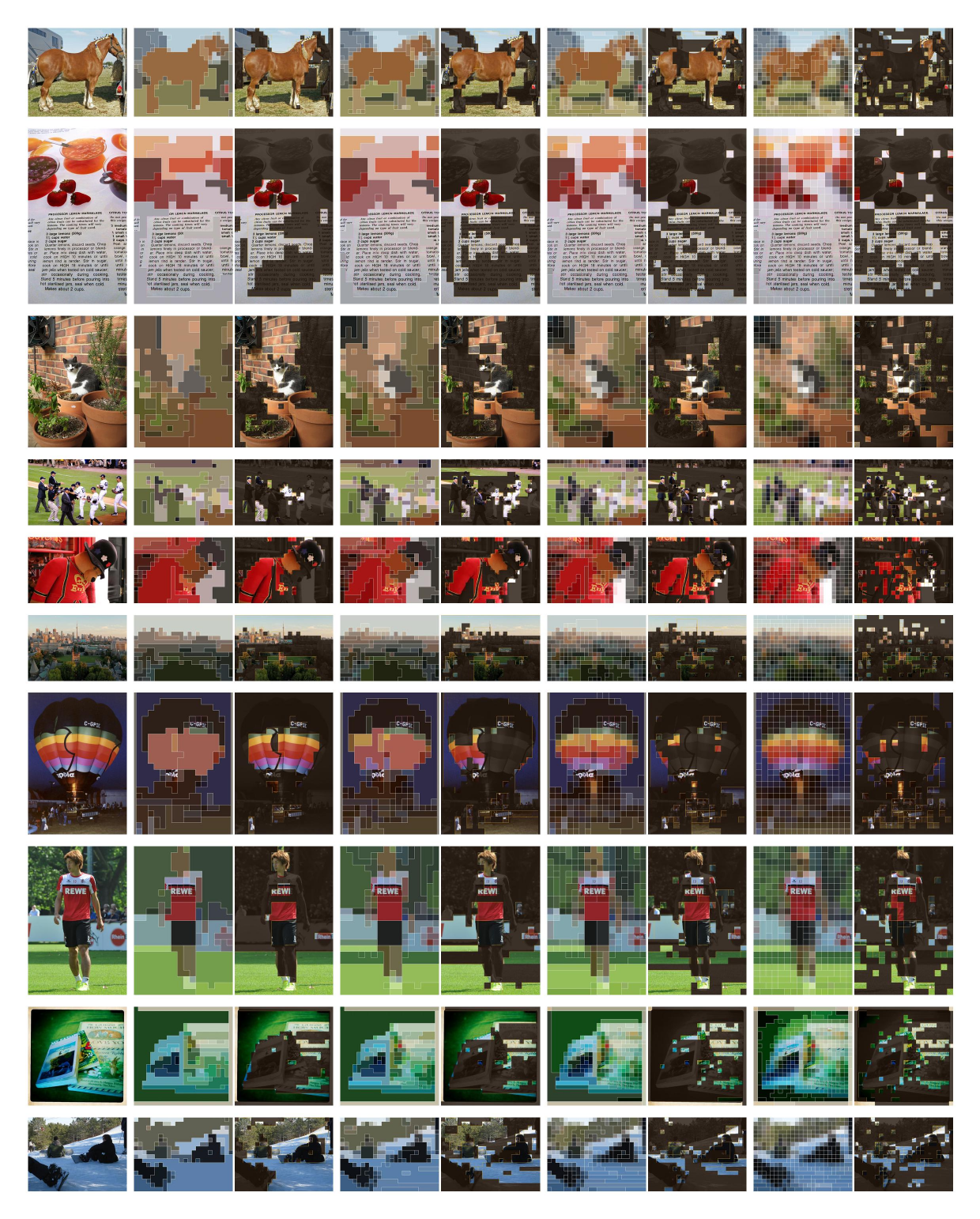}
    \caption{\textbf{Gallery of \method{} tokenization and \texttt{[REL]} necessity pruning on LLaVA-1.5-13B.}}
    \label{fig:gallery_llava15_13b}
\end{figure*}

\begin{figure*}[p]
    \centering
    \includegraphics[width=1\linewidth]{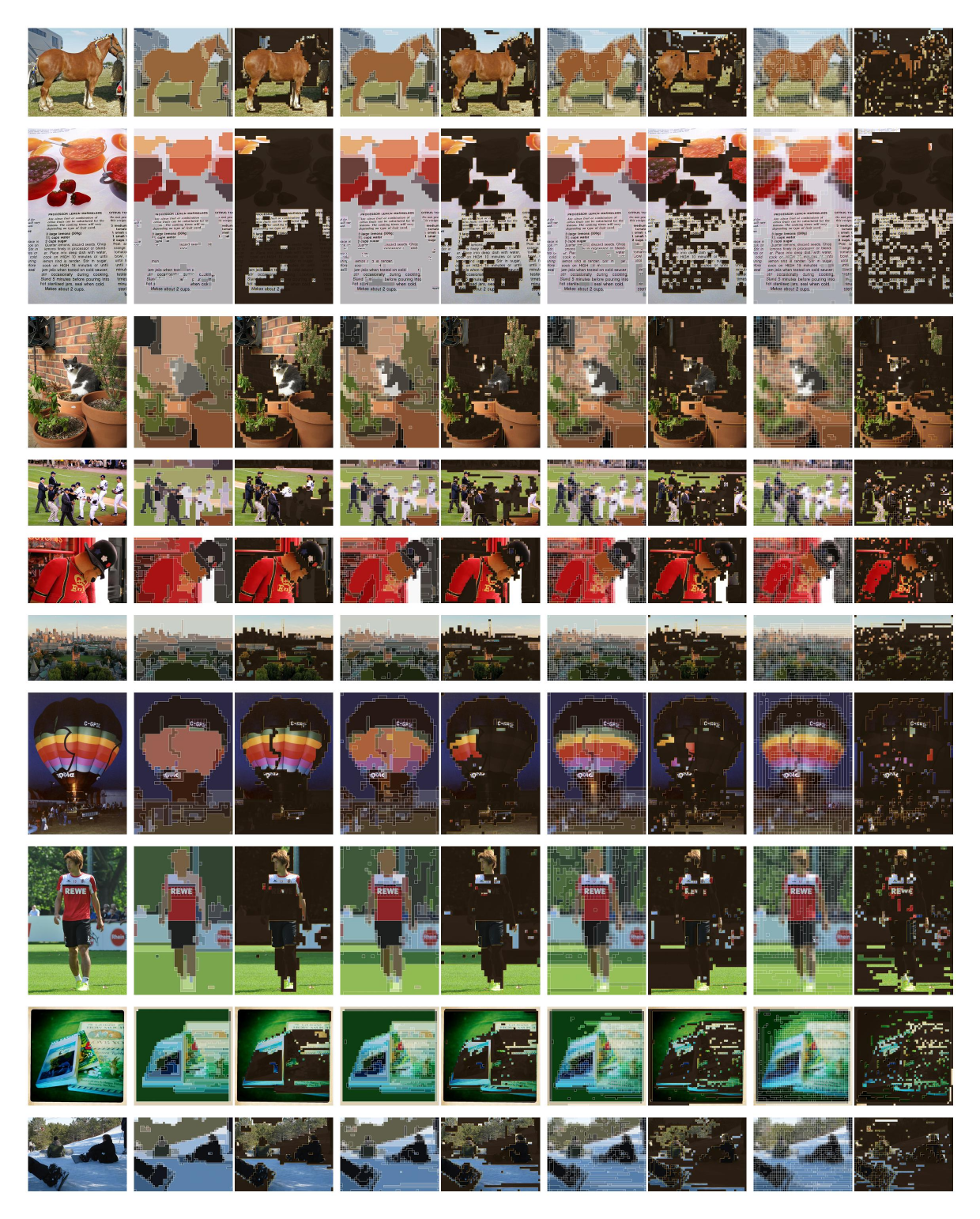}
    \caption{\textbf{Gallery of \method{} tokenization and \texttt{[REL]} necessity pruning on LLaVA-NeXT-8B.}}
    \label{fig:gallery_llava_next}
\end{figure*}

\begin{figure*}[p]
    \centering
    \includegraphics[width=1\linewidth]{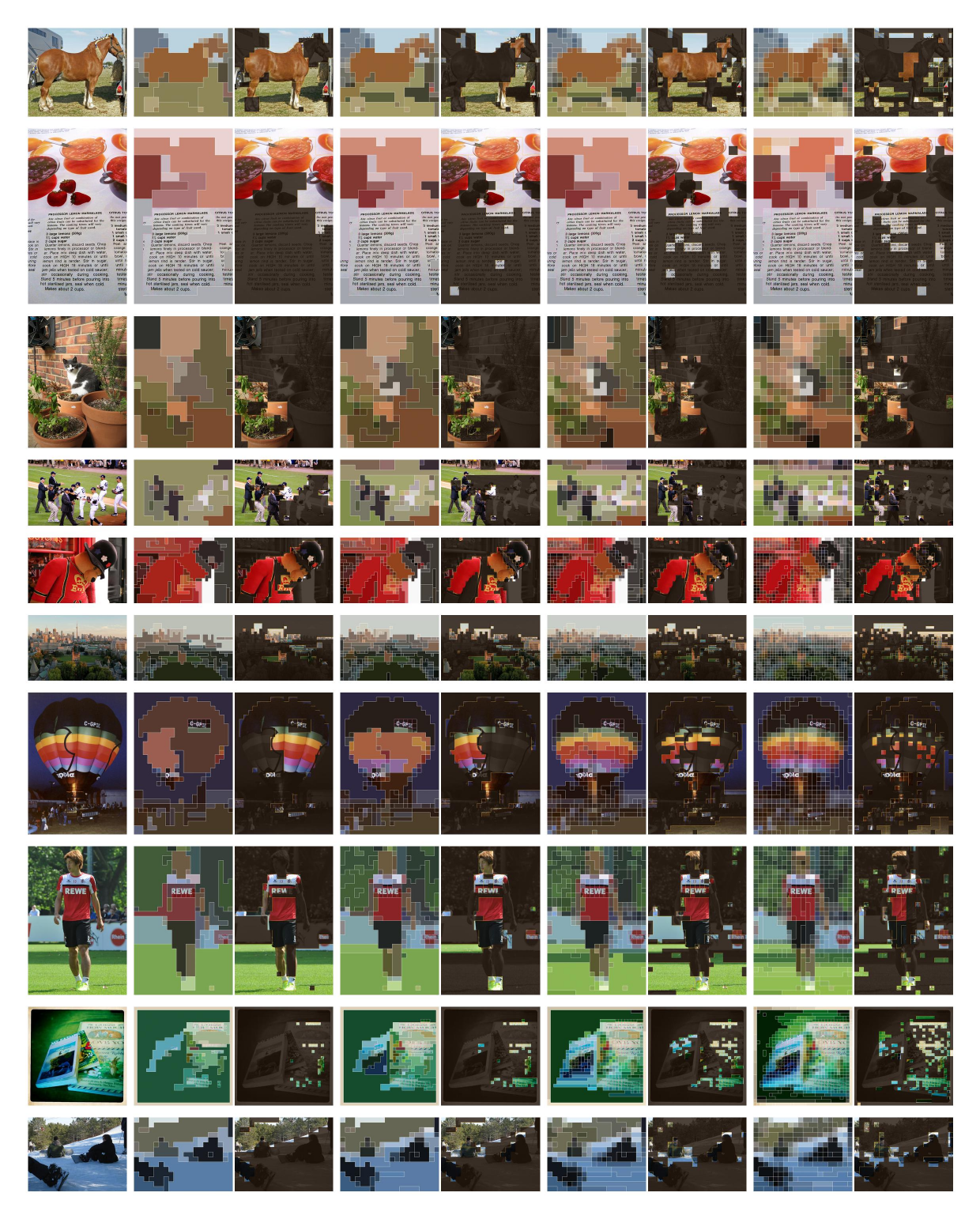}
    \caption{\textbf{Gallery of \method{} tokenization and \texttt{[REL]} necessity pruning on Qwen3.8-27B.}}
    \label{fig:gallery_qwen38_27b}
\end{figure*}

\begin{figure*}[p]
    \centering
    \includegraphics[width=0.96\linewidth]{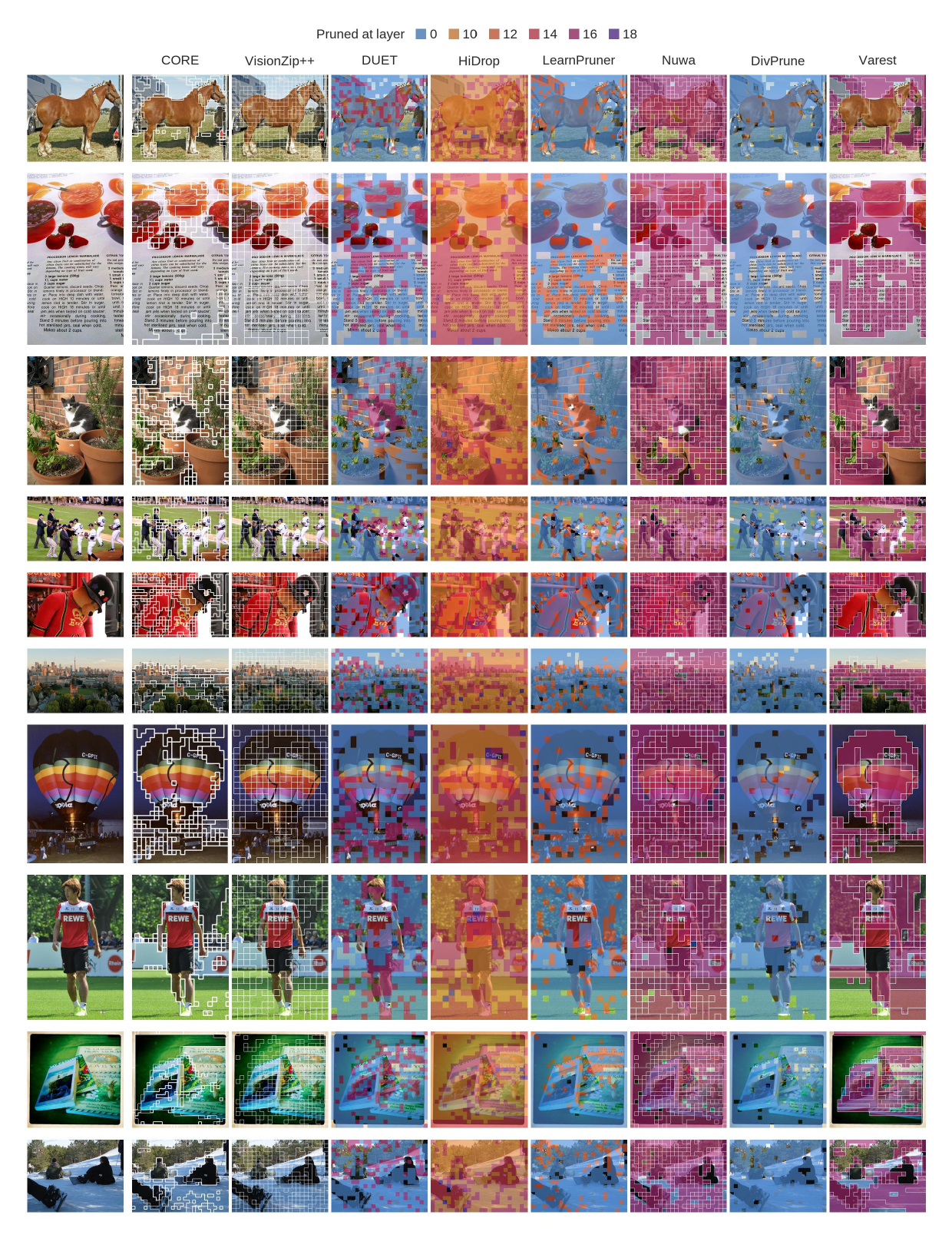}
    \caption{\textbf{Token retention of baselines and \method{} on LLaVA-1.5-7B at $K=64$.}}
    \label{fig:baselines_llava15_7b_b64}
\end{figure*}

\begin{figure*}[p]
    \centering
    \includegraphics[width=0.96\linewidth]{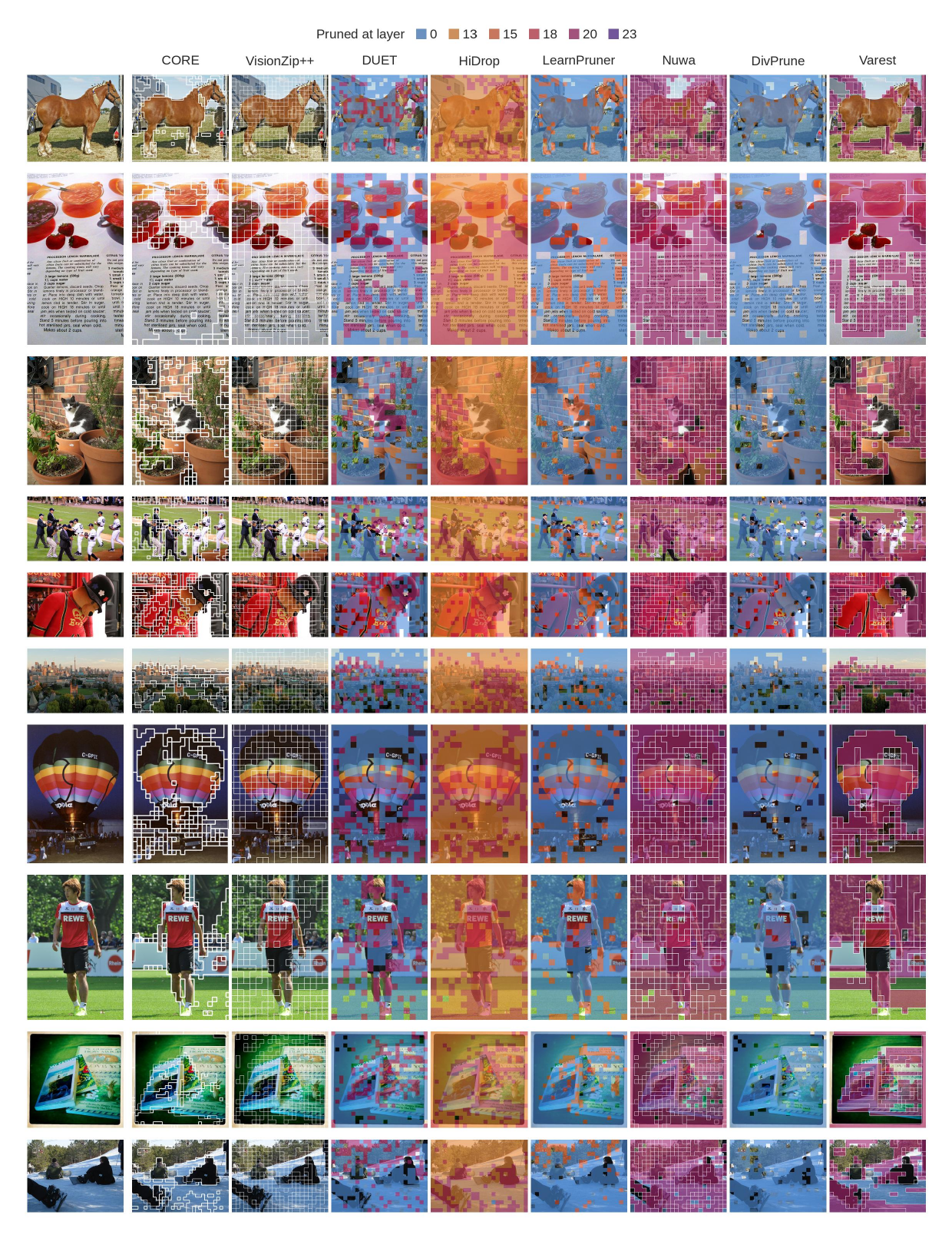}
    \caption{\textbf{Token retention of baselines and \method{} on LLaVA-1.5-13B at $K=64$.}}
    \label{fig:baselines_llava15_13b_b64}
\end{figure*}

\begin{figure*}[p]
    \centering
    \includegraphics[width=0.96\linewidth]{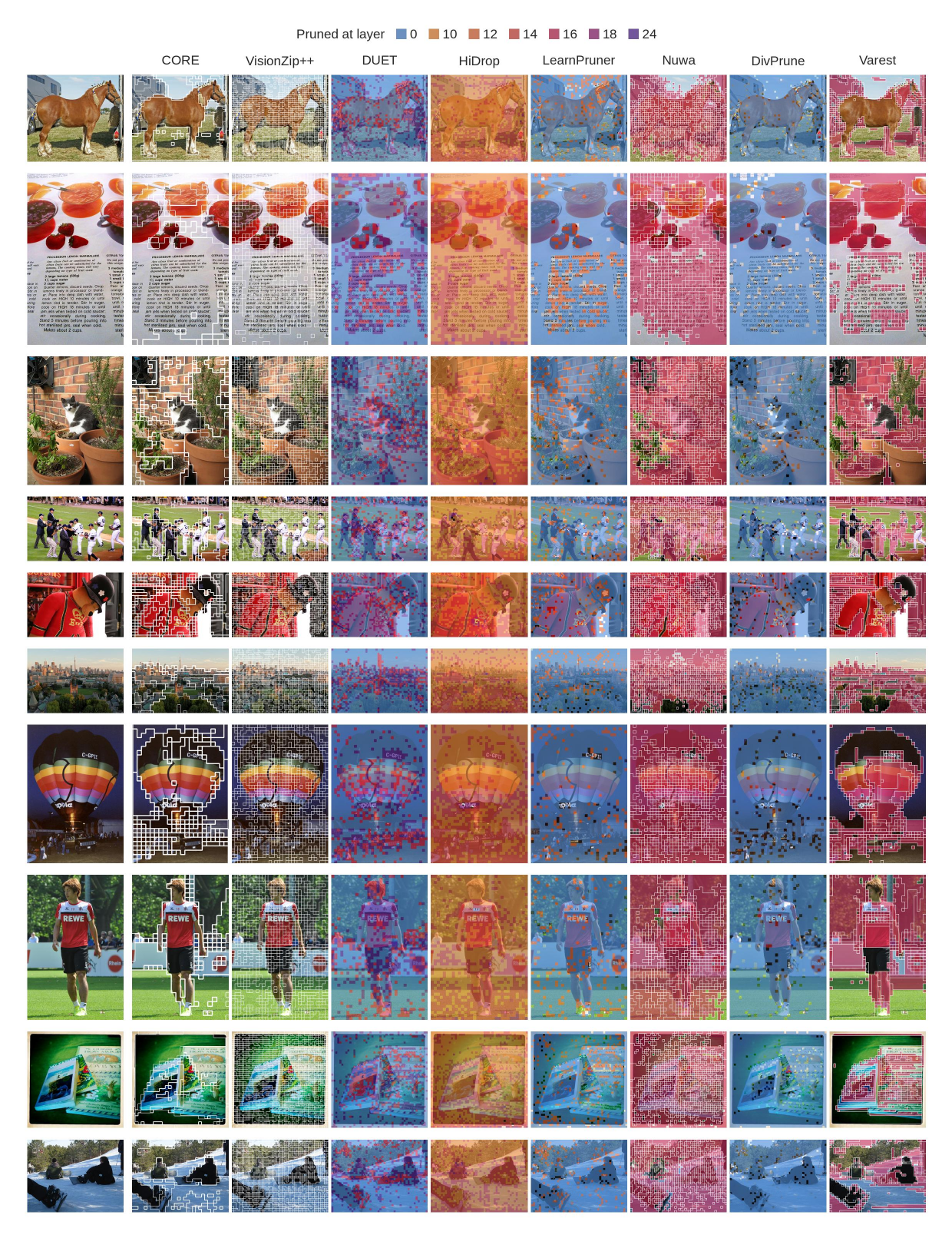}
    \caption{\textbf{Token retention of baselines and \method{} on LLaVA-NeXT-8B at $K=64$.}}
    \label{fig:baselines_llava_next_8b_b64}
\end{figure*}

\end{document}